\pdfoutput=1
\documentclass[10pt,twocolumn,letterpaper]{article}

\usepackage[pagenumbers]{cvpr}
\AddToShipoutPicture{\color[rgb]{.45,.45,.45}\AtTextUpperLeft{\put(0,\LenToUnit{0.8cm}){\parbox{\textwidth}{\centering\cvprtenhv Preprint. Under review.}}}}      

\usepackage{cuted}      
\usepackage{currfile}   
\usepackage{caption}    
\usepackage{graphicx}
\usepackage{amsmath,amssymb}
\usepackage{booktabs}
\usepackage{multirow}
\usepackage{makecell}
\usepackage{xcolor}
\usepackage{xspace}
\usepackage{tikz}
\usetikzlibrary{arrows.meta,positioning,fit,calc,backgrounds}

\newcommand{\R}{\mathbb{R}}
\newcommand{\x}{x}
\newcommand{\z}{z}

\newcommand{\enc}{\mathcal{E}}

\newcommand{\Lexe}{\mathcal{L}_{\mathrm{exe}}}
\newcommand{\Lren}{\mathcal{L}_{\mathrm{ren}}}
\newcommand{\Lpred}{\mathcal{L}_{\mathrm{JEPA}}}
\newcommand{\Lprobe}{\mathcal{L}_{\mathrm{probe}}}
\providecommand{\LDelta}{\mathcal{L}_{\Delta}}

\newcommand{\Csem}{C^{\mathrm{sem}}_t}
\newcommand{\Hout}{H^{\mathrm{out}}_t}
\newcommand{\Qout}{Q_{\mathrm{out}}}

\newcommand{\softmax}{\operatorname{softmax}}
\newcommand{\normalize}{\operatorname{normalize}}

\renewcommand{\dbltopfraction}{0.92}
\renewcommand{\topfraction}{0.92}
\renewcommand{\bottomfraction}{0.85}
\renewcommand{\textfraction}{0.06}
\renewcommand{\floatpagefraction}{0.85}
\renewcommand{\dblfloatpagefraction}{0.85}
\definecolor{cvprblue}{rgb}{0.21,0.49,0.74}
\usepackage[pagebackref,breaklinks,colorlinks,allcolors=cvprblue]{hyperref}

\def\confName{CVPR}
\def\confYear{2027}

\title{Interactive Neural Computers}

\author{Ross Murphy\\
University College London\\
{\tt\small r.murphy@cs.ucl.ac.uk}
\and
Marcel Hedman\\
University of Oxford\\
{\tt\small marcel.hedman@stats.ox.ac.uk}
\and
Gabriel Brostow\\
University College London\\
{\tt\small g.brostow@cs.ucl.ac.uk}
}

\begin{document}
\twocolumn[{
\renewcommand\twocolumn[1][]{#1}
\maketitle
\begin{center}
  \centering
  \captionsetup{type=figure}
  \vspace{-0.2em}
\includegraphics[width=0.60\textwidth]{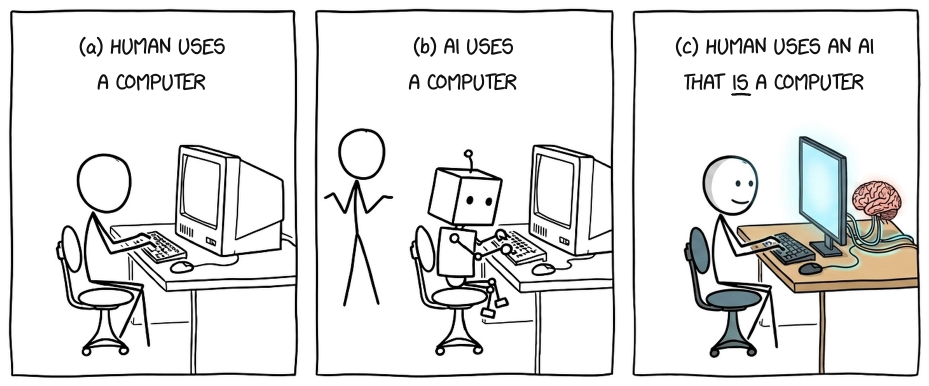}

  \vspace{-0.2em}
  \captionof{figure}{\textbf{From AI \emph{using} a computer to AI \emph{being} a computer.} From today's paradigm, where a human uses a computer (a), the current trend trains an AI agent to use a computer on the user's behalf (b). We take a different path (c) where a human uses an AI that \emph{is} a computer.}
  \label{fig:teaser}
  \vspace{0.1em}
\end{center}
}]

\begin{abstract}
Computers and their interfaces are usually treated as fixed, so a person, or increasingly an AI agent, must adapt to whatever the operating system presents. This is limiting when someone needs a different workflow, or an interface suited to hands-free use or a disability. We explore an overlooked alternative. Rather than training an agent to operate an existing computer, we train a generative model to \emph{be} the computer, predicting each next screen from past screens and the user's mouse and keyboard actions. Treating a desktop as an interactive world model raises problems that natural-video world models largely avoid.
Without a camera pose to index memory, the model must remember content that has left the screen without any geometric cues. Arithmetic in a terminal has to be exactly right, not merely plausible. Small interface text must also stay legible at interactive frame rates. Our Interactive Neural Computer is structured like a Neural Turing Machine and starts to address these problems with three modules trained end to end from pixels and actions. A surprise-gated slot memory, supervised by latent next-frame prediction (\textsc{Memory-JEPA}), stores what is off screen. An exploratory query bridge brings in the reasoning of a pretrained VLM, and a residual renderer predicts only the latent changes between frames. In long closed-loop rollouts, our model recalls occluded files in 64.2\% of trials (chance is 33.3\%), compared with at most 56.0\% for recurrent and
sliding-window baselines. It also raises terminal-text OCR accuracy from 1.3\% to 50.1\%. Finally, because the memory is explicit, it can be transplanted at test time to causally change what the model renders.
\end{abstract}

\section{Introduction}
\label{sec:intro}

Much recent work on computer use aims to build agents that operate a computer
in the way a person does, reading the screen, moving a cursor, and selecting
from menus, to complete tasks on users' behalf
\citep{xie2024osworld,rawles2024androidworld,anthropic2024computeruse}. This setting has three fundamental issues. First, the user is removed from the loop, which serves to replace human capabilities rather than augment them.
Second, removing the user makes execution hard to supervise, which is risky,
given documented deceptive and alignment-faking behavior in frontier models
\citep{hubinger2024sleeper,greenblatt2024alignmentfaking,meinke2024scheming}.
Third, this setting treats computer interfaces as fixed environments to be `completed'. In reality, an interface is an arbitrary construct, so we should care less about
whether an agent can `solve' it \`a la MiniGrid~\citep{chevalierboisvert2023minigrid}, and more about whether an AI can
generate a useful interface that augments human capabilities. Ultimately, if the AI \textit{is} the computer, human-AI collaboration can move toward \textit{open-ended} interfaces. Following \citet{hughes2024oeasi}, we call a system open-ended with respect to an observer when it keeps producing artifacts that the observer finds both novel and learnable. Here the observer is the user, and the artifacts are interfaces that adapt to their evolving workflow while remaining easy to learn, without anyone having to rewrite, recompile, and reload static source code. 

%

We therefore set out to build a generative interactive neural computer (\Cref{fig:teaser}) that learns, directly from pixel and action trajectories, to remember abstract concepts, execute symbolic operations, and render crisp updates, all without a hand-crafted symbolic memory state. Each of these three requirements is an open problem for world models at large, and each takes a specific form on a computer. The first is long-horizon object permanence. World models of 3D scenes can use camera pose and scene geometry to decide where to store an observation and when to retrieve it~\citep{xiao2025worldmem,li2025vmem}, but a desktop offers no such index. A file opened in a pane overwrites exactly the pixels of the file that was there before, so the model must learn for itself what to store, when, and where. The second is symbolic reasoning inside a generated world. Approximate dynamics look convincing in natural video, but a terminal that receives \texttt{3+3} must print \texttt{6} every time. The third is rendering. Most of the screen is static from one frame to the next, which tempts a model to simply copy the previous frame, yet the regions that do change include 10--12\,px text that must stay legible at interactive frame rates.

To address these three problems, we introduce the Interactive Neural Computer (\Cref{fig:architecture}), an end-to-end architecture analogous to a Neural Turing Machine~\cite{graves2014ntm,graves2016dnc} organized into three core modules:
\begin{itemize}[leftmargin=1.3em,itemsep=1pt,topsep=2pt]
  \item \textbf{Memory (surprise-gated concept slots):} To sustain object permanence without camera pose or hand-crafted state, we introduce a coordinate-free concept-slot memory grounded in predictive coding~\cite{rao1999predictive,friston2010free} and surprise-driven memory selection (\textsc{Cambrian-S}~\cite{yang2025cambrians}), whose differentiable writes are gated by latent prediction surprise and self-supervised in representation space (\textsc{Memory-JEPA}).
  \item \textbf{Compute (exploratory single-pass VLM query bridge):} As a preliminary step toward symbolic reasoning inside a generated world, we extend learnable query bridges~\cite{pan2025metaqueries,chen2025blip3o,panteam2025pan} beyond diffusion-only supervision by placing causal queries strictly before training-only outcome text, extracting symbolic execution states into latent tokens ($\Hout$) in a single forward pass without test-time text decoding.
  \item \textbf{Interface (residual latent-diffusion rendering):} To render crisp updates at interactive speeds without collapsing into frame copying, we pair both modules with residual latent diffusion~\cite{yang2023diffusion,salehi2026rfdm} that predicts frame-to-frame latent deltas rather than re-synthesizing the full screen at every step, lifting persistent terminal-text OCR from $0.2\%$ with full-frame diffusion to $43.5\%$ at a matched training budget. Our full model reaches $50.1\%$ at $512{\times}384$, close to the 4-channel codec's $56.3\%$ ceiling, and $88.7\%$ with a 16-channel \texttt{FLUX.1-schnell} autoencoder at $1024{\times}768$.
\end{itemize}

Standard perceptual metrics cannot tell whether a world model remembers what is off screen, because a frame can look plausible while its content is wrong. We therefore evaluate memory directly (\Cref{sec:exp:permanence}). In our hidden-shape recall task, a file containing a shape is closed, the user works elsewhere, and the file is later reopened, so the correct shape can only come from memory. A harder variant first overwrites the same pane with a competing file. Because our memory is an explicit bank rather than part of an entangled hidden state, we can also transplant or erase it at test time and check that what the model renders changes accordingly (\Cref{sec:exp:mechanistic}).

Public GUI corpora rarely revisit occluded content or test symbolic execution after the screen is cleared, so we built two controlled datasets, \emph{Desktop} and \emph{Maths}, that do (\Cref{sec:data}). We will release the full five-source aligned corpus, including both controlled datasets and our conversion and alignment pipelines, together with all training code and model weights.

\section{Related Work}
\label{sec:related}

\paragraph{World models of computers.}
Modeling the machine rather than the agent has begun to be explored in recent work.
Neural Computers~\citep{zhuge2026neural} models the desktop as an open-loop generated video of a user interacting with a computer, which generates compelling open-loop videos, but does not support closed-loop user interaction.
StateCraft~\citep{venturoli2026statecraftaugmentingvideoworld} similarly uses video diffusion as its substrate, but augments its model with an explicit symbolic memory. This helps recall (object permanence), but its design requires access to hand-crafted symbolic accessibility-tree state graphs.
Other concurrent GUI world models rely on external scaffolds or omit persistent memory: Solaris~\citep{alaluf2026solaris} stores state in text prompts, while CUWM~\citep{guan2026cuwm} and \citet{yu2026latentgui} pair VLM or latent predictors with separate decoders without a persistent learned memory bank.
NeuralOS~\citep{rivard2025neuralos}, the precursor to these approaches, is the only open-weight, interactive, pixel-to-pixel desktop world model, available for direct closed-loop rollouts, without privileged symbolic state requirements. 
NeuralOS requires only pixel supervision, but compressing observation history into a single recurrent LSTM state degrades performance over long rollouts.
Our model differs from all of the above in that we learn an addressable external memory end-to-end from pixels and actions without hand-crafted state schemas.

\paragraph{Generative interactive world models and memory.}
More broadly, action-conditioned world models have progressed from single-vector recurrent state spaces~\citep{ha2018worldmodels,hafner2023dreamerv3} to large-scale video generation for driving~\citep{hu2023gaia1}, 2D and 3D worlds~\citep{bruce2024genie,parkerholder2024genie2,parkerholder2025genie3}, and games~\citep{valevski2025diffusion,alonso2024diamond}.
Across all of these domains, world models face a trade-off between storing raw history frames, which scales poorly with rollout length, and compressing history into a fixed-size state, which risks overwriting older information~\citep{graves2014ntm,graves2016dnc,zhang2025framepack,po2025longcontext}.
Fixed-size memory is much easier to manage when the environment provides an external index for reads and writes.
In 3D physical environments, geometry provides that index `for free' via camera pose, field-of-view overlap~\citep{xiao2025worldmem,yu2025context}, 3D surfels~\citep{li2025vmem}, or raymaps~\citep{cetin2026smart,wu2026worldt3}.
A 2D interface offers no such index (\Cref{sec:intro}), so the model must learn \emph{what}, \emph{where}, and \emph{when} to write.
Existing 2D world models avoid this indexing problem by keeping a short sliding window of frames~\citep{alonso2024diamond,valevski2025diffusion,zhang2025framepack} or compressing history into a single recurrent vector~\citep{ha2018worldmodels,hafner2023dreamerv3,rivard2025neuralos}, both of which lose occluded state over long rollouts.
To build an indexed, fixed-capacity memory without 3D geometry, we instead draw on three complementary ideas from the representation-learning and sequence-modeling literature.
First, to decide \emph{where} and \emph{when} to write without spatial coordinates, sequence models such as Differentiable Neural Computers~\citep{graves2014ntm,graves2016dnc}, slot attention~\citep{locatello2020object,zhang2024gated}, and gated delta networks~\citep{schlag2021linear,katharopoulos2020transformers,yang2024gated,gu2024mamba,yang2025gated} address discrete slots by content similarity and free capacity (\emph{where}). Predictive-coding models~\citep{rao1999predictive,friston2010free,sun2025learning,behrouz2025titans} and \textsc{Cambrian-S}~\citep{yang2025cambrians} gate memory writes by whether a new observation is surprising (\emph{when}).
However, content addressing and surprise gating only work if the underlying representations capture high-level semantics rather than static background pixels. For \emph{what} to store, we therefore draw on Joint-Embedding Predictive Architectures~\citep{assran2023ijepa,bardes2024vjepa,zhou2025dinowm,assran2025vjepa2,zhang2026rlawm}, self-supervising our concept slots via latent next-state prediction (\textsc{Memory-JEPA}) rather than pixel reconstruction alone.

\paragraph{Symbolic reasoning in generated worlds.}
Learning symbolic rules and multi-digit arithmetic purely through pixel diffusion~\citep{bruce2024genie,parkerholder2024genie2,parkerholder2025genie3} is sample-inefficient.
Pretrained vision--language models (VLMs)~\citep{bai2025qwen25vl,bai2025qwen3vl} already possess strong symbolic reasoning, though in computer use research they are typically trained as external agents operating existing interfaces~\citep{lee2023pix2struct,shaw2023pix2act,toyama2021androidenv,rawles2023aitw,deng2023mind2web,zhou2024webarena,xie2024osworld,rawles2024androidworld,anthropic2024computeruse}.
We instead bootstrap symbolic execution inside the world model from a pretrained VLM, supervising its query tokens~\citep{pan2025metaqueries,chen2025blip3o,panteam2025pan} with the outcome text rather than only the diffusion loss (\Cref{sec:method:vlm}).

\section{Method}
\label{sec:method}

\definecolor{Steal}{HTML}{3F7A5D}\definecolor{Sviolet}{HTML}{5D4E96}\definecolor{Sblue}{HTML}{3A5A8C}
\definecolor{Sorange}{HTML}{BF7A38}\definecolor{Sgrey}{HTML}{7C7C7C}\definecolor{Sred}{HTML}{B5523B}
\definecolor{SfB}{HTML}{DCE6F2}\definecolor{SfK}{HTML}{ECECEC}\definecolor{Spanel}{HTML}{F5F5F5}\definecolor{SA}{HTML}{6E6E6E}
\newcommand{\Ssub}{\fontsize{5}{5.8}\selectfont}
\newcommand{\Sbg}{Spanel}
\tikzset{
  S/tpos/.style={anchor=north east, xshift=-2pt, yshift=-1.5pt},
  S/f/.style={font=\sffamily\fontsize{5.8}{6.6}\selectfont},
  S/learned/.style={S/f, rectangle, rounded corners=2pt, draw=Sblue, fill=SfB, line width=0.6pt,
                    align=center, inner sep=2pt, minimum height=0.48cm},
  S/frozen/.style={S/f, rectangle, rounded corners=2pt, draw=Sgrey, fill=SfK, line width=0.6pt,
                   align=center, inner sep=2pt, minimum height=0.40cm},
  S/data/.style={S/f, text=black!72, inner sep=2pt},
  S/group/.style={rounded corners=3pt, draw=black!38, line width=0.5pt, dash pattern=on 2pt off 1.5pt},
  S/stage/.style={rounded corners=4pt, draw=#1, fill=Spanel, line width=0.8pt},
  S/title/.style={font=\sffamily\fontsize{6.6}{7.6}\selectfont, text=#1, inner sep=1.5pt}, S/title/.default=black!80,
  S/side/.style={S/f, text=black!58},
  S/legendbox/.style={rounded corners=3pt, draw=black!20, fill=white, line width=0.5pt},
  S/a/.style={-{Stealth[length=3.4pt,width=2.8pt]}, line width=0.6pt, draw=#1, rounded corners=3pt, shorten >=0.6pt},
  S/a/.default=SA,
  S/g/.style={S/a=Sred, dash pattern=on 2pt off 1.5pt},
  S/halo/.style={preaction={draw=white, line width=2.4pt, -}},
  S/loss/.style={circle, draw=Sred, line width=0.6pt, fill=white, minimum size=0.28cm, inner sep=0pt,
                 path picture={\draw[Sred, line width=0.6pt]
                   ([shift={(0.05,0.05)}]path picture bounding box.south west)--([shift={(-0.05,-0.05)}]path picture bounding box.north east)
                   ([shift={(0.05,-0.05)}]path picture bounding box.north west)--([shift={(-0.05,0.05)}]path picture bounding box.south east);}},
  S/lname/.style={font=\fontsize{5.8}{6.8}\selectfont, text=Sred, inner sep=1.2pt},
  S/jn/.style={circle, draw=SA, fill=white, line width=0.6pt, minimum size=0.28cm, inner sep=0pt,
               font=\sffamily\fontsize{6.5}{6.5}\selectfont},
  S/lanelabel/.style={S/f, text=#1, fill=white, inner sep=0.7pt, rotate=90},
}
\newcommand{\Sstrip}[4]{\begin{scope}[shift={(#1,#2)}]\foreach \i in {0,...,#3}{%
  \filldraw[fill=#4!28, draw=#4!80, line width=0.35pt] (\i*0.085,-0.05) rectangle ++(0.085,0.10);}\end{scope}}
\newcommand{\Sbank}[1]{\begin{scope}[shift={(#1)}]\foreach \r in {0,...,3}{\foreach \c in {0,...,6}{%
  \pgfmathsetmacro{\sh}{12+12*mod(\r*5+\c*3,4)}%
  \filldraw[fill=Sviolet!\sh, draw=Sviolet!75, line width=0.25pt] (-0.385+\c*0.11,-0.22+\r*0.11) rectangle ++(0.11,0.11);}}\end{scope}}
\newcommand{\Sscreen}[3]{\IfFileExists{#3}{%
    \node[inner sep=0pt, draw=black!40, line width=0.4pt] (#2) at (#1) {\includegraphics[width=1.0cm]{#3}};}{%
    \node[inner sep=0pt, draw=black!45, line width=0.4pt, fill=white, minimum width=1.0cm, minimum height=0.75cm] (#2) at (#1) {};%
    \fill[Sblue!55] (#2.north west) rectangle ([yshift=-0.08cm]#2.north east);%
    \fill[black!10] ([shift={(0.08,-0.16)}]#2.north west) rectangle ([shift={(0.44,0.08)}]#2.south west);%
    \foreach \k in {0,1,2,3}{\fill[black!30] ([shift={(0.52,-0.20-\k*0.1)}]#2.north west) rectangle ++(0.36,-0.035);}}}
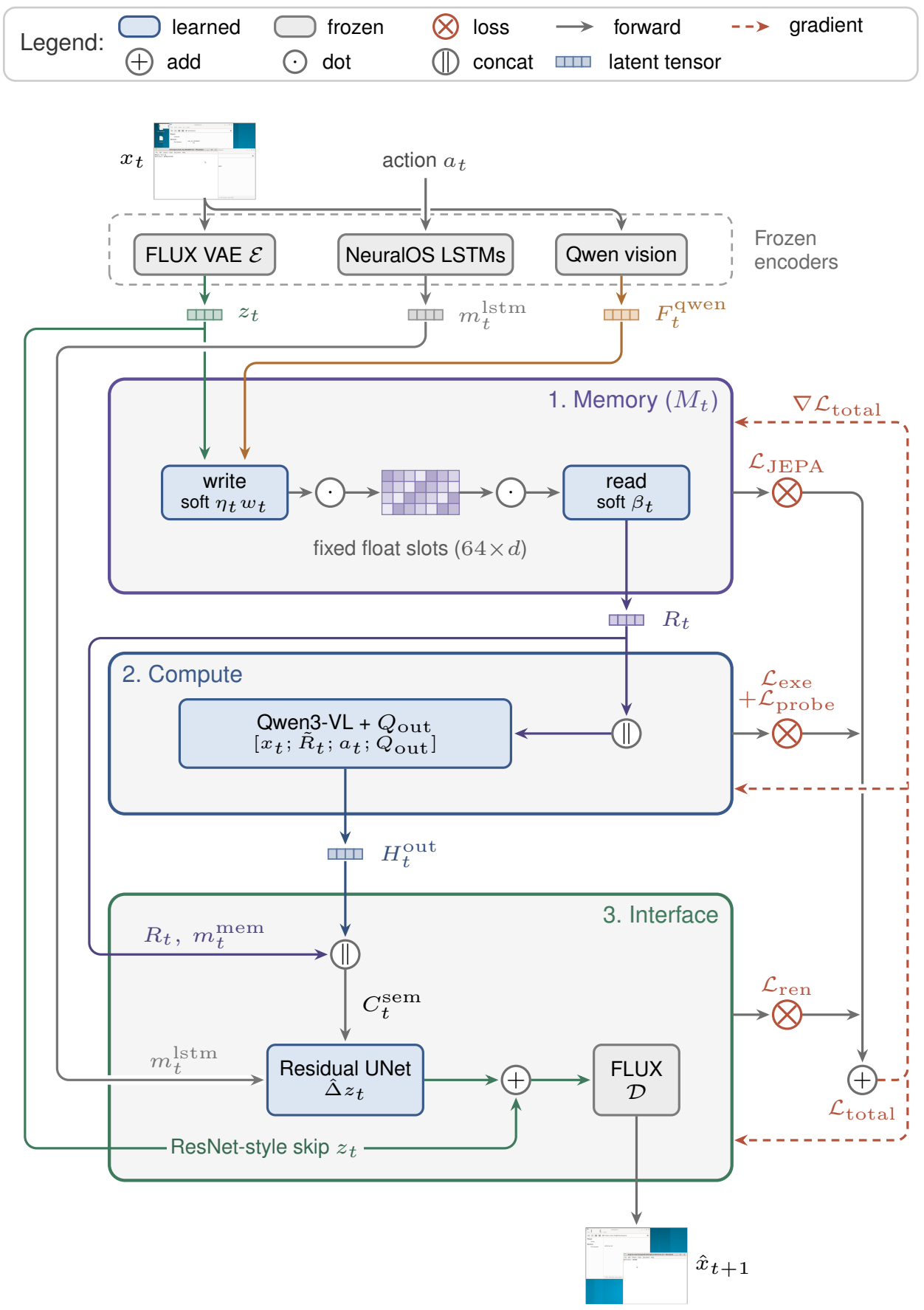
\begin{figure}[t]
  \centering
  \begin{tikzpicture}[S/f, scale=0.90, transform shape]
  \draw[S/legendbox] (-0.05,16.25) rectangle (9.05,16.98);
  \node[S/title, anchor=west] at (0.05,16.615) {Legend:};
  \node[S/learned, minimum width=0.40cm, minimum height=0.20cm] at (1.30,16.79) {}; \node[S/f, anchor=west] at (1.50,16.79) {learned};
  \node[S/frozen,  minimum width=0.40cm, minimum height=0.20cm] at (2.85,16.79) {}; \node[S/f, anchor=west] at (3.05,16.79) {frozen};
  \node[S/loss, minimum size=0.24cm] at (4.35,16.79) {};                          \node[S/f, anchor=west] at (4.50,16.79) {loss};
  \draw[S/a] (5.45,16.79) -- ++(0.40,0);                                           \node[S/f, anchor=west] at (5.90,16.79) {forward};
  \draw[S/g] (7.20,16.79) -- ++(0.40,0);                                           \node[S/f, anchor=west] at (7.65,16.79) {gradient};
  \node[S/jn, minimum size=0.22cm] at (1.30,16.44) {$+$};       \node[S/f, anchor=west] at (1.45,16.44) {add};
  \node[S/jn, minimum size=0.22cm] at (2.85,16.44) {$\cdot$};   \node[S/f, anchor=west] at (3.00,16.44) {dot};
  \node[S/jn, minimum size=0.22cm] at (4.35,16.44) {$\parallel$};\node[S/f, anchor=west] at (4.50,16.44) {concat};
  \Sstrip{5.45}{16.44}{3}{Sblue} \node[S/f, anchor=west] at (5.85,16.44) {latent tensor};

  \Sscreen{1.95,15.45}{xt}{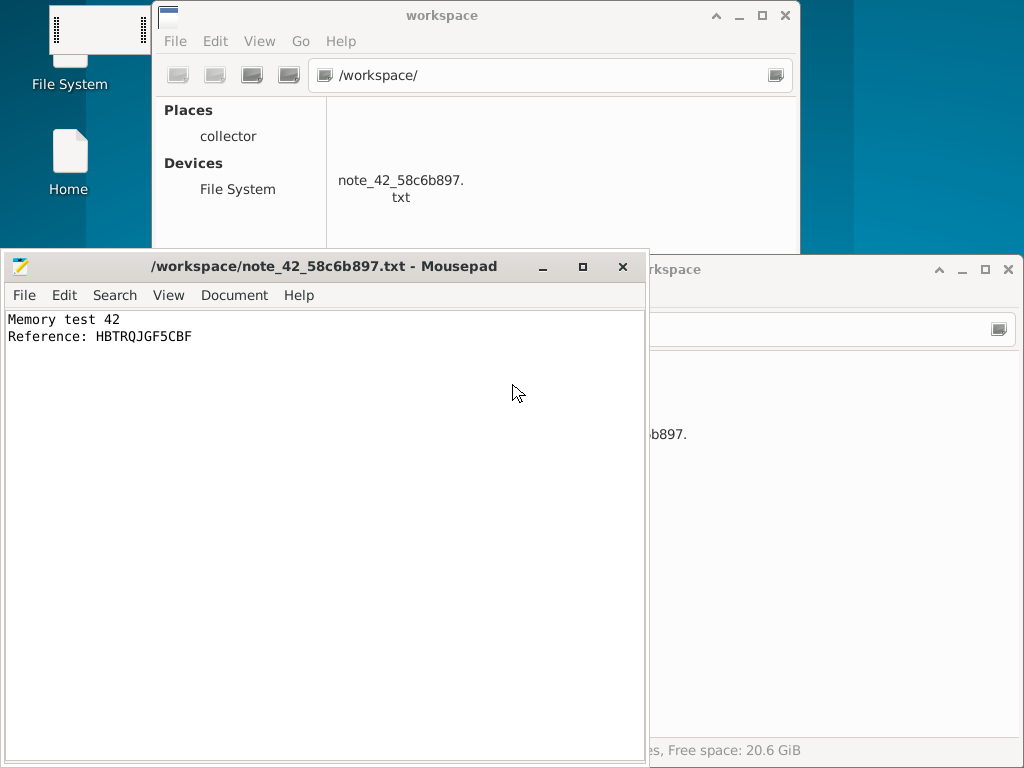}
  \node[S/f, anchor=east, inner sep=2pt] at (xt.west) {$x_t$};
  \node[S/data] (at) at (4.15,15.45) {action $a_t$};
  \draw[S/group] (1.00,14.22) rectangle (7.20,14.92);
  \node[S/side, anchor=west, align=left] at (7.30,14.57) {Frozen\\encoders};
  \node[S/frozen, minimum width=1.40cm] (vae)  at (1.95,14.52) {FLUX VAE $\mathcal{E}$};
  \node[S/frozen, minimum width=1.55cm] (lstm) at (4.15,14.52) {NeuralOS LSTMs};
  \node[S/frozen, minimum width=1.30cm] (qwv)  at (6.10,14.52) {Qwen vision};
  \draw[S/a] (xt.south) -- (vae.north);
  \draw[S/a] (xt.south) -- ++(0,-0.10) -| (qwv.north);
  \draw[S/a, S/halo] (at.south) -- (lstm.north);
  \Sstrip{1.78}{13.92}{3}{Steal}   \node[S/f, anchor=west, text=Steal!85!black]  at (2.15,13.92) {$z_t$};
  \Sstrip{3.98}{13.92}{3}{Sgrey}   \node[S/f, anchor=west, text=Sgrey!95!black]  at (4.35,13.92) {$m^{\mathrm{lstm}}_t$};
  \Sstrip{5.93}{13.92}{3}{Sorange} \node[S/f, anchor=west, text=Sorange!90!black] at (6.30,13.92) {$F^{\mathrm{qwen}}_t$};
  \draw[S/a=Steal]           (vae.south)  -- (1.95,13.99);
  \draw[S/a=Sgrey!95!black]  (lstm.south) -- (4.15,13.99);
  \draw[S/a=Sorange!90!black](qwv.south)  -- (6.10,13.99);

  \draw[S/stage=Sviolet] (1.00,11.15) rectangle (7.20,13.28);
  \node[S/title=Sviolet, S/tpos] at (7.20,13.28) {1.\ Memory ($M_t$)};
  \node[S/learned, minimum width=1.25cm] (mw) at (2.15,12.15) {write\\[-1pt]{\Ssub soft $\eta_t w_t$}};
  \node[S/jn] (wd) at (3.20,12.15) {$\cdot$};
  \Sbank{4.10,12.15}
  \node[S/f, text=black!65, anchor=north] at (4.10,11.80) {fixed float slots ($64{\times}d$)};
  \node[S/jn] (rd) at (5.00,12.15) {$\cdot$};
  \node[S/learned, minimum width=1.25cm] (mr) at (6.15,12.15) {read\\[-1pt]{\Ssub soft $\beta_t$}};
  \draw[S/a] (mw.east) -- (wd.west);
  \draw[S/a] (wd.east) -- (3.72,12.15);
  \draw[S/a] (4.48,12.15) -- (rd.west);
  \draw[S/a] (rd.east) -- (mr.west);
  \draw[S/a=Steal] (1.95,13.85) -- (1.95,12.45);
  \draw[S/a=Sorange!90!black] (6.10,13.85) -- (6.10,13.44) -- (2.35,13.44) -- (2.35,12.45);
  \draw[S/a=Sviolet!85!black] (mr.south) -- (6.15,10.95);
  \Sstrip{5.98}{10.88}{3}{Sviolet} \node[S/f, anchor=west, text=Sviolet!85!black] at (6.37,10.88) {$R_t$};

  \draw[S/stage=Sblue] (1.00,8.95) rectangle (7.20,10.55);
  \node[S/title=Sblue, anchor=north west, xshift=2pt, yshift=-1.5pt] at (1.00,10.55) {2.\ Compute};
  \node[S/learned, minimum width=3.30cm, minimum height=0.68cm] (vlm) at (3.35,9.75)
        {Qwen3-VL + $Q_{\mathrm{out}}$\\[-1pt]{\Ssub $[x_t;\tilde R_t;a_t;Q_{\mathrm{out}}]$}};
  \node[S/jn] (vc) at (6.15,9.75) {$\parallel$};
  \draw[S/a=Sviolet!85!black] (6.15,10.82) -- (vc.north);
  \draw[S/a=Sviolet!85!black] (vc.west) -- (vlm.east);
  \draw[S/a=Sblue!90!black] (vlm.south) -- (3.35,8.62);
  \Sstrip{3.18}{8.55}{3}{Sblue} \node[S/f, anchor=west, text=Sblue!90!black] at (3.57,8.55) {$H^{\mathrm{out}}_t$};

  \draw[S/stage=Steal] (1.00,5.30) rectangle (7.20,8.15);
  \node[S/title=Steal, S/tpos] at (7.20,8.15) {3.\ Interface};
  \node[S/jn] (ic) at (3.35,7.55) {$\parallel$};
  \draw[S/a=Sblue!90!black] (3.35,8.48) -- (ic.north);
  \node[S/learned, minimum width=1.55cm, minimum height=0.70cm] (unet) at (3.35,6.30) {Residual UNet\\[-1pt]{\Ssub $\hat\Delta z_t$}};
  \draw[S/a] (ic.south) -- node[S/f, right=1pt]{$C^{\mathrm{sem}}_t$} (unet.north);
  \node[S/jn] (add) at (5.05,6.30) {$+$};
  \node[S/frozen, minimum width=0.85cm, minimum height=0.70cm] (dec) at (6.25,6.30) {FLUX\\$\mathcal{D}$};
  \draw[S/a=Steal] (unet.east) -- (add.west);
  \draw[S/a=Steal] (add.east) -- (dec.west);
  \Sscreen{6.25,4.45}{xo}{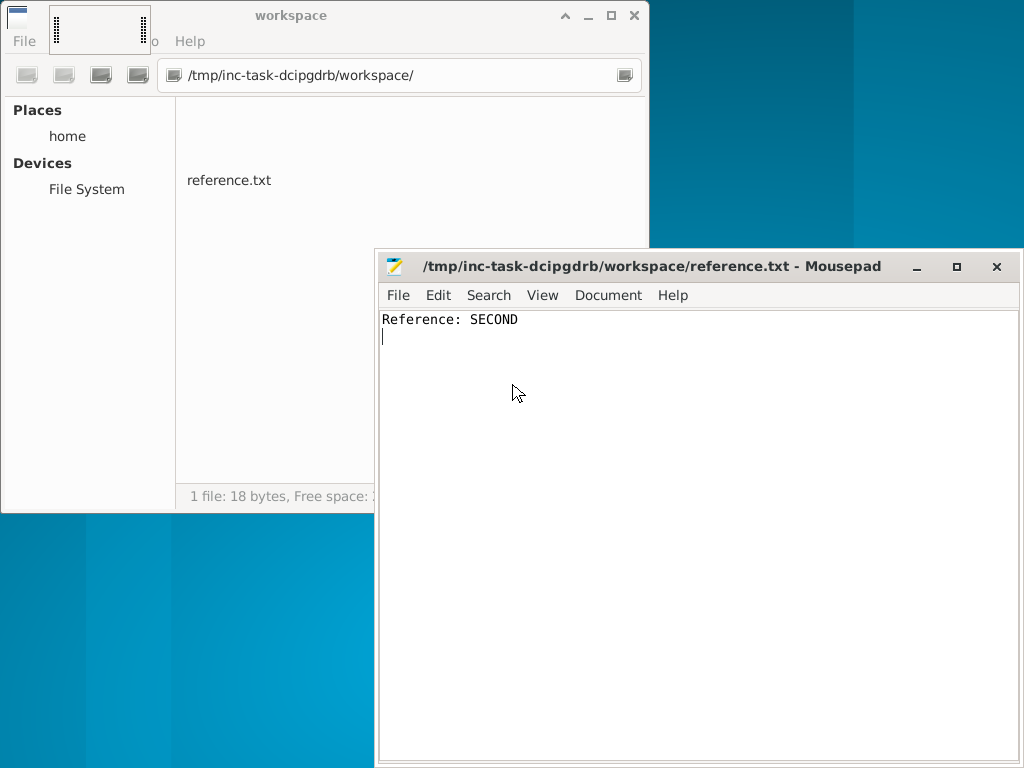}
  \node[S/f, anchor=west, inner sep=2pt] at (xo.east) {$\hat x_{t+1}$};
  \draw[S/a] (dec.south) -- (xo.north);

  \draw[S/a=Sviolet!85!black] (6.15,10.70) -- (0.80,10.70) -- (0.80,7.55) -- (ic.west);
  \node[S/f, text=Sviolet!85!black, anchor=south, inner sep=1.5pt] at (1.95,7.55) {$R_t,\ m^{\mathrm{mem}}_t$};
  \draw[S/a=Sgrey!95!black, S/halo] (4.15,13.85) -- (4.15,13.60) -- (0.48,13.60) -- (0.48,6.30) -- (unet.west);
  \node[S/f, text=Sgrey!95!black, anchor=south, inner sep=1.5pt] at (1.75,6.30) {$m^{\mathrm{lstm}}_t$};
  \draw[S/a=Steal] (1.95,13.78) -- (0.16,13.78) -- (0.16,5.62) -- (5.05,5.62) -- (add.south);
  
  \node[S/f, text=Steal!85!black, fill=\Sbg, inner sep=1pt] at (2.55,5.62) {ResNet-style skip $z_t$};

  \node[S/loss] (lj) at (7.75,12.15) {}; \node[S/lname, anchor=south] at (lj.north) {$\mathcal{L}_{\mathrm{JEPA}}$};
  \node[S/loss] (le) at (7.75,9.75)  {}; \node[S/lname, anchor=south, align=center] at (le.north) {$\mathcal{L}_{\mathrm{exe}}$\\[-1pt]$+\mathcal{L}_{\mathrm{probe}}$};
  \node[S/loss] (lr) at (7.75,6.95)  {}; \node[S/lname, anchor=south] at (lr.north) {$\mathcal{L}_{\mathrm{ren}}$};
  \draw[S/a] (7.20,12.15) -- (lj.west);
  \draw[S/a] (7.20,9.75)  -- (le.west);
  \draw[S/a] (7.20,6.95)  -- (lr.west);
  \node[S/jn] (lt) at (8.50,6.30) {$+$};
  \node[S/lname, anchor=north] at (lt.south) {$\mathcal{L}_{\mathrm{total}}$};
  \draw[S/a] (lj.east) -- (8.50,12.15) -- (lt.north);
  \draw[S/a] (le.east) -- (8.50,9.75);
  \draw[S/a] (lr.east) -- (8.50,6.95);
  \draw[S/g] (lt.east) -- (8.95,6.30) -- (8.95,12.85) -- (7.20,12.85);
  \draw[S/g, S/halo] (8.95,9.20) -- (7.20,9.20);
  \draw[S/g] (8.95,6.30) -- (8.95,5.70) -- (7.20,5.70);
  \node[S/lname, anchor=south] at (8.25,12.87) {$\nabla\mathcal{L}_{\mathrm{total}}$};
  \end{tikzpicture}
  \vspace{-6pt}
  \caption{\textbf{End-to-end Interactive Neural Computer.} Data flows vertically through frozen encoders and three learned modules: \textbf{Memory} ($M_t$), \textbf{Compute} ($Q_{\mathrm{out}}$), and \textbf{Interface} ($\hat{\Delta}z_t$). 
  }
  \label{fig:architecture}
  \vspace{-10pt}
\end{figure}

An interactive world model predicts the next video frame $\x_{t+1} \in \R^{3\times H\times W}$ from observations $\x_{1:t}$ and user actions $a_{1:t}$ (mouse coordinates, clicks, and keystrokes).
Following the three problems set out in \Cref{sec:intro}, we split the model into three modules, Memory, Compute, and Interface (\Cref{fig:architecture}), and train them jointly end to end.


\subsection{Memory}
\label{sec:method:memory}
\label{sec:method:jepa}

Following Neural Turing Machines~\cite{graves2014ntm} and Differentiable Neural Computers~\cite{graves2016dnc}, we separate computation from memory in an explicit, fixed-capacity bank $M_t = (K_t \in \R^{M\times 64}, V^{\mathrm{sem}}_t \in \R^{M\times 512}, V^{\mathrm{vis}}_t \in \R^{M\times 8\times 128}, u_t \in [0,1]^M)$ of $M=64$ concept slots ($m \in \{1,\dots,M\}$), storing a unit-norm key $K_{t,m}$, semantic vector $V^{\mathrm{sem}}_{t,m}$, $8$ visual tokens $V^{\mathrm{vis}}_{t,m}$, and scalar occupancy $u_{t,m}$.
The separation of the memory component also opens the door to greater interpretability and causal interventions later down the line (such as transplanting a donor memory $M^{\mathrm{donor}}_t$ into a running rollout, for example (\Cref{sec:exp:mechanistic})).
Operating $M_t$ reduces to four decisions, namely \emph{what}, \emph{when}, and \emph{where} to write, and \emph{how} to read (\Cref{fig:architecture}, Module~1, with a cycle diagram in Figure~S1 of Section~S1).

\paragraph{What to write: intrinsic keys and self-supervised \textsc{Memory-JEPA}.}
At step $t$, $8$ learned queries cross-attend over frozen vision features $F^{\mathrm{qwen}}_t$ (\texttt{Qwen2.5-VL-3B-Instruct}~\cite{bai2025qwen25vl} at $512{\times}384$, and \texttt{Qwen3-VL-2B-Instruct}~\cite{bai2025qwen3vl} at $1024{\times}768$) and the latent frame encoding $\z_t = \enc(\x_t)$ from our frozen autoencoder (16-channel \texttt{FLUX.1-schnell} f8 VAE~\cite{blackforestlabs2024flux} at $1024{\times}768$, and 4-channel Stable Diffusion KL-VAE at $512{\times}384$) to extract tokens $O_t \in \R^{8\times 512}$ and a pooled summary $s_t \in \R^{512}$.
As desktop frames generally share a static background, we subtract a running EMA background vector $\mu_s$ before unit-sphere projection to form the write key $k_t = \normalize(W_k(s_t - \mu_s)) \in \mathbb{S}^{63}$, while combining the centered summary, action embedding $e(a_t)$, and $O_t$ into the slot write values $(v^{\mathrm{sem}}_t, v^{\mathrm{vis}}_t)$.

When we initially trained these write and read heads using only the outer diffusion loss $\Lren$, the memory bank failed to learn (\Cref{sec:exp:ablations}). The few pixels that depend on memory were swamped by static background, and nothing in the loss told the read head which stored key to ask for, leaving recall at chance.
To fix this problem without requiring ground-truth slot labels, we introduce an explicit self-supervised objective directly on the memory module, \textsc{Memory-JEPA}~\cite{assran2023ijepa,bardes2024vjepa}, which trains a predictor $g_\theta$ to predict the representation of the true next frame $\x_{t+1}$ from only the current screen summary $s_t$, the user's action $e(a_t)$, and the memory readout $r^{\mathrm{sem}}_t$ (\Cref{eq:read}), so that its prediction is
\begin{equation}
  \hat{s}_{t+1} = \normalize\!\big(s_t + g_\theta([s_t;\, e(a_t);\, r^{\mathrm{sem}}_t])\big).
  \label{eq:jepa_pred}
\end{equation}
To obtain a stable target, we follow BYOL~\cite{grill2020byol} and I-JEPA~\cite{assran2023ijepa} and pass the ground-truth next frame $\x_{t+1}$ through an exponential moving average (EMA) copy of the frame encoder with gradients stopped ($\mathrm{sg}$), producing the target summary $s^*_{t+1}$ and target key $k^*_{t+1}$.
Why does this force the memory to learn across long occlusions? Suppose at step $t$ the user clicks a tab ($a_t$) to reopen a file closed $\Delta t$ steps earlier. The current screen $\x_t$ and click $a_t$ contain no information about what was drawn inside that file, so $g_\theta$ can match $s^*_{t+1}$ only if the write head stored the file in $M$ before it closed and the read head retrieved it into $r^{\mathrm{sem}}_t$.
\textsc{Memory-JEPA} supervises \Cref{eq:jepa_pred} at every step via $\Lpred = \mathcal{L}_{\mathrm{InfoNCE}}(\hat{s}_{t+1}, s^*_{t+1}) + \lambda_{\mathrm{align}}\|\hat{s}_{t+1} - s^*_{t+1}\|_2^2 + \lambda_q(1 - \langle q^{(1)}_t, k^*_{t+1}\rangle) + \lambda_{\mathrm{VICReg}}\mathcal{L}_{\mathrm{VICReg}}(k_t) + \lambda_{\mathrm{swap}}\mathcal{L}_{\mathrm{swap}}$. The first term is the InfoNCE contrastive loss~\cite{oord2018cpc}, which pulls the prediction $\hat{s}_{t+1}$ toward its own target $s^*_{t+1}$ and away from the targets of other samples in the batch, and the second term aligns the two directly. The third term aligns the first read query $q^{(1)}_t$ (\Cref{eq:read}) with the target key $k^*_{t+1}$, so the read address is supervised without a gradient having to cross the occlusion. The VICReg term~\cite{bardes2022vicreg} keeps each key dimension's variance high and decorrelates the dimensions, which prevents all keys from collapsing to the same vector. Finally, $\mathcal{L}_{\mathrm{swap}}$ requires the prediction to degrade when $M_t$ is swapped for another episode's memory (Equation~(S1)).

\paragraph{When to write: predictive-coding surprise gate.}

Writing unconditionally on every frame exhausts a fixed quantity of ($M$) slots quite quickly, and can also result in writing self-generated rollout noise into memory. We take inspiration from predictive coding theory~\cite{rao1999predictive,friston2010free} and surprise-driven memory selection~\cite{yang2025cambrians}, and employ a measure of `prediction surprise' to modulate our memory-writes. In a nutshell, if the actual observed frame $\x_t$ confounds our model's expectations around what it \textit{thought} it was going to see at step $t$ ($\hat{s}_t$), then we say it is `surprising'. Formally, prediction surprise $\varepsilon_t$ is the cosine distance between $s_t$ and $\mathrm{sg}(\hat{s}_t)$, normalized by its running average to form the write gate $\eta_t \in (0,1)$, computed as
\begin{equation}
  \begin{aligned}
  \varepsilon_t &= 1 - \big\langle \normalize(s_t),\, \normalize(\mathrm{sg}(\hat{s}_t)) \big\rangle, \\[-2pt]
  \eta_t &= \sigma\!\Big(\alpha_s \Big(\frac{\varepsilon_t}{\mathrm{EMA}(\varepsilon) + 10^{-5}} - \beta_s\Big)\Big).
  \end{aligned}
  \label{eq:surprise_gate}
\end{equation}
Predictable frames keep $\eta_t \approx 0$, whereas unpredicted transitions open $\eta_t \to 1$.
Because $g_\theta$ is inaccurate early in training, normalizing $\varepsilon_t$ by $\mathrm{EMA}(\varepsilon)$ ensures $\eta_t$ responds only to \emph{relative} surprise spikes, while a learned per-slot factor, $\eta_{t,m} = \eta_t\,\sigma(w_g^\top[V^{\mathrm{sem}}_{t-1,m};\, v^{\mathrm{sem}}_t] + b_g)$ ($b_g{=}1.0$ at initialization), lets outer gradients modulate writes during warm-up.

\paragraph{Where to write: a matching slot, or else the first free one.}
A write should update the slot that already holds this content, if there is one, and otherwise claim an empty slot.
We compute both candidate destinations following the Differentiable Neural Computer (DNC)~\cite{graves2016dnc} (Equation~(S2) in Section~S1).
\emph{Content addressing} $a_{c,m}$ points at the occupied slot whose key best matches $k_t$, like a soft nearest-neighbor lookup, and \emph{allocation} $a_{a,m}$ points at the first empty slot, since a slot is penalized while any earlier slot $j < m$ is still free.
A gate $\gamma_t = \sigma(\kappa(\max_m u_{t-1,m}\langle K_{t-1,m}, k_t\rangle - \vartheta)) \in (0,1)$, which measures how well $k_t$ matches its best occupied slot, chooses between them via $w_{t,m} = \gamma_t\, a_{c,m} + (1-\gamma_t)\, a_{a,m}$.
Re-editing a stored file gives $\gamma_t \approx 1$ and updates its slot, whereas a new file gives $\gamma_t \approx 0$ and takes a free one.
Scaling by the surprise gate gives the write strength $e_{t,m} = \eta_{t,m}\, w_{t,m} \in [0,1]$, so a slot is written only when it is the chosen destination \emph{and} the frame is surprising.
Each slot's key, semantic vector, and visual tokens then move toward $(k_t, v^{\mathrm{sem}}_t, v^{\mathrm{vis}}_t)$ in proportion to $e_{t,m}$, so an unaddressed slot is left untouched and a strongly addressed one is overwritten.
Its occupancy $u_{t,m} \in [0,1]$ rises toward $1$ when written and is lowered by a forget gate $f_{t,m}$ that releases slots whose contents have been superseded.
Together, these give the slot updates
\begin{equation}
  \begin{aligned}
  K_{t,m} &= \operatorname{clip}_1\!\big((1 - e_{t,m}) K_{t-1,m} + e_{t,m} k_t\big), \\[-2pt]
  V^{\mathrm{sem}}_{t,m} &= (1 - e_{t,m}) V^{\mathrm{sem}}_{t-1,m} + e_{t,m} v^{\mathrm{sem}}_t, \\[-2pt]
  V^{\mathrm{vis}}_{t,m} &= (1 - e_{t,m}) V^{\mathrm{vis}}_{t-1,m} + e_{t,m} v^{\mathrm{vis}}_t, \\[-2pt]
  u_{t,m} &= \big(u_{t-1,m} + (1 - u_{t-1,m})\, e_{t,m}\big)\,(1 - \eta_t f_{t,m}),
  \end{aligned}
  \label{eq:mem_update}
\end{equation}
where $\operatorname{clip}_1(v) = v / \max(1, \|v\|_2)$ keeps keys in the unit ball and leaves empty slots at zero norm, and multiplying $f_{t,m}$ by $\eta_t$ releases slots only at surprising steps, so memory does not decay while the screen is static.

\paragraph{How to read: action-conditioned predictive retrieval.}
Once $M_t$ has been updated, the model must decide what information to read back out to render the next frame $\x_{t+1}$. In many cases, the current screen $\x_t$ alone does not tell us which stored slot is needed next. We also need the user's action $a_t$. For instance, if a user is viewing \texttt{file\_B} on screen ($\x_t$) and clicks the tab for \texttt{file\_A} ($a_t$), it is the combination of the screen and the click that tells the model to retrieve \texttt{file\_A} from memory. Each of our $H_r=4$ read heads therefore forms its query $q^{(h)}_t = \normalize(\phi^{(h)}_q([s_t - \mu_s; e(a_t)]))\in\mathbb{S}^{63}$ from both the centered screen summary and the action, and attends over the occupied slots. A learnable no-op logit $b_\emptyset$ lets it abstain when no off-screen memory is needed, and the attention weights are
\begin{equation}
  \begin{aligned}
  \beta^{(h)}_{t} = \softmax_{m\in\{1,\dots,M,\emptyset\}}\!\Big(&\Big[\frac{\langle K_{t,m}, q^{(h)}_t\rangle}{\tau_r} \\[-3pt]
  &+ \log(u_{t,m}{+}\epsilon),\; b_\emptyset\Big]\Big)_{1:M},
  \end{aligned}
  \label{eq:read}
\end{equation}
where each head returns one semantic vector $r^{\mathrm{sem}}_t$ and $8$ visual tokens ($4{\times}9 = 36$ tokens $R_t\in\R^{36\times 512}$), over which every screen patch attends to form a spatial map $m^{\mathrm{mem}}_t$ (\Cref{fig:architecture}, violet skip).

\subsection{Compute}
\label{sec:method:vlm}

We bootstrap symbolic execution capabilities from a pretrained vision--language model.
Learning exact arithmetic purely from pixels is prohibitively slow, and hand-crafted state graphs~\cite{venturoli2026statecraftaugmentingvideoworld} do not scale across open-ended applications or general world models.
However, pretrained VLMs already have symbolic reasoning capabilities. Given this, we asked whether a VLM backbone could guide our interface diffusion module directly.

The problem is that a VLM cannot be plugged into a diffusion model out of the box. Attending over prompt-prefix states $[\x_t; \tilde{R}_t; a_t]$ ($\tilde{R}_t$: $R_t$ projected into the VLM's embedding space) without explicit symbolic supervision ($\Lexe=0$), yields $0.0\%$ arithmetic accuracy (Section~S4.3). Another alternative, conditioning on decoded text $y_t$, made each state-generation step of the outer world model very slow, as we had to wait for slow test-time autoregressive decoding from the VLM.

To get explicit text supervision during training without paying the test-time decoding penalty, we build on recent methods that connect autoregressive VLMs to diffusion decoders using learnable query tokens~\cite{pan2025metaqueries,chen2025blip3o,panteam2025pan} (Figure~S2 in Section~S1). We insert $N_{\mathrm{out}}=16$ learnable query tokens $\Qout\in\R^{16\times 2048}$ into the input sequence of a Qwen-VL model with a frozen vision tower and LoRA-adapted language backbone (\texttt{Qwen3-VL-2B-Instruct}~\cite{bai2025qwen3vl} / \texttt{Qwen2.5-VL-3B-Instruct}~\cite{bai2025qwen25vl,hu2021lora}) strictly between the multimodal prefix and the training target $y_t$, the text of the action's outcome (e.g.\ the printed answer), so that the sequence is ordered as
\begin{equation}
  \underbrace{\big[\x_t;\; \tilde{R}_t;\; a_t\big]}_{\text{multimodal prefix}} \;\longrightarrow\; \underbrace{\Qout}_{\text{causal queries}} \;\longrightarrow\; \underbrace{y_t}_{\text{train target}}.
  \label{eq:vlm_seq}
\end{equation}
Under the causal mask, $\Qout$ precedes $y_t$, so its output states $\Hout\in\R^{16\times 2048}$ cannot see the answer ($\partial \Hout / \partial y_t = 0$) and are computed in one forward pass at test time. During training, $y_t$ attends backward over $\Qout$, so next-token cross-entropy ($\Lexe$) and a linear probe ($\Lprobe$) force the VLM to retrieve cleared variables from $\tilde{R}_t$, execute the operation, and store the answer in $\Hout$.

\subsection{Interface}
\label{sec:method:renderer}
\label{sec:method:objectives}


Once \textbf{Memory} has retrieved the off-screen slots $R_t$ and \textbf{Compute} has produced the outcome states $\Hout$, the \textbf{Interface} module renders $\hat{\x}_{t+1}$ at interactive speeds by solving two sub-problems: updating only changing regions without blurring static typography, and routing non-spatial memory and VLM tokens, to 2D screen coordinates.
Rather than training low-level cursor movement and desktop-interface rendering from scratch, we build directly on the open-source release of NeuralOS~\cite{rivard2025neuralos}, whose authors generously released well-documented training code and weights pretrained for many GPU-hours across millions of desktop transitions. We initialize our short-term action LSTM ($m^{\mathrm{lstm}}_t$) and base UNet channels from their pretrained backbone to inherit strong local action-to-pixel priors.

\paragraph{Routing memory and compute into 2D space.}
Whereas the UNet operates on a 2D latent grid ($96{\times}128$), $R_t\in\R^{36\times 512}$ and $\Hout\in\R^{16\times 2048}$ are 1D token sequences specifying \emph{what} file or arithmetic result should appear, not \emph{where} on the screen to draw it.
We bridge this gap via two pathways (\Cref{fig:architecture}, Module~3, with the full conditioning diagram in Figure~S3 of Section~S1).
First, to localize where actions occur, 2D maps carrying screen coordinates are stacked at the UNet input for 2D convolutions. These are the noised residual $\Delta\z^{(k)}_t$ (the latent change the UNet predicts, described below), short-term LSTM map $m^{\mathrm{lstm}}_t$, $(y,x)$ cursor heat map, spatial memory readout $m^{\mathrm{mem}}_t$, and noise-corrupted reference $\tilde{\z}_t$ (a noised copy of $\z_t$).
Second, to let any screen region look up non-local content from $R_t$ and $\Hout$, we project both into $\R^{512}$ to form a $52$-token semantic bank $\Csem = \mathrm{LN}([W_R R_t;\, W_H \Hout] + \alpha_{\mathrm{pos}} P_{1\mathrm{D}}) \in \R^{52\times 512}$, where 1D sinusoidal encodings $P_{1\mathrm{D}}$ preserve left-to-right character and digit order (e.g.\ \texttt{842} vs.\ \texttt{248}).
At each UNet stage $\ell \in \{1,\dots,7\}$, every 2D patch in $U^{(\ell)}$ acts as a \textbf{query} (with 2D coordinates $P_{2\mathrm{D}}$) attending over $\Csem$ as \textbf{keys and values}, so that each stage is updated as
\begin{equation}
  \begin{aligned}
  \tilde{U}^{(\ell)} = U^{(\ell)} + \tanh(g_\ell)\,\mathrm{Proj}_\ell\!\Big(
  &\mathrm{MHA}_\ell\!\big(\mathrm{GN}(U^{(\ell)}) + P_{2\mathrm{D}}, \\[-3pt]
  &\Csem,\; \Csem\big)\Big),
  \end{aligned}
  \label{eq:cross_attn}
\end{equation}
retrieving the content relevant to its $(y,x)$ region and adding it into $U^{(\ell)}$ via a learnable gate $\tanh(g_\ell)$ ($g_0 = 0.25$).

\begin{table*}[t]
  \centering
  \vspace{-12pt}
  \small
  \setlength{\tabcolsep}{3.6pt}
  \begin{tabular}{@{}lcccc@{\hspace{12pt}}ccc@{}}
    \toprule
    & \multicolumn{2}{c}{Desktop} & \multicolumn{2}{c}{Active CLI} & Text OCR & $261$-step & Render Speed \\
    \cmidrule(lr){2-3} \cmidrule(lr){4-5}
    \textbf{(A) $512{\times}384$} & LPIPS $\downarrow$ & PSNR $\uparrow$ & LPIPS $\downarrow$ & PSNR $\uparrow$ & (\%) $\uparrow$ & PSNR $\uparrow$ & (fps) $\uparrow$ \\
    \midrule
    \textsc{NeuralOS} (recurrent LSTM)~\cite{rivard2025neuralos} & $0.115 \pm 0.002$ & $25.51$ & $0.128 \pm 0.009$ & $23.82$ & $0.0$ & $20.64$ & $\mathbf{20.5}$ \\
    \textsc{Window} ($8$-frame diffusion) & $0.111 \pm 0.002$ & $25.70$ & $0.094 \pm 0.002$ & $25.23$ & $1.3$ & $20.69$ & $19.8$ \\
    \textbf{Ours} & $\mathbf{0.107 \pm 0.002}$ & $\mathbf{25.80}$ & $\mathbf{0.086 \pm 0.003}$ & $\mathbf{25.75}$ & $\mathbf{50.1 \pm 2.0}$ & $\mathbf{22.09}$ & $18.2$ \\
    \midrule
    & & & Whole & Active-edit & Text OCR & $16$-step & Speed \\
    \textbf{(B) $1024{\times}768$} & LPIPS $\downarrow$ & FID $\downarrow$ & PSNR $\uparrow$ & gain (dB) $\uparrow$ & (\%) $\uparrow$ & PSNR $\uparrow$ & (fps) $\uparrow$ \\
    \midrule
    \textsc{Copy-Only} (no Memory/Compute) & $0.051 \pm 0.001$ & $66.1$ & $29.50 \pm 0.23$ & $+2.73 \pm 0.21$ & $83.4 \pm 6.4$ & $17.38$ & $\mathbf{21.0}$ \\
    \textbf{Ours} & $\mathbf{0.044 \pm 0.004}$ & $\mathbf{56.1}$ & $\mathbf{31.22 \pm 0.18}$ & $\mathbf{+5.20 \pm 0.11}$ & $\mathbf{88.7 \pm 4.6}$ & $\mathbf{22.34}$ & $10.5$ \\
    \bottomrule
  \end{tabular}
  \vspace{-6pt}
  \caption{\textbf{Visual fidelity, text legibility, and speed.}
  (A)~Controlled $512{\times}384$ suite (\texttt{SD KL-VAE}).
  (B)~$1024{\times}768$ suite (\texttt{FLUX.1-schnell} 16-ch VAE) against \textsc{Copy-Only}, our renderer without Memory or Compute.
  Mean $\pm$ s.d.\ over $3$ seeds. Active-edit gain is changed-pixel PSNR minus that of copying the previous frame. NeuralOS is fine-tuned from the public release on our data, following their exact training configuration. Zero-shot, it ignores our screens and renders its own desktop. 
}
  \label{tab:main}
  \label{tab:visual}
  \vspace{0pt}
\end{table*}

\paragraph{Residual latent prediction.}
Standard 4-channel full-frame diffusion tends to blur dense features such as 10--12\,px glyphs, and also wastes capacity regenerating the ${\sim}95\%$ of pixels that stay static per step.
At both resolutions, we train an $x_0$-parameterized Residual UNet~\cite{ho2020denoising,yang2023diffusion,yue2023resshift,salehi2026rfdm} to predict only the standardized frame-to-frame delta $\Delta\z_t = (\z_{t+1} - \z_t)/s_c$ ($s_c$ per-channel training std) and add it to the skip $\z_t$ to form $\hat{\z}_{t+1} = \z_t + s_c \odot \hat{\Delta}\z_t$ (\Cref{fig:architecture}, green skip), converging in $2$--$4$ DDIM steps~\cite{song2021denoising} (\Cref{sec:exp:setup}; Figure~S4 in Section~S1).

\Cref{sec:experiments} uses NeuralOS's native 4-channel $512{\times}384$ SD KL-VAE for compute-matched baseline comparisons, and pairs the same Memory and Compute modules (via parameter-free $2{\times}2$ latent pooling) with the frozen 16-channel $1024{\times}768$ \texttt{FLUX.1-schnell} VAE~\cite{blackforestlabs2024flux} to overcome the 4-channel codec's $56.3\%$ text-legibility ceiling.

\subsection{End-to-End Training}
\label{sec:method:training}

Because most of the screen is static, an unweighted loss can predict $\hat{\Delta}\z_t \approx 0$. To combat this, we define our rendering loss as $\Lren = \frac{\sum_p \omega_p \|\hat{\Delta}\z_{t,p} - \Delta\z_{t,p}\|_2^2}{\sum_p \omega_p} + \frac{\lambda_{\mathrm{static}}}{|\mathcal{S}_t|C}\sum_{p\in\mathcal{S}_t}\|\hat{\Delta}\z_{t,p}\|_1$ (Equation~(S3) in Section~S1). This has the effect of upweighting active patches via $\omega_p = 1 + 4\cdot\mathbf{1}[\max_c|\Delta\z_{t,p,c}|>\delta_{\mathrm{chg}}]$ alongside a static-patch $L_1$ quietness penalty over $\mathcal{S}_t$. We also corrupt two inputs during training, adding schedule-dependent noise to $\tilde{\z}_t$~\cite{zhao2024identifying} and masking $\x_t$ ($p_{\mathrm{drop}}=0.5$, Section~S4.4), so that the model cannot simply copy the current frame and has to rely on $\tilde{R}_t$ and $\Hout$.
All three modules train jointly over $128$-step trajectories (a $64$-frame warm-up without gradients, followed by $64$ frames of truncated backpropagation through time (BPTT)~\cite{werbos1990bptt,williams1990efficient}) under $\mathcal{L}_{\mathrm{total}} = \Lren + \lambda_{\mathrm{JEPA}}\Lpred + \lambda_{\mathrm{exe}}\Lexe + \lambda_{\mathrm{probe}}\Lprobe$, backpropagating outer losses along one trunk (\Cref{fig:architecture}, dashed red) into the memory heads and surprise gate (\Cref{sec:exp:ablations} and Section~S4).

\section{Training Corpus and Data Mixture}
\label{sec:data}

\begin{figure*}[t]
  \centering
  \vspace{-2pt}
  \includegraphics[width=0.95\linewidth,trim=0 99pt 0 0,clip]{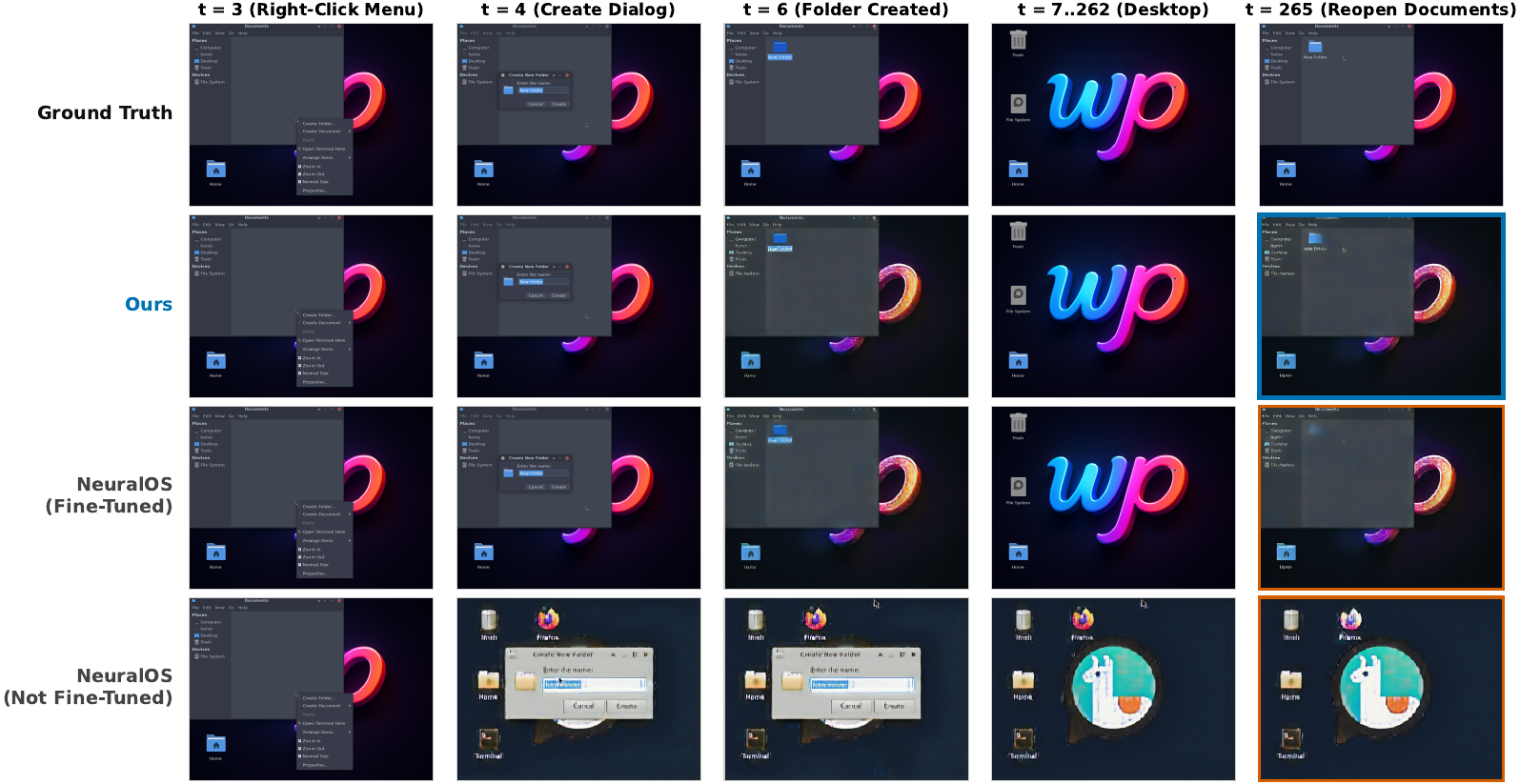}
  \vspace{0pt}
  \caption{\textbf{Long-horizon $512{\times}384$ persistence on the XFCE File Manager task.} The user creates \texttt{New Folder} in \texttt{Documents} ($t{=}3..6$), closes the File Manager for $256$ steps, and reopens \texttt{Documents} at $t{=}265$ (col.~5). \textbf{Ours} preserves the \texttt{New Folder} icon, whereas fine-tuned \textbf{\textsc{NeuralOS}} draws an empty pane. Note how the two models agree while the folder is on screen and diverge only once it has been hidden for 256 steps.}
  \label{fig:rollout_xfce_g256_r36_512}
  \vspace{-2pt}
\end{figure*}

Training a world model to function as a usable computer requires training examples of both (a) open-world GUI diversity (cursor dynamics, window management, and varied typography), and (b) controlled multi-step workflows where off-screen files or cleared variables uniquely determine future frames.
Public GUI corpora alone rarely revisit occluded state or test symbolic execution after screen clears. Due to this, we chose to curate five separate datasets, which we group into three regimes (Figures~S5--S7 and Table~S2 in Section~S3):
\begin{enumerate}[leftmargin=1.3em,itemsep=0.5pt,topsep=1.5pt]
  \item \textbf{Three open-world GUI corpora (AgentNet, ProCUA-SFT, and CUA-Gym, with $>2.1$\,TB raw, $173{,}975$ trajectories / $8.83\,\text{M}$ steps):} Human recordings on Windows, macOS, and Ubuntu from (1)~AgentNet~\cite{wang2025opencua} ($21{,}957$ cleaned episodes), plus Ubuntu rollouts from (2)~ProCUA-SFT~\cite{jung2026procua} ($92{,}964$ Kimi-K2.5 episodes) and (3)~CUA-Gym ($55{,}139$ Qwen3.5 VM episodes).
  \item \textbf{Controlled XFCE file persistence (\emph{Desktop}, source 4):} Hand-collected VS Code sessions ($768$ train, $96$ eval episodes in $32$ matched triplets) where a geometric diagram (triangle, circle, or star) is edited, occluded for $13$--$256$ steps while competing files are edited in the same pane, and reopened from pixel-identical frames.
  \item \textbf{Controlled XFCE cleared-screen CLI (\emph{Maths}, source 5):} Interactive Python REPL sessions ($4{,}000$ train, $300$ eval episodes, four task families) where variables (\texttt{q = 854}) are cleared via \texttt{Ctrl+L} before evaluating queries (\texttt{q * 738}).
\end{enumerate}

\vspace{-3pt}
\paragraph{Synthesis of sources.}
All $16{:}9$ recordings are letterboxed onto a $4{:}3$ canvas, and \texttt{pyautogui} scripts are parsed into abstract syntax trees without being executed, which lets us filter $194$ corrupted gap/scroll records and fix an undocumented post-action frame offset in ProCUA-SFT so every step obeys $(\x_t, a_t) \to \x_{t+1}$ (Section~S3).
Each step emits a verbatim keystroke string (preserving character order and \texttt{Ctrl+L}) for \textbf{Compute}, plus $(y,x)$ cursor heat maps, click flags, and held-key vectors for \textbf{Memory} and \textbf{Interface}.

\vspace{-3pt}
\paragraph{Data mixtures, training windows, and public release.}
Three of our five sources are already public. Upon publication, we will release the full aligned corpus, including \emph{Desktop} and \emph{Maths}, along with our alignment pipelines, mixture manifests, training code, and model weights.

To fully evaluate all of our architectural mechanisms under academic compute budgets, our multi-seed ablations and long-horizon rollouts train on $128$-step windows ($64$-frame warm-up + $64$-frame BPTT) for $3{,}000\text{--}9{,}000$ steps ($48{,}000\text{--}144{,}000$ $128$-step trajectories) on the \textbf{Dense Interactive Mixture} ($50\%$ \emph{Desktop}, $50\%$ \emph{Maths}), complemented by a high-resolution $1024{\times}768$ transfer run on the \textbf{Broad Mixture} ($75.0\%$ equally stratified AgentNet, ProCUA-SFT, and CUA-Gym, $12.5\%$ \emph{Desktop}, and $12.5\%$ \emph{Maths}, comprising $12{,}268$ \texttt{FLUX.1-schnell} episodes / $325{,}573$ transitions resampled to $694{,}266$ windows (for more details see Table~S2 and Sections~S4, S6, and~S7 in the Supplementary Material)).

\section{Experiments and Results}
\label{sec:experiments}
\renewcommand{\dbltopfraction}{0.95}
\renewcommand{\topfraction}{0.95}
\renewcommand{\bottomfraction}{0.90}
\renewcommand{\textfraction}{0.04}
\renewcommand{\floatpagefraction}{0.95}
\renewcommand{\dblfloatpagefraction}{0.95}
\setcounter{dbltopnumber}{3}
\setcounter{topnumber}{3}


We ask three questions of each model. Is it a competent renderer (\Cref{sec:exp:setup})? Does it remember off-screen state (\Cref{sec:exp:permanence})? And is the slot memory what drives that recall (\Cref{sec:exp:mechanistic})?

We use two evaluation settings (Section~S1): a \textbf{Controlled $512{\times}384$ Suite} (\texttt{SD KL-VAE} + \texttt{Qwen2.5-VL-3B}) that matches \textsc{NeuralOS}~\cite{rivard2025neuralos} in codec and resolution for fair $3$-seed comparisons, and a \textbf{High-Resolution $1024{\times}768$ Suite} (\texttt{FLUX.1-schnell} 16-ch VAE + \texttt{Qwen3-VL-2B}). We do not compare models across resolutions.

\begin{table*}[t]
  \centering
  \vspace{-8pt}
  \small
  \setlength{\tabcolsep}{5pt}
  \begin{tabular}{@{}lccccc@{}}
    \toprule
    & Edited-file & \multicolumn{3}{c}{Same-pane overwrite for $\Delta t$ steps (\%) $\uparrow$} & Donor memory \\
    \cmidrule(lr){3-5}
    Model ($512{\times}384$) & recall (\%) $\uparrow$ & $\Delta t{=}32$ & $\Delta t{=}64$ & $\Delta t{=}128$ & swap (\%) $\uparrow$ \\
    \midrule
    \textsc{NeuralOS} (recurrent LSTM)~\cite{rivard2025neuralos} & $56.0$ & $36.8$ & $36.5$ & $38.2$ & -- \\
    \textsc{Window} ($8$-frame diffusion) & $49.6$ & $37.8$ & $34.0$ & $39.9$ & -- \\
    \textsc{Coordinate Slots} ($(y,x)$-addressed bank) & $51.9$ & $38.9$ & $39.6$ & $37.5$ & -- \\
    \textbf{Ours} (content-addressed slots) & $\mathbf{64.2}$ & $\mathbf{46.2}$ & $\mathbf{50.0}$ & $\mathbf{42.4}$ & $\mathbf{64.7 \pm 1.1}$ \\
    \bottomrule
  \end{tabular}
  \vspace{-6pt}
  \caption{\textbf{Off-screen memory: 3-way hidden-shape accuracy ($33.3\%$ chance).}
  \emph{Edited-file recall} reopens an edited file after occlusion ($3$ seeds).
  In \emph{Overwrite}, a competing file is first edited in the same pane for $\Delta t$ steps ($288$ evaluations each).
  \emph{Donor swap} counts trials that render the donor's shape after a memory transplant.
  Bold marks the best result in each column.}
  \label{tab:functional}
  \label{tab:ablations}
  \vspace{-8pt}
\end{table*}

\subsection{Visual Fidelity}
\label{sec:exp:setup}
\label{sec:exp:rendering}
\label{sec:exp:quality}
\label{sec:exp:hires}

\Cref{tab:visual} compares \textbf{Ours} against \textbf{\textsc{NeuralOS}}~\citep{rivard2025neuralos}, \textbf{\textsc{Window}} ($8$-frame sliding-window diffusion~\citep{alonso2024diamond,valevski2025diffusion}), for the $512{\times}384$ setting. At $1024{\times}768$, we compare with \textbf{\textsc{Copy-Only}}, our residual renderer without Memory or Compute, trained on the same mixture.

At $512{\times}384$ the three models are close on desktop LPIPS ($0.107$ vs.\ $0.111$ and $0.115$), and \textbf{Ours} pulls ahead where content changes or must be kept: active CLI transitions ($0.086$ vs.\ $0.094$ and $0.128$ LPIPS) and persistent terminal text ($50.1\%$ vs.\ $1.3\%$ and $0.0\%$ OCR).

At $1024{\times}768$, the comparison with \textsc{Copy-Only} isolates what Memory and Compute add to an otherwise identical renderer. Adding them improves every fidelity metric, most clearly $16$-step rollout PSNR ($22.34$ vs.\ $17.38$\,dB) and active-edit gain ($+5.20$ vs.\ $+2.73$\,dB).

\subsection{Our Learned Memory Module Survives Occlusion and Overwriting}
\label{sec:exp:permanence}
\label{sec:exp:horizon}
\label{sec:exp:rollout}
\label{sec:exp:rollouts}

We test object permanence directly with \emph{hidden-shape recall}: across $32$ matched VS Code triplets ($96$ episodes $\times\,3$ seeds $= 288$ evaluations), a file containing a triangle, circle, or star is closed, an agent navigates away from the closed file (and carries out distracting tasks on screen), before eventually navigating back to the file.
At that moment nothing on screen says which shape to draw, so the answer can only come from memory. A recall counts as correct when the rendered frame is closer in $L_1$ to the true shape than to either foil ($33.3\%$ chance, with margin $\Delta L_1 = L_1^{\mathrm{foil}} - L_1^{\mathrm{true}}$).

\begin{figure}[t]
\centering
\vspace{-4pt}
\includegraphics[width=\linewidth]{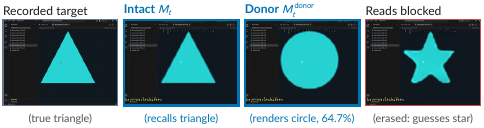}
\vspace{-8pt}
\caption{\textbf{Test-time memory transplant and erasure.}
With $\x_t$, $a_t$, and the LSTM fixed, keeping $M_t$ recalls the true triangle, transplanting $M_t^{\mathrm{donor}}$ renders the donor's circle ($64.7\%$), and blocking reads erases the file, leaving the model to guess (here, a star). Since the screen, click, and LSTM state are identical across panels, the change in shape can only come from the slot memory.}
\label{fig:mechanistic_interventions}
\vspace{-10pt}
\end{figure}



The results of this test can be seen under `Edited-file recall' in Column 1 of \Cref{tab:functional}. When an edited file is reopened after occlusion, \textbf{Ours} recalls the correct shape in $64.2\%$ of episodes, against $49.6$--$56.0\%$ for the baselines.

A harder test is \emph{overwriting}: before the file is reopened, a competing file (with another shape, or text) is edited in the same VS Code pane for $\Delta t$ steps, so the content to be remembered and the distractor occupy identical pixels.
\textsc{Window} and \textsc{NeuralOS} now sit close to chance ($34.0$--$39.9\%$).
\textsc{Coordinate Slots}, the same slot bank addressed by $(y,x)$ position rather than by content, fares no better ($37.5$--$39.6\%$), because two files in one pane can share one address.
\textbf{Ours} holds $50.0\%$ at $\Delta t{=}64$ ($\Delta L_1 = +0.070$, $95\%$ CI $[+0.046, +0.095]$).
The lead narrows at $\Delta t{=}128$ ($42.4\%$) (as all models eventually tend toward chance over long enough time horizons with fixed memory buffers).

\Cref{fig:rollout_xfce_g256_r36_512} shows the effect visually on \textsc{NeuralOS}'s native XFCE file-manager task. A folder hidden for $256$ steps is still there when \texttt{Documents} is reopened by \textbf{Ours}, where fine-tuned \textsc{NeuralOS} draws an empty pane.

\subsection{Memory is Causal and Editable}
\label{sec:exp:mechanistic}
\label{sec:exp:failure}

High recall does not by itself show that the learned memory module is responsible. For example, abstract concepts such as shapes could have been carried by the LSTM, or by the renderer.
However, because our memory is an explicit bank $M_t$, rather than part of an entangled latent state, we can test this directly. We hold the screen $\x_t$, the action $a_t$, and the LSTM (hidden-state) fixed, and change only $M_t$ (\Cref{fig:mechanistic_interventions}).
With $M_t$ intact, the model recalls the true shape (a triangle). When we transplant the memory with a donor $M_t \leftarrow M^{\mathrm{donor}}_t$ from an episode in which a circle was drawn, the model renders the donor's circle on $64.7 \pm 1.1\%$ of trials (\Cref{tab:functional}), and when we block memory reads, the stored file is erased.
What the model draws on reopening a file is therefore determined by the contents of its slots, which a supervisor can inspect, replace, or potentially even wipe without resetting the rest of the latent state of our model.

\paragraph{Ablations.}
\label{sec:exp:ablations}
In ablations at $3{,}000$ steps (Section~S4), removing \textsc{Memory-JEPA} ($\Lpred{=}0$) drops edited-file recall to chance ($32.1\%$), and removing residual prediction drops terminal OCR from $43.5\%$ to $0.2\%$ ($p{=}0.0007$). For full ablations of every component of our model, see the Supplementary Material.

\section{Conclusion}
\label{sec:conclusion}
\label{sec:limitations}

We introduce a world model of a computer, with a novel \textit{surprise-gated} memory component. Our model improves on the baselines in edited-region fidelity, text legibility, and resistance to overwriting. It learns how to interact with a modular and detachable memory state, and is significantly more functional (useful) than the learned alternatives. 

Due to academic compute limits, we trained on only a small portion of our full data mixture, and performance has not yet saturated, so scaling up training is a clear next step. 

Another exciting angle for future work is to explore open-ended interface generation, where applications which augment human capabilities are generated in real time around a user depending on their goals, abilities, and environment.

{
    \small
    \bibliographystyle{ieeenat_fullname}
    \bibliography{refs}

\begin{thebibliography}{76}
\providecommand{\natexlab}[1]{#1}
\providecommand{\url}[1]{\texttt{#1}}
\expandafter\ifx\csname urlstyle\endcsname\relax
  \providecommand{\doi}[1]{doi: #1}\else
  \providecommand{\doi}{doi: \begingroup \urlstyle{rm}\Url}\fi

\bibitem[Alaluf et~al.(2026)Alaluf, Avrahami, Leshem, Geyer, Goldberg,
  Richardson, et~al.]{alaluf2026solaris}
Yuval Alaluf, Omri Avrahami, Guy~Bukchin Leshem, Michal Geyer, Kfir Goldberg,
  Elad Richardson, et~al.
\newblock Solaris: Towards interfaces that are generated, not coded, 2026.

\bibitem[Alonso et~al.(2024)Alonso, Jelley, Micheli, Kanervisto, Storkey,
  Pearce, and Fleuret]{alonso2024diamond}
Eloi Alonso, Adam Jelley, Vincent Micheli, Anssi Kanervisto, Amos Storkey, Tim
  Pearce, and Fran{\c{c}}ois Fleuret.
\newblock Diffusion for world modeling: Visual details matter in atari.
\newblock In \emph{Advances in Neural Information Processing Systems
  (NeurIPS)}, 2024.

\bibitem[{Anthropic}(2024)]{anthropic2024computeruse}
{Anthropic}.
\newblock Developing a computer use model.
\newblock \url{https://www.anthropic.com/news/3-5-models-and-computer-use},
  2024.

\bibitem[Assran et~al.(2023)Assran, Duval, Misra, Bojanowski, Vincent, Rabbat,
  LeCun, and Ballas]{assran2023ijepa}
Mahmoud Assran, Quentin Duval, Ishan Misra, Piotr Bojanowski, Pascal Vincent,
  Michael Rabbat, Yann LeCun, and Nicolas Ballas.
\newblock Self-supervised learning from images with a joint-embedding
  predictive architecture.
\newblock In \emph{IEEE/CVF Conference on Computer Vision and Pattern
  Recognition (CVPR)}, 2023.

\bibitem[Assran et~al.(2025)Assran, Bardes, Fan, et~al.]{assran2025vjepa2}
Mahmoud Assran, Adrien Bardes, David Fan, et~al.
\newblock {V-JEPA 2}: Self-supervised video models enable understanding,
  prediction and planning.
\newblock \emph{arXiv preprint arXiv:2506.09985}, 2025.

\bibitem[Bai et~al.(2025{\natexlab{a}})Bai, Cai, Chen, Chen, Chen, Cheng, Deng,
  Ding, Gao, Ge, et~al.]{bai2025qwen3vl}
Shuai Bai, Yuxuan Cai, Ruizhe Chen, Keqin Chen, Xionghui Chen, Zesen Cheng,
  Lianghao Deng, Wei Ding, Chang Gao, Chunjiang Ge, et~al.
\newblock {Qwen3-VL} technical report, 2025{\natexlab{a}}.

\bibitem[Bai et~al.(2025{\natexlab{b}})Bai, Chen, Liu, Wang, Ge, Song, Dang,
  Wang, Wang, Tang, et~al.]{bai2025qwen25vl}
Shuai Bai, Keqin Chen, Xuejing Liu, Jialin Wang, Wenbin Ge, Sibo Song, Kai
  Dang, Peng Wang, Shijie Wang, Jun Tang, et~al.
\newblock {Qwen2.5-VL} technical report, 2025{\natexlab{b}}.

\bibitem[Bardes et~al.(2022)Bardes, Ponce, and LeCun]{bardes2022vicreg}
Adrien Bardes, Jean Ponce, and Yann LeCun.
\newblock {VICReg}: Variance-invariance-covariance regularization for
  self-supervised learning.
\newblock In \emph{International Conference on Learning Representations
  (ICLR)}, 2022.

\bibitem[Bardes et~al.(2024)Bardes, Garrido, Ponce, Chen, Rabbat, LeCun,
  Assran, and Ballas]{bardes2024vjepa}
Adrien Bardes, Quentin Garrido, Jean Ponce, Xinlei Chen, Michael Rabbat, Yann
  LeCun, Mahmoud Assran, and Nicolas Ballas.
\newblock Revisiting feature prediction for learning visual representations
  from video.
\newblock \emph{Transactions on Machine Learning Research (TMLR)}, 2024.

\bibitem[Behrouz et~al.(2025)Behrouz, Zhong, and Mirrokni]{behrouz2025titans}
Ali Behrouz, Peilin Zhong, and Vahab Mirrokni.
\newblock Titans: Learning to memorize at test time.
\newblock In \emph{Advances in Neural Information Processing Systems
  (NeurIPS)}, 2025.

\bibitem[{Black Forest Labs}(2024)]{blackforestlabs2024flux}
{Black Forest Labs}.
\newblock {FLUX}.
\newblock \url{https://github.com/black-forest-labs/flux}, 2024.

\bibitem[Bruce et~al.(2024)Bruce, Dennis, Edwards, Parker-Holder, Shi, Hughes,
  Lai, Mavalankar, Steigerwald, Apps, Aytar, Bechtle, Behbahani, Chan, Heess,
  Gonzalez, Osindero, Ozair, Reed, Zhang, Zolna, Clune, de~Freitas, Singh, and
  Rockt{\"a}schel]{bruce2024genie}
Jake Bruce, Michael Dennis, Ashley Edwards, Jack Parker-Holder, Yuge Shi,
  Edward Hughes, Matthew Lai, Aditi Mavalankar, Richie Steigerwald, Chris Apps,
  Yusuf Aytar, Sarah Bechtle, Feryal Behbahani, Stephanie Chan, Nicolas Heess,
  Lucy Gonzalez, Simon Osindero, Sherjil Ozair, Scott Reed, Jingwei Zhang,
  Konrad Zolna, Jeff Clune, Nando de Freitas, Satinder Singh, and Tim
  Rockt{\"a}schel.
\newblock Genie: Generative interactive environments.
\newblock In \emph{International Conference on Machine Learning (ICML)}, 2024.

\bibitem[Cetin et~al.(2026)Cetin, Erbach, Bozic, Go, Zhao, Kalischek, Tombari,
  Schindler, and Narnhofer]{cetin2026smart}
Eren Cetin, Julius Erbach, Vukasin Bozic, Hyojun Go, Zixiang Zhao, Nikolai
  Kalischek, Federico Tombari, Konrad Schindler, and Dominik Narnhofer.
\newblock Long-horizon video generation with slot memory addressed by region
  tokens, 2026.

\bibitem[Chen et~al.(2025)Chen, Xu, Pan, Hu, Qin, Goldstein, Huang, Zhou, Xie,
  Savarese, Xue, Xiong, and Xu]{chen2025blip3o}
Jiuhai Chen, Zhiyang Xu, Xichen Pan, Yushi Hu, Can Qin, Tom Goldstein, Lifu
  Huang, Tianyi Zhou, Saining Xie, Silvio Savarese, Le Xue, Caiming Xiong, and
  Ran Xu.
\newblock {BLIP3-o}: A family of fully open unified multimodal models ---
  architecture, training and dataset, 2025.

\bibitem[Chevalier-Boisvert et~al.(2023)Chevalier-Boisvert, Dai, Towers,
  de~Lazcano, Willems, Lahlou, Pal, Castro, and
  Terry]{chevalierboisvert2023minigrid}
Maxime Chevalier-Boisvert, Bolun Dai, Mark Towers, Rodrigo de Lazcano, Lucas
  Willems, Salem Lahlou, Suman Pal, Pablo~Samuel Castro, and Jordan Terry.
\newblock {Minigrid} \& {Miniworld}: Modular \& customizable reinforcement
  learning environments for goal-oriented tasks.
\newblock In \emph{Advances in Neural Information Processing Systems (NeurIPS),
  Datasets and Benchmarks Track}, 2023.

\bibitem[Deng et~al.(2023)Deng, Gu, Zheng, Chen, Stevens, Wang, Sun, and
  Su]{deng2023mind2web}
Xiang Deng, Yu Gu, Boyuan Zheng, Shijie Chen, Samuel Stevens, Boshi Wang, Huan
  Sun, and Yu Su.
\newblock {Mind2Web}: Towards a generalist agent for the web.
\newblock In \emph{Advances in Neural Information Processing Systems (NeurIPS)
  Datasets and Benchmarks Track}, 2023.

\bibitem[Friston(2010)]{friston2010free}
Karl Friston.
\newblock The free-energy principle: a unified brain theory?
\newblock \emph{Nature Reviews Neuroscience}, 11\penalty0 (2):\penalty0
  127--138, 2010.

\bibitem[Graves et~al.(2014)Graves, Wayne, and Danihelka]{graves2014ntm}
Alex Graves, Greg Wayne, and Ivo Danihelka.
\newblock Neural turing machines.
\newblock \emph{arXiv preprint arXiv:1410.5401}, 2014.

\bibitem[Graves et~al.(2016)Graves, Wayne, Reynolds, Harley, Danihelka,
  Grabska-Barwi{\'n}ska, Colmenarejo, Grefenstette, Ramalho, Agapiou,
  et~al.]{graves2016dnc}
Alex Graves, Greg Wayne, Malcolm Reynolds, Tim Harley, Ivo Danihelka, Agnieszka
  Grabska-Barwi{\'n}ska, Sergio~G{\'o}mez Colmenarejo, Edward Grefenstette,
  Tiago Ramalho, John Agapiou, et~al.
\newblock Hybrid computing using a neural network with dynamic external memory.
\newblock \emph{Nature}, 538\penalty0 (7626):\penalty0 471--476, 2016.

\bibitem[Greenblatt et~al.(2024)Greenblatt, Denison, Wright, Roger,
  et~al.]{greenblatt2024alignmentfaking}
Ryan Greenblatt, Carson Denison, Benjamin Wright, Fabien Roger, et~al.
\newblock Alignment faking in large language models, 2024.

\bibitem[Grill et~al.(2020)Grill, Strub, Altch{\'e}, Tallec, Richemond,
  Buchatskaya, Doersch, Avila~Pires, Guo, Gheshlaghi~Azar, Piot, Kavukcuoglu,
  Munos, and Valko]{grill2020byol}
Jean-Bastien Grill, Florian Strub, Florent Altch{\'e}, Corentin Tallec, Pierre
  Richemond, Elena Buchatskaya, Carl Doersch, Bernardo Avila~Pires, Zhaohan
  Guo, Mohammad Gheshlaghi~Azar, Bilal Piot, Koray Kavukcuoglu, R{\'e}mi Munos,
  and Michal Valko.
\newblock Bootstrap your own latent: A new approach to self-supervised
  learning.
\newblock In \emph{Advances in Neural Information Processing Systems
  (NeurIPS)}, pages 21271--21284, 2020.

\bibitem[Gu and Dao(2024)]{gu2024mamba}
Albert Gu and Tri Dao.
\newblock Mamba: Linear-time sequence modeling with selective state spaces.
\newblock In \emph{Conference on Language Modeling (COLM)}, 2024.

\bibitem[Guan et~al.(2026)Guan, Yu, Zhang, Wang, Zhang, Li, Qiao, Qin, Huang,
  Yang, et~al.]{guan2026cuwm}
Yiming Guan, Rui Yu, John Zhang, Lu Wang, Chaoyun Zhang, Liqun Li, Bo Qiao, Si
  Qin, He Huang, Fangkai Yang, et~al.
\newblock Computer-using world model.
\newblock In \emph{International Conference on Learning Representations
  (ICLR)}, 2026.

\bibitem[Ha and Schmidhuber(2018)]{ha2018worldmodels}
David Ha and J{\"u}rgen Schmidhuber.
\newblock Recurrent world models facilitate policy evolution.
\newblock In \emph{Advances in Neural Information Processing Systems
  (NeurIPS)}, 2018.

\bibitem[Hafner et~al.(2023)Hafner, Pasukonis, Ba, and
  Lillicrap]{hafner2023dreamerv3}
Danijar Hafner, Jurgis Pasukonis, Jimmy Ba, and Timothy Lillicrap.
\newblock Mastering diverse domains through world models.
\newblock \emph{arXiv preprint arXiv:2301.04104}, 2023.

\bibitem[Heusel et~al.(2017)Heusel, Ramsauer, Unterthiner, Nessler, and
  Hochreiter]{heusel2017gans}
Martin Heusel, Hubert Ramsauer, Thomas Unterthiner, Bernhard Nessler, and Sepp
  Hochreiter.
\newblock {GANs} trained by a two time-scale update rule converge to a local
  {Nash} equilibrium.
\newblock In \emph{Advances in Neural Information Processing Systems (NIPS)},
  2017.

\bibitem[Ho et~al.(2020)Ho, Jain, and Abbeel]{ho2020denoising}
Jonathan Ho, Ajay Jain, and Pieter Abbeel.
\newblock Denoising diffusion probabilistic models.
\newblock In \emph{Advances in Neural Information Processing Systems
  (NeurIPS)}, 2020.

\bibitem[Hu et~al.(2023)Hu, Russell, Yeo, Murez, Fedoseev, Kendall, Shotton,
  and Corrado]{hu2023gaia1}
Anthony Hu, Lloyd Russell, Hudson Yeo, Zak Murez, George Fedoseev, Alex
  Kendall, Jamie Shotton, and Gianluca Corrado.
\newblock {GAIA-1}: A generative world model for autonomous driving.
\newblock \emph{arXiv preprint arXiv:2309.17080}, 2023.

\bibitem[Hu et~al.(2022)Hu, Shen, Wallis, Allen-Zhu, Li, Wang, Wang, and
  Chen]{hu2021lora}
Edward~J. Hu, Yelong Shen, Phillip Wallis, Zeyuan Allen-Zhu, Yuanzhi Li, Shean
  Wang, Lu Wang, and Weizhu Chen.
\newblock {LoRA}: Low-rank adaptation of large language models.
\newblock In \emph{International Conference on Learning Representations
  (ICLR)}, 2022.

\bibitem[Hubinger et~al.(2024)Hubinger, Denison, Mu, Lambert,
  et~al.]{hubinger2024sleeper}
Evan Hubinger, Carson Denison, Jesse Mu, Mike Lambert, et~al.
\newblock Sleeper agents: Training deceptive {LLMs} that persist through safety
  training, 2024.

\bibitem[Hughes et~al.(2024)Hughes, Dennis, Parker-Holder, Behbahani,
  Mavalankar, Shi, Schaul, and Rockt{\"a}schel]{hughes2024oeasi}
Edward Hughes, Michael Dennis, Jack Parker-Holder, Feryal Behbahani, Aditi
  Mavalankar, Yuge Shi, Tom Schaul, and Tim Rockt{\"a}schel.
\newblock Open-endedness is essential for artificial superhuman intelligence.
\newblock \emph{arXiv preprint arXiv:2406.04268}, 2024.

\bibitem[Jung et~al.(2026)Jung, Lu, Cui, Khalifa, Zhang, Zhang, Xu, Deshmukh,
  Sapra, Tao, Choi, Kautz, Liu, and Dong]{jung2026procua}
Jaehun Jung, Ximing Lu, Brandon Cui, Muhammad Khalifa, Shaokun Zhang, Hao
  Zhang, Jin Xu, Amala~Sanjay Deshmukh, Karan Sapra, Andrew Tao, Yejin Choi,
  Jan Kautz, Mingjie Liu, and Yi Dong.
\newblock {ProCUA-SFT} technical report, 2026.

\bibitem[Katharopoulos et~al.(2020)Katharopoulos, Vyas, Pappas, and
  Fleuret]{katharopoulos2020transformers}
Angelos Katharopoulos, Apoorv Vyas, Nikolaos Pappas, and Fran{\c{c}}ois
  Fleuret.
\newblock Transformers are {RNNs}: Fast autoregressive transformers with linear
  attention.
\newblock In \emph{Proceedings of the 37th International Conference on Machine
  Learning (ICML)}, pages 5156--5165, 2020.

\bibitem[Lee et~al.(2023)Lee, Joshi, Turc, Hu, Liu, Eisenschlos, Khandelwal,
  Shaw, Chang, and Toutanova]{lee2023pix2struct}
Kenton Lee, Mandar Joshi, Iulia Turc, Hexiang Hu, Fangyu Liu, Julian
  Eisenschlos, Urvashi Khandelwal, Peter Shaw, Ming-Wei Chang, and Kristina
  Toutanova.
\newblock Pix2struct: Screenshot parsing as pretraining for visual language
  understanding.
\newblock In \emph{International Conference on Machine Learning (ICML)}, 2023.

\bibitem[Li et~al.(2025)Li, Torr, Vedaldi, and Jakab]{li2025vmem}
Runjia Li, Philip Torr, Andrea Vedaldi, and Tomas Jakab.
\newblock {VMem}: Consistent interactive video scene generation with
  surfel-indexed view memory.
\newblock In \emph{Proceedings of the IEEE/CVF International Conference on
  Computer Vision (ICCV)}, 2025.

\bibitem[Locatello et~al.(2020)Locatello, Weissenborn, Unterthiner, Mahendran,
  Heigold, Uszkoreit, Dosovitskiy, and Kipf]{locatello2020object}
Francesco Locatello, Dirk Weissenborn, Thomas Unterthiner, Aravindh Mahendran,
  Georg Heigold, Jakob Uszkoreit, Alexey Dosovitskiy, and Thomas Kipf.
\newblock Object-centric learning with slot attention.
\newblock In \emph{Advances in Neural Information Processing Systems
  (NeurIPS)}, 2020.

\bibitem[Meinke et~al.(2024)Meinke, Schoen, Scheurer, Balesni, Shah, and
  Hobbhahn]{meinke2024scheming}
Alexander Meinke, Bronson Schoen, J{\'e}r{\'e}my Scheurer, Mikita Balesni,
  Rusheb Shah, and Marius Hobbhahn.
\newblock Frontier models are capable of in-context scheming, 2024.

\bibitem[Pan et~al.(2025)Pan, Shukla, Singh, Zhao, Mishra, Wang, Xu, Chen, Li,
  Juefei-Xu, Hou, and Xie]{pan2025metaqueries}
Xichen Pan, Satya~Narayan Shukla, Aashu Singh, Zhuokai Zhao, Shlok~Kumar
  Mishra, Jialiang Wang, Zhiyang Xu, Jiuhai Chen, Kunpeng Li, Felix Juefei-Xu,
  Ji Hou, and Saining Xie.
\newblock Transfer between modalities with {MetaQueries}, 2025.

\bibitem[{PAN Team} et~al.(2025){PAN Team}, Liu, Gu, Deng, Liu, Feng, Gao, Hu,
  Huang, Yang, Zhou, Xiang, Hu, Liu, and Xing]{panteam2025pan}
{PAN Team}, Zihan Liu, Yi Gu, Mingkai Deng, Guangyi Liu, Zeyu Feng, Qiyue Gao,
  Yiyan Hu, Benhao Huang, Yichi Yang, Kun Zhou, Jiannan Xiang, Zhiting Hu,
  Zhengzhong Liu, and Eric~P. Xing.
\newblock {PAN}: A world model for general, actionable, and long-horizon world
  simulation, 2025.

\bibitem[Parker-Holder and Fruchter(2025)]{parkerholder2025genie3}
Jack Parker-Holder and Shlomi Fruchter.
\newblock Genie 3: A new frontier for world models.
\newblock Google DeepMind blog, 2025.

\bibitem[Parker-Holder et~al.(2024)Parker-Holder, Ball, Bruce, Dasagi,
  Holsheimer, Kaplanis, Moufarek, Scully, Shar, Shi,
  et~al.]{parkerholder2024genie2}
Jack Parker-Holder, Philip Ball, Jake Bruce, Vibhavari Dasagi, Kristian
  Holsheimer, Christos Kaplanis, Alexandre Moufarek, Guy Scully, Jeremy Shar,
  Jimmy Shi, et~al.
\newblock Genie 2: A large-scale foundation world model.
\newblock Google DeepMind blog, 2024.

\bibitem[Po et~al.(2025)Po, Nitzan, Zhang, Chen, Dao, Shechtman, Wetzstein, and
  Huang]{po2025longcontext}
Ryan Po, Yotam Nitzan, Richard Zhang, Berlin Chen, Tri Dao, Eli Shechtman,
  Gordon Wetzstein, and Xun Huang.
\newblock Long-context state-space video world models.
\newblock In \emph{Proceedings of the IEEE/CVF International Conference on
  Computer Vision (ICCV)}, 2025.

\bibitem[Rao and Ballard(1999)]{rao1999predictive}
Rajesh P.~N. Rao and Dana~H. Ballard.
\newblock Predictive coding in the visual cortex: a functional interpretation
  of some extra-classical receptive-field effects.
\newblock \emph{Nature Neuroscience}, 2\penalty0 (1):\penalty0 79--87, 1999.

\bibitem[Rawles et~al.(2023)Rawles, Li, Rodriguez, Riva, and
  Lillicrap]{rawles2023aitw}
Christopher Rawles, Alice Li, Daniel Rodriguez, Oriana Riva, and Timothy
  Lillicrap.
\newblock {AndroidInTheWild}: A large-scale dataset for android device control.
\newblock In \emph{Advances in Neural Information Processing Systems (NeurIPS)
  Datasets and Benchmarks Track}, 2023.

\bibitem[Rawles et~al.(2024)Rawles, Clinckemaillie, Chang, Waltz, Lau, Fair,
  Li, Bishop, Li, Campbell-Ajala, Riva, Lillicrap, Toyama, and
  Riva]{rawles2024androidworld}
Christopher Rawles, Sarah Clinckemaillie, Yifan Chang, Jonathan Waltz,
  Gabrielle Lau, Marybeth Fair, Alice Li, William Bishop, Wei Li, Folawiyo
  Campbell-Ajala, Daniel Riva, Timothy Lillicrap, Kris Toyama, and Oriana Riva.
\newblock {AndroidWorld}: A dynamic benchmarking environment for autonomous
  agents.
\newblock \emph{arXiv preprint arXiv:2405.14573}, 2024.

\bibitem[Rivard et~al.(2026)Rivard, Sun, Guo, Chen, and
  Deng]{rivard2025neuralos}
Luke Rivard, Sun Sun, Hongyu Guo, Wenhu Chen, and Yuntian Deng.
\newblock {NeuralOS}: Towards simulating operating systems via neural
  generative models.
\newblock In \emph{International Conference on Learning Representations
  (ICLR)}, 2026.

\bibitem[Salehi et~al.(2026)Salehi, Noroozi, Morreale, Chavhan, Chadwick,
  Ramos, and Mehrotra]{salehi2026rfdm}
Mohammadreza Salehi, Mehdi Noroozi, Luca Morreale, Ruchika Chavhan, Malcolm
  Chadwick, Alberto~Gil Ramos, and Abhinav Mehrotra.
\newblock {RFDM}: Residual flow diffusion model for efficient causal video
  editing.
\newblock In \emph{Proceedings of the IEEE/CVF Conference on Computer Vision
  and Pattern Recognition (CVPR)}, 2026.

\bibitem[Schlag et~al.(2021)Schlag, Irie, and Schmidhuber]{schlag2021linear}
Imanol Schlag, Kazuki Irie, and J{\"u}rgen Schmidhuber.
\newblock Linear transformers are secretly fast weight programmers.
\newblock In \emph{Proceedings of the 38th International Conference on Machine
  Learning (ICML)}, pages 9355--9366, 2021.

\bibitem[Shaw et~al.(2023)Shaw, Joshi, Cohan, Berant, Pasupat, Hu, Khandelwal,
  Lee, and Toutanova]{shaw2023pix2act}
Peter Shaw, Mandar Joshi, James Cohan, Jonathan Berant, Panupong Pasupat,
  Hexiang Hu, Urvashi Khandelwal, Kenton Lee, and Kristina Toutanova.
\newblock From pixels to {UI} actions: Learning to follow instructions via
  graphical user interfaces.
\newblock In \emph{Advances in Neural Information Processing Systems
  (NeurIPS)}, 2023.

\bibitem[Song et~al.(2021)Song, Meng, and Ermon]{song2021denoising}
Jiaming Song, Chenlin Meng, and Stefano Ermon.
\newblock Denoising diffusion implicit models.
\newblock In \emph{International Conference on Learning Representations
  (ICLR)}, 2021.

\bibitem[Sun et~al.(2025)Sun, Li, Dalal, Xu, Vikram, Zhang, Dubois, Chen, Wang,
  Koyejo, Hashimoto, and Guestrin]{sun2025learning}
Yu Sun, Xinhao Li, Karan Dalal, Jiarui Xu, Arjun Vikram, Genghan Zhang, Yann
  Dubois, Xinlei Chen, Xiaolong Wang, Sanmi Koyejo, Tatsunori Hashimoto, and
  Carlos Guestrin.
\newblock Learning to (learn at test time): {RNNs} with expressive hidden
  states.
\newblock In \emph{Proceedings of the 42nd International Conference on Machine
  Learning (ICML)}, pages 57503--57522, 2025.

\bibitem[Toyama et~al.(2021)Toyama, Hamel, Gergely, Comanici, Glaese, Ahmed,
  Jackson, Mourad, and Precup]{toyama2021androidenv}
Daniel Toyama, Philippe Hamel, Anita Gergely, Gheorghe Comanici, Amelia Glaese,
  Zafarali Ahmed, Tyler Jackson, Shibl Mourad, and Doina Precup.
\newblock {AndroidEnv}: A reinforcement learning platform for android.
\newblock \emph{arXiv preprint arXiv:2105.13231}, 2021.

\bibitem[Valevski et~al.(2025)Valevski, Leviathan, Arar, and
  Fruchter]{valevski2025diffusion}
Dani Valevski, Yaniv Leviathan, Moab Arar, and Shlomi Fruchter.
\newblock Diffusion models are real-time game engines.
\newblock In \emph{International Conference on Learning Representations
  (ICLR)}, 2025.

\bibitem[van~den Oord et~al.(2018)van~den Oord, Li, and Vinyals]{oord2018cpc}
A{\"a}ron van~den Oord, Yazhe Li, and Oriol Vinyals.
\newblock Representation learning with contrastive predictive coding, 2018.

\bibitem[Venturoli et~al.(2026)Venturoli, Savov, Wang, Pollefeys, and
  Gool]{venturoli2026statecraftaugmentingvideoworld}
Lorenzo Venturoli, Nedko Savov, Xi Wang, Marc Pollefeys, and Luc~Van Gool.
\newblock Statecraft: Augmenting video world models with abstract-state
  reasoning, 2026.

\bibitem[Wang et~al.(2025)Wang, Wang, Lu, Yang, Xie, Wang, Deng, Guo, Xu, Wu,
  et~al.]{wang2025opencua}
Xinyuan Wang, Bowen Wang, Dunjie Lu, Junlin Yang, Tianbao Xie, Junli Wang,
  Jiaqi Deng, Xiaole Guo, Yiheng Xu, Chen~Henry Wu, et~al.
\newblock {OpenCUA}: Open foundations for computer-use agents, 2025.

\bibitem[Werbos(1990)]{werbos1990bptt}
Paul~J. Werbos.
\newblock Backpropagation through time: What it does and how to do it.
\newblock \emph{Proceedings of the IEEE}, 78\penalty0 (10):\penalty0
  1550--1560, 1990.

\bibitem[Williams and Peng(1990)]{williams1990efficient}
Ronald~J. Williams and Jing Peng.
\newblock An efficient gradient-based algorithm for on-line training of
  recurrent network trajectories.
\newblock \emph{Neural Computation}, 2\penalty0 (4):\penalty0 490--501, 1990.

\bibitem[Wu et~al.(2026)Wu, Li, Yang, Guo, Yang, Zhang, Zhang, Hu, Rymkowski,
  Sridhar, Yan, and Huang]{wu2026worldt3}
Junyi Wu, Zekun Li, Chenhongyi Yang, Chen Guo, Lingchen Yang, Yongan Zhang,
  Xinyi Zhang, Hongbing Hu, Bartek Rymkowski, Srinath Sridhar, Yan Yan, and
  Xijie Huang.
\newblock {World-T3}: Test-time training as persistent memory for real-time
  video world models, 2026.

\bibitem[Xiao et~al.(2025)Xiao, Lan, Zhou, Ouyang, Yang, Zeng, and
  Pan]{xiao2025worldmem}
Zeqi Xiao, Yushi Lan, Yifan Zhou, Wenqi Ouyang, Shuai Yang, Yanhong Zeng, and
  Xingang Pan.
\newblock {WorldMem}: Long-term consistent world simulation with memory.
\newblock In \emph{Advances in Neural Information Processing Systems
  (NeurIPS)}, 2025.

\bibitem[Xie et~al.(2024)Xie, Zhang, Chen, Li, Zhao, Cao, Hua, Cheng, Shin,
  Lei, Liu, Xu, Zhou, Savarese, Xiong, Zhong, and Yu]{xie2024osworld}
Tianbao Xie, Danyang Zhang, Jixuan Chen, Xiaochuan Li, Siheng Zhao, Ruisheng
  Cao, Toh~Jing Hua, Zhoujun Cheng, Dongchan Shin, Fangyu Lei, Yitao Liu,
  Yiheng Xu, Shuyan Zhou, Silvio Savarese, Caiming Xiong, Victor Zhong, and Tao
  Yu.
\newblock {OSWorld}: Benchmarking multimodal agents for open-ended tasks in
  real computer environments.
\newblock In \emph{Advances in Neural Information Processing Systems (NeurIPS)
  Datasets and Benchmarks Track}, 2024.

\bibitem[Yang et~al.(2023)Yang, Srivastava, and Mandt]{yang2023diffusion}
Ruihan Yang, Prakhar Srivastava, and Stephan Mandt.
\newblock Diffusion probabilistic modeling for video generation.
\newblock \emph{Entropy}, 25\penalty0 (10):\penalty0 1469, 2023.

\bibitem[Yang et~al.(2024)Yang, Wang, Shen, Panda, and Kim]{yang2024gated}
Songlin Yang, Bailin Wang, Yikang Shen, Rameswar Panda, and Yoon Kim.
\newblock Gated linear attention transformers with hardware-efficient training.
\newblock In \emph{Proceedings of the 41st International Conference on Machine
  Learning (ICML)}, pages 56501--56523, 2024.

\bibitem[Yang et~al.(2025{\natexlab{a}})Yang, Kautz, and
  Hatamizadeh]{yang2025gated}
Songlin Yang, Jan Kautz, and Ali Hatamizadeh.
\newblock Gated delta networks: Improving {Mamba2} with delta rule.
\newblock In \emph{International Conference on Learning Representations
  (ICLR)}, 2025{\natexlab{a}}.

\bibitem[Yang et~al.(2025{\natexlab{b}})Yang, Yang, Huang, Brown, Yang, Yu,
  Tong, Zheng, Xu, Wang, Fergus, LeCun, Fei-Fei, and Xie]{yang2025cambrians}
Shusheng Yang, Jihan Yang, Pinzhi Huang, Ellis Brown, Zihao Yang, Yue Yu,
  Shengbang Tong, Zihan Zheng, Yifan Xu, Muhan Wang, Rob Fergus, Yann LeCun, Li
  Fei-Fei, and Saining Xie.
\newblock {Cambrian-S}: Towards spatial supersensing in video,
  2025{\natexlab{b}}.

\bibitem[Yu et~al.(2025)Yu, Bai, Qin, Liu, Wang, Wan, Zhang, and
  Liu]{yu2025context}
Jiwen Yu, Jianhong Bai, Yiran Qin, Quande Liu, Xintao Wang, Pengfei Wan, Di
  Zhang, and Xihui Liu.
\newblock Context as memory: Scene-consistent interactive long video generation
  with memory retrieval.
\newblock In \emph{SIGGRAPH Asia 2025 Conference Papers}, 2025.

\bibitem[Yu et~al.(2026)Yu, Choi, Park, Shin, and Shim]{yu2026latentgui}
Seungjun Yu, Hojun Choi, Seojeong Park, Jaeyo Shin, and Hyunjung Shim.
\newblock Latent world action model for {GUI} agents, 2026.

\bibitem[Yue et~al.(2023)Yue, Wang, and Loy]{yue2023resshift}
Zongsheng Yue, Jianyi Wang, and Chen~Change Loy.
\newblock {ResShift}: Efficient diffusion model for image super-resolution by
  residual shifting.
\newblock In \emph{Advances in Neural Information Processing Systems
  (NeurIPS)}, 2023.

\bibitem[Zhang et~al.(2025)Zhang, Cai, Li, Wetzstein, and
  Agrawala]{zhang2025framepack}
Lvmin Zhang, Shengqu Cai, Muyang Li, Gordon Wetzstein, and Maneesh Agrawala.
\newblock Frame context packing and drift prevention in next-frame-prediction
  video diffusion models.
\newblock In \emph{Advances in Neural Information Processing Systems
  (NeurIPS)}, 2025.

\bibitem[Zhang et~al.(2018)Zhang, Isola, Efros, Shechtman, and
  Wang]{zhang2018unreasonable}
Richard Zhang, Phillip Isola, Alexei~A. Efros, Eli Shechtman, and Oliver Wang.
\newblock The unreasonable effectiveness of deep features as a perceptual
  metric.
\newblock In \emph{Proceedings of the IEEE Conference on Computer Vision and
  Pattern Recognition (CVPR)}, 2018.

\bibitem[Zhang et~al.(2026)Zhang, Xu, Tao, Wang, She, and
  Boularias]{zhang2026rlawm}
Xinyu Zhang, Zhengtong Xu, Yutian Tao, Yeping Wang, Yu She, and Abdeslam
  Boularias.
\newblock Learning visual feature-based world models via residual latent
  action.
\newblock In \emph{Advances in Neural Information Processing Systems
  (NeurIPS)}, 2026.

\bibitem[Zhang et~al.(2024)Zhang, Yang, Zhu, Zhang, Cui, Wang, Wang, Shi, Wang,
  Bi, Zhou, and Fu]{zhang2024gated}
Yu Zhang, Songlin Yang, Ruijie Zhu, Yue Zhang, Leyang Cui, Yiqiao Wang, Bolun
  Wang, Freda Shi, Bailin Wang, Wei Bi, Peng Zhou, and Guohong Fu.
\newblock Gated slot attention for efficient linear-time sequence modeling.
\newblock In \emph{Advances in Neural Information Processing Systems
  (NeurIPS)}, 2024.

\bibitem[Zhao et~al.(2024)Zhao, Zhu, Xiang, Zheng, Li, and
  Zhu]{zhao2024identifying}
Min Zhao, Hongzhou Zhu, Chendong Xiang, Kaiwen Zheng, Chongxuan Li, and Jun
  Zhu.
\newblock Identifying and solving conditional image leakage in image-to-video
  diffusion model.
\newblock In \emph{Advances in Neural Information Processing Systems
  (NeurIPS)}, 2024.

\bibitem[Zhou et~al.(2025)Zhou, Pan, LeCun, and Pinto]{zhou2025dinowm}
Gaoyue Zhou, Hengkai Pan, Yann LeCun, and Lerrel Pinto.
\newblock {DINO-WM}: World models on pre-trained visual features enable
  zero-shot planning.
\newblock In \emph{Proceedings of the 42nd International Conference on Machine
  Learning (ICML)}, 2025.

\bibitem[Zhou et~al.(2024)Zhou, Xu, Zhu, Zhou, Lo, Sridhar, Cheng, Ou, Bisk,
  Fried, Alon, and Neubig]{zhou2024webarena}
Shuyan Zhou, Frank~F. Xu, Hao Zhu, Xuhui Zhou, Robert Lo, Abishek Sridhar,
  Xianyi Cheng, Tianyue Ou, Yonatan Bisk, Daniel Fried, Uri Alon, and Graham
  Neubig.
\newblock {WebArena}: A realistic web environment for building autonomous
  agents.
\newblock In \emph{International Conference on Learning Representations
  (ICLR)}, 2024.

\bibitem[Zhuge et~al.(2026)Zhuge, Schmidhuber, et~al.]{zhuge2026neural}
Mingchen Zhuge, J{\"u}rgen Schmidhuber, et~al.
\newblock Neural computers, 2026.

\end{thebibliography}
}

\clearpage
\raggedbottom
\renewcommand{\dbltopfraction}{0.92}
\renewcommand{\dblfloatpagefraction}{0.80}
\renewcommand{\topfraction}{0.92}
\renewcommand{\textfraction}{0.05}
\setcounter{dbltopnumber}{3}
\maketitlesupplementary
\setcounter{section}{0}\setcounter{figure}{0}\setcounter{table}{0}\setcounter{equation}{0}
\renewcommand{\thesection}{S\arabic{section}}\renewcommand{\theHsection}{S\arabic{section}}
\renewcommand{\thefigure}{S\arabic{figure}}\renewcommand{\theHfigure}{S\arabic{figure}}
\renewcommand{\thetable}{S\arabic{table}}\renewcommand{\theHtable}{S\arabic{table}}
\renewcommand{\theequation}{S\arabic{equation}}\renewcommand{\theHequation}{S\arabic{equation}}
\newcommand{\tfbanner}{{\sffamily\footnotesize\colorbox{black!12}{\strut\textbf{TEACHER-FORCED}}\enspace every frame is predicted from the \emph{real} frames before it}\par\vspace{3pt}}
\newcommand{\sfbanner}[1]{{\sffamily\footnotesize\colorbox{orange!30}{\strut\textbf{STUDENT-FORCED}}\enspace #1}\par\vspace{3pt}}

\section{Method Equations, Diagrams, and Evaluation Protocols}
\label{sec:suppl:impl}

The main paper asks whether a generative world model can \emph{be} a usable computer rather than merely render plausible frames.
In order to achieve this goal, we first needed to overcome three critical problems:
\begin{enumerate}
    \item Given the lack of 3D camera pose to rely on, world models of computers forget off-screen state, even over short time-horizons.
    \item World models \textit{in general} tend to struggle with explicit symbolic reasoning, but nowhere is this more apparent than in a world model of a computer. Baseline models often generate plausible-looking but patently illogical text on screen.
    \item As ${\sim}95\%$ of a desktop can remain static from one frame to the next (think of a terminal shell), diffusing the next screen state without weighting the rendering loss can cause the model to collapse into simply copying the previous frame.
\end{enumerate}

Our Interactive Neural Computer (Main Figure~2) addresses these three failures in turn by separating the system into three modules, trained end-to-end: \textbf{Memory}, \textbf{Compute}, and \textbf{Interface}. We give each module targeted supervision rather than asking pixel diffusion alone to teach all three.

This section `opens the hood' of the main paper's Figure~2, and provides the intuition, exact equations, and evaluation protocols behind the architecture introduced in the main paper's Sections~3 and~5.


We organize the details into four parts: why each training objective and addressing rule takes the form it does (\Cref{sec:suppl:losses}), how the architecture stays compact and scales to $1024{\times}768$ (\Cref{sec:suppl:arch}), how we test functional usefulness beyond single-step image metrics (\Cref{sec:suppl:eval_protocol}), and how we train long-horizon memory under bounded GPU memory (\Cref{sec:suppl:training}).

\begin{figure}[t]
\centering
\input{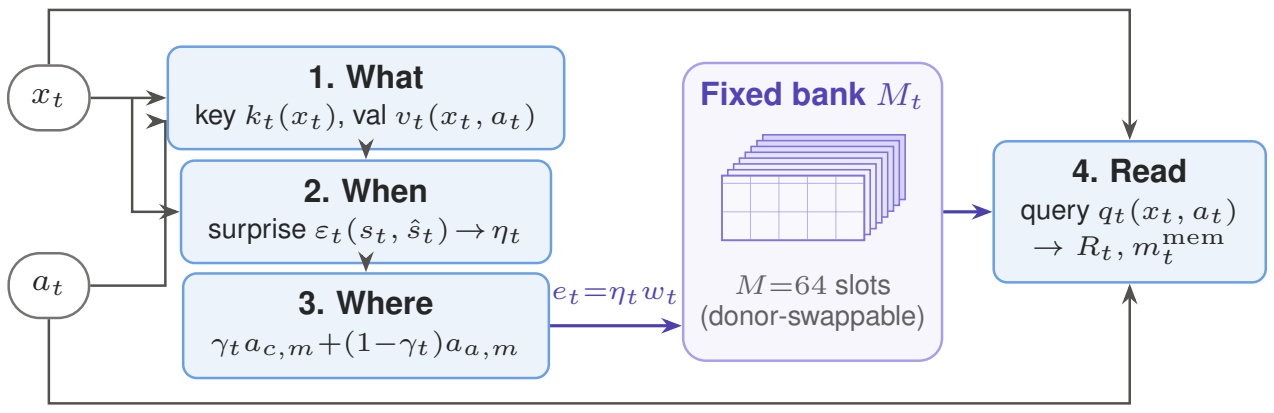}
\resizebox{\linewidth}{!}{%
\begin{tikzpicture}[archpic, x=1cm, y=1cm]
  \node[dat] (obs) at (-0.30, 1.30) {$\x_t$};
  \node[dat] (act) at (-0.30, -0.15) {$a_t$};

  \node[mod, minimum width=2.25cm, minimum height=0.62cm] (what) at (2.15, 1.30) {\textbf{1. What}\\{\archsmall key $k_t(\x_t)$, val $v_t(\x_t,a_t)$}};
  \node[mod, minimum width=2.25cm, minimum height=0.62cm] (when) at (2.15, 0.42) {\textbf{2. When}\\{\archsmall surprise $\varepsilon_t(s_t,\hat{s}_t)\!\to\!\eta_t$}};
  \node[mod, minimum width=2.25cm, minimum height=0.62cm] (where) at (2.15, -0.46) {\textbf{3. Where}\\{\archsmall $\gamma_t a_{c,m}{+}(1{-}\gamma_t)a_{a,m}$}};

  \node[memgrp, minimum width=2.00cm, minimum height=2.30cm, align=center] (bank) at (5.60, 0.42) {};
  \node[font=\sffamily\scriptsize\bfseries, text=violc!85!black] at (5.60, 1.32) {Fixed bank $M_t$};
  \pic at (4.90, 0.20) {slotsmed};
  \node[note] at (5.60, -0.28) {$M{=}64$ slots\\(donor-swappable)};

  \node[mod, minimum width=1.95cm, minimum height=0.88cm] (read) at (8.05, 0.42) {\textbf{4. Read}\\{\archsmall query $q_t(\x_t, a_t)$}\\{\archsmall $\to R_t, m^{\mathrm{mem}}_t$}};

  \draw[fwd] (obs.east) -- (what.west);
  \draw[fwd] (obs.east) ++(0.32,0) |- (when.west);
  \draw[fwd] (obs.north) -- ++(0,0.42) -| (read.north);
  \draw[fwd] (act.east) -- ++(0.58,0) |- ([yshift=-1.8mm]what.west);
  \draw[fwd] (what) -- (when);
  \draw[fwd] (when) -- (where);
  \draw[fwd, draw=violc!80!black, line width=0.7pt] (where.east) -- (bank.west |- where.east) node[midway, above=1pt, font=\sffamily\archsmall, text=violc!85!black] {$e_t{=}\eta_t w_t$};
  \draw[fwd, draw=violc!80!black, line width=0.7pt] (bank.east) -- (read.west);
  \draw[fwd] (act.south) -- ++(0,-0.65) -| (read.south);
\end{tikzpicture}}
\vspace{-5pt}
\caption{\textbf{Zoom-in on Module~1 (Memory) of Main Figure~2: concept-slot memory splits writing and reading into four learned decisions.} Inside the violet Memory box of Main Figure~2, step $t$ extracts keys and values from $(\x_t, a_t)$ (\emph{what}), gates writes by prediction surprise $\varepsilon_t$ via $\eta_t$ so static frames do not overwrite the bank (\emph{when}), blends content matching with free-slot allocation via $w_t$ to place updates in the $64$-slot bank $M_t$ (\emph{where}, \cref{eq:suppl_addr}), and issues action-conditioned queries to retrieve off-screen state into $R_t$ and $m^{\mathrm{mem}}_t$ (\emph{read}).}
\label{fig:mem_cycle}
\end{figure}

\subsection{Training Objectives and Slot Addressing}
\label{sec:suppl:losses}
Each component of our loss fixes a concrete failure mode we encountered when training this neural computer end to end.

\paragraph{Why can't a vanilla world model—or the outer rendering loss alone—learn long-horizon memory, and how does \textsc{Memory-JEPA} fix it?}
Suppose a user edits a file, closes it, and works elsewhere for roughly a hundred steps (several minutes) before clicking the tab of the original file to reopen it.
A vanilla visual world model has no explicit off-screen storage: it relies on a short sliding window of recent frames or a single recurrent hidden state, trained purely to reconstruct pixels.
Over a long gap, the closed file is crowded out of that finite context, and even when the file reappears, the handful of pixels that reveal its contents are drowned out by thousands of unchanged background pixels.

To remember abstract concepts across long horizons, we add a fixed-capacity slot memory—akin to a neural hard disk (\Cref{fig:mem_cycle})—with learned read and write heads that decide what to store and what to retrieve.
Training is still purely from unlabeled frame-and-action trajectories, without ground-truth slot labels.
However, if these read and write heads receive gradients \textit{only} from the outer rendering loss $\Lren$ backpropagated through the UNet, they suffer the same drowning-out problem.

Our \textsc{Memory-JEPA} architecture solves both failures---drowning out and crowding out---with one self-supervised predictor, trained directly on the memory controller in representation space.

First, instead of reconstructing hundreds of thousands of pixels through the UNet, the predictor must predict a single $512$-dimensional semantic vector summarizing the next screen (pooled from Qwen-VL vision features and the VAE latent) from only the current screen, the user's action, and what the read head retrieves from memory.
Trained contrastively against other screens in the batch (via $\mathcal{L}_{\mathrm{InfoNCE}}$), this loss asks which screen is about to appear rather than averaging over thousands of unchanged background pixels; and because neither the current screen nor the click reveals what was inside a closed file, the predictor can match the right target only if the write head filed the content earlier and the read head fetched it back.

Second, because the predictor produces a live expectation of the next screen at every step, it gives us a principled measure of \emph{prediction surprise} (Main Section~3.1; \Cref{fig:mem_cycle}, Step~2), inspired by predictive coding~\cite{rao1999predictive,friston2010free} and \textsc{Cambrian-S}~\cite{yang2025cambrians}.
We can thus gate every write we make to the slot bank by the magnitude of surprise: predictable, static frames keep the gate shut so repetitive desktop frames do not crowd out the bank, and only unpredicted transitions---such as editing or closing a file---open the gate to commit a write.

This JEPA-style architecture can be subject to two potential types of collapse, and we guard against both.
First, consider representation collapse: if the encoder and predictor simply map every screen to the same constant vector, prediction error drops to zero without learning anything about the desktop.
To close \textit{this} loophole, we employ an exponential moving average (EMA) copy of the encoder---with gradients stopped---to compute a \textit{stable target summary} $s^*_{t+1} \in \R^{512}$ from the true next frame $\x_{t+1}$. Then, we train the controller's prediction $\hat{s}_{t+1} \in \R^{512}$ to match $s^*_{t+1}$ both contrastively ($\mathcal{L}_{\mathrm{InfoNCE}}$) and directly (squared $L_2$ distance).
Second, memory-write collapse: even when screen summaries stay informative, the controller can file every screen under the same write key $k_t \in \mathbb{S}^{63}$ (a unit vector in $\R^{64}$), or the read head's first query $q^{(1)}_t \in \mathbb{S}^{63}$ can fail to discover which stored key matches the screen about to appear. We close this loophole by adding $\mathcal{L}_{\mathrm{VICReg}}$~\cite{bardes2022vicreg} to spread write keys across the unit sphere, and directly aligning $q^{(1)}_t$ with the EMA encoder's target key $k^*_{t+1} \in \mathbb{S}^{63}$ for the next frame---supervising the read address without asking gradients to survive across the occlusion.

However, one last shortcut still remains. In many computer-interface transitions, the predictor could simply \textit{ignore} the memory bank $M_t$ and rely on the current screen alone---it would do quite well in this regime, in many instances, given how static computer interfaces are from one timestep to the next.
To block this, we form a counterfactual prediction $\hat{s}^{\,\mathrm{swap}}_{t+1}$ by replacing $M_t$ with another episode's bank, and require the swapped prediction to be worse by a margin $\rho = 0.1$:
\begin{subequations}
\begin{equation}
  \mathcal{L}_{\mathrm{swap}} = \max\!\big(0,\; \rho + \|\hat{s}_{t+1} - s^*_{t+1}\|_2^2 - \|\mathrm{sg}(\hat{s}^{\,\mathrm{swap}}_{t+1}) - s^*_{t+1}\|_2^2\big),
  \label{eq:suppl_lswap}
\end{equation}
with all three vectors normalized, averaged over frames on which the screen changes and $k^*_{t+1}$ has cosine similarity above $0.7$ with the key of an occupied slot.

Putting all five components together---contrastive prediction, direct $L_2$ alignment, read-query supervision, key-spread regularization, and the counterfactual swap margin---gives the full \textsc{Memory-JEPA} loss:
\begin{equation}
  \begin{aligned}
  \Lpred =\;& \mathcal{L}_{\mathrm{InfoNCE}}\!\big(\normalize(\hat{s}_{t+1}),\, \normalize(s^*_{t+1})\big) \\
  &+ \lambda_{\mathrm{align}}\big\|\normalize(\hat{s}_{t+1}) - \normalize(s^*_{t+1})\big\|_2^2 \\
  &+ \lambda_q\big(1 - \langle q^{(1)}_t, k^*_{t+1}\rangle\big) \\
  &+ \lambda_{\mathrm{VICReg}}\,\mathcal{L}_{\mathrm{VICReg}}(k_t)
   + \lambda_{\mathrm{swap}}\,\mathcal{L}_{\mathrm{swap}}.
  \end{aligned}
  \label{eq:suppl_ljepa}
\end{equation}
\end{subequations}
Removing $\Lpred$ drops edited-file recall back to chance (\Cref{sec:suppl:fail_jepa}).

\paragraph{How does the memory decide whether to update an existing slot or claim a free one?}
Unlike 3D physical world models, a computer interface has no camera pose for memory addressing. Screen coordinates are not a viable replacement either, because a new application window routinely opens over the exact coordinates of a closed one---which would simply overwrite old files.
The third alternative, a fixed-length queue, eventually fills up and evicts long-lived state.

We confront these issues by following the Differentiable Neural Computer (DNC)~\cite{graves2016dnc}. A write chooses between two destinations (\Cref{fig:mem_cycle}, Step~3): it either updates the occupied slot whose stored content best matches the current screen, or claims the first empty slot.

Each slot $m \in \{1,\dots,M\}$ stores a key $K_{t-1,m} \in \R^{64}$ and an occupancy $u_{t-1,m} \in [0,1]$, both initialized to zero.
A gate $\gamma_t \in (0,1)$ (Main Section~3.1) measures how strongly the current write key $k_t$ matches the best occupied slot and blends a content distribution $a_{c,m}$ with an allocation distribution $a_{a,m}$ as $w_{t,m} = \gamma_t a_{c,m} + (1-\gamma_t) a_{a,m}$—updating an existing slot when re-editing a known file ($\gamma_t \approx 1$), and claiming a free slot for a new file ($\gamma_t \approx 0$).
For content matching, we score each slot by its similarity to $k_t$ and add a log-occupancy bias so empty slots receive negligible weight once any slot is occupied.
For allocation, rather than maintaining the DNC's discrete usage-sorted free list, we claim free slots from left to right differentiably: we favor low-occupancy slots and penalize every slot $m$ by how many earlier slots $j < m$ are still below the free threshold $u_{\mathrm{free}}$:
\begin{equation}
  \begin{aligned}
  a_{c,m} &= \softmax_m\!\Big(\frac{\langle K_{t-1,m}, k_t\rangle}{\tau_w} + \log(u_{t-1,m}{+}\epsilon)\Big), \\
  a_{a,m} &= \softmax_m\!\Big(\frac{-u_{t-1,m}}{\tau_a} \\[-2pt]
  &\qquad - \lambda_{\mathrm{prec}}\sum_{j=1}^{m-1}\sigma\!\Big(\frac{u_{\mathrm{free}} - u_{t-1,j}}{\tau_a}\Big)\Big).
  \end{aligned}
  \label{eq:suppl_addr}
\end{equation}
Here $\tau_w$ and $\tau_a$ are softmax temperatures, $\lambda_{\mathrm{prec}}$ weights the left-to-right precedence penalty, $\sigma$ is the logistic sigmoid, and $\epsilon{=}10^{-6}$ prevents $\log 0$.

\begin{figure}[t]
\centering
\input{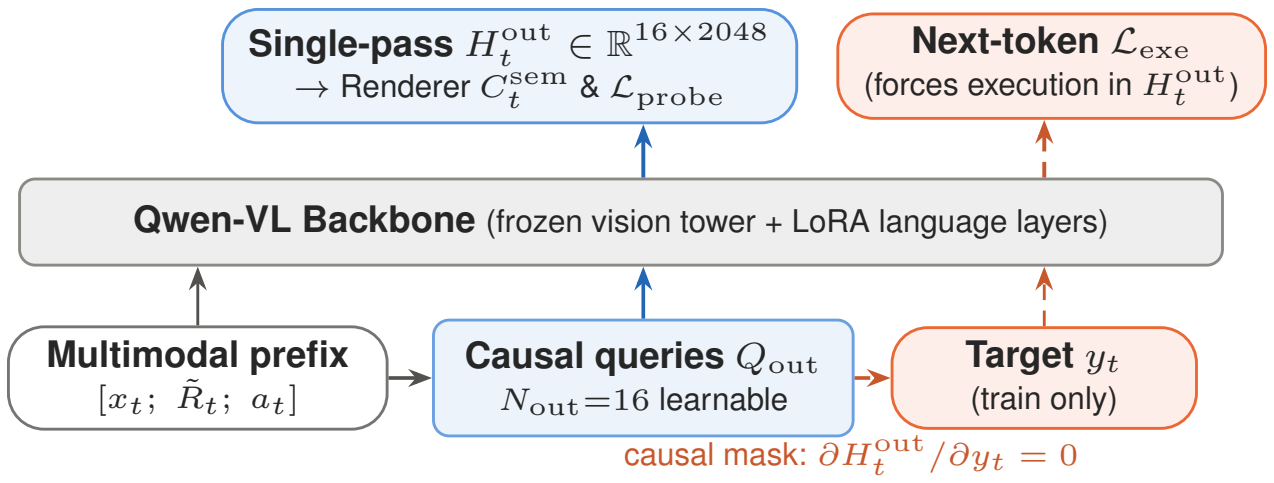}
\resizebox{\linewidth}{!}{%
\begin{tikzpicture}[archpic, x=1cm, y=1cm]

  \node[dat, minimum width=2.55cm, minimum height=0.65cm] (pref) at (1.35, 0.0)
    {\textbf{Multimodal prefix}\\{\archsmall $[\x_t;\; \tilde{R}_t;\; a_t]$}};

  \node[mod, minimum width=2.25cm, minimum height=0.65cm] (qout) at (4.35, 0.0)
    {\textbf{Causal queries $\Qout$}\\{\archsmall $N_{\mathrm{out}}{=}16$ learnable}};

  \node[gt, minimum width=2.05cm, minimum height=0.65cm] (yt) at (7.05, 0.0)
    {\textbf{Target $y_t$}\\{\archsmall (train only)}};

  \draw[fwd, line width=0.7pt] (pref.east) -- (qout.west);
  \draw[fwd, draw=orc!85!black, line width=0.7pt, densely dashed] (qout.east) -- (yt.west);

  \node[font=\sffamily\archsmall, text=orc!85!black] at (5.75, -0.52) {causal mask: $\partial \Hout / \partial y_t = 0$};

  \node[frz, minimum width=8.10cm, minimum height=0.58cm] (vlm) at (4.20, 1.05)
    {\textbf{Qwen-VL Backbone} {\archsmall (frozen vision tower + LoRA language layers)}};

  \draw[fwd] (pref.north) -- (pref.north |- vlm.south);
  \draw[fwd, draw=bluec!85!black, line width=0.7pt] (qout.north) -- (qout.north |- vlm.south);
  \draw[fwd, draw=orc!85!black, densely dashed] (yt.north) -- (yt.north |- vlm.south);

  \node[dat, draw=bluec!80, fill=bluec!8, minimum width=2.55cm, minimum height=0.62cm] (hout) at (3.45, 2.10)
    {\textbf{Single-pass $\Hout \in \R^{16\times 2048}$}\\{\archsmall $\to$ Renderer $\Csem$ \& $\Lprobe$}};

  \node[gt, minimum width=2.15cm, minimum height=0.62cm] (lexe) at (7.10, 2.10)
    {\textbf{Next-token $\Lexe$}\\{\archsmall (forces execution in $\Hout$)}};

  \draw[fwd, draw=bluec!85!black, line width=0.75pt] (qout.north |- vlm.north) -- (qout.north |- hout.south);
  \draw[fwd, draw=orc!85!black, line width=0.75pt, densely dashed] (yt.north |- vlm.north) -- (yt.north |- lexe.south);

\end{tikzpicture}}
\vspace{-5pt}
\caption{\textbf{Zoom-in on Module~2 (Compute) of Main Figure~2: post-prefix causal queries extract single-pass VLM execution without test-time decoding.} Inside the blue Compute box of Main Figure~2, $16$ learnable queries $\Qout$ sit strictly between the multimodal prefix $[\x_t;\, \tilde{R}_t;\, a_t]$ and the training-only outcome text $y_t$. Under causal masking, $\Qout$ reads the prefix without seeing $y_t$ ($\partial\Hout/\partial y_t = 0$), so $y_t$ is dropped at test time; during training, next-token loss $\Lexe$ over $y_t$ (dashed orange) forces the VLM to compress the executed answer into $\Hout$.}
\label{fig:vlm_bridge}
\vspace{-6pt}
\end{figure}

\paragraph{Why does standard diffusion loss fail on a mostly static desktop?}
Most desktop pixels do not move much between state transitions. Re-synthesizing the entire frame at every step wastes capacity redrawing static backgrounds, and tends to blur small interface text.
Predicting only the frame-to-frame change, then, seems like a good idea. It would avoid touching static text, by only predicting the delta in the state on each timestep. However, this format introduces another trap: an unweighted loss is dominated by the unchanged patches, tempting the renderer to predict zero change everywhere and freeze the screen.

We resolve this tension with an active-weighted residual loss in latent space (\Cref{fig:unetcond_suppl}).
Let $\Delta\z_t = (\z_{t+1}-\z_t)/s_c$ be the latent increment standardized by its per-channel scale $s_c$, and let $\hat{\Delta}\z_t$ be the Residual UNet's prediction.
We mark a latent patch $p$ as static if its largest change across the $C$ channels stays below a quiet threshold $\delta_{\mathrm{stat}}$, forming the static set $\mathcal{S}_t = \{p : \max_c |\Delta\z_{t,p,c}| < \delta_{\mathrm{stat}}\}$.
Conversely, patches whose change exceeds $\delta_{\mathrm{chg}}$ receive extra weight $\omega_p = 1 + \omega_{\mathrm{chg}}\cdot\mathbf{1}[\max_c |\Delta\z_{t,p,c}| > \delta_{\mathrm{chg}}]$.
The rendering loss then balances active edits against static quietness:
\begin{equation}
  \Lren = \frac{\sum_p \omega_p \|\hat{\Delta}\z_{t,p} - \Delta\z_{t,p}\|_2^2}{\sum_p \omega_p} + \frac{\lambda_{\mathrm{static}}}{|\mathcal{S}_t|\,C}\sum_{p\in\mathcal{S}_t}\|\hat{\Delta}\z_{t,p}\|_1.
  \label{eq:suppl_lren}
\end{equation}
The first term ensures changing regions cannot be drowned out by the static background, while the second applies an $L_1$ quietness penalty on static patches so unchanged regions do not accumulate drift across rollouts.
Replacing this residual target with full-frame prediction collapses terminal OCR (\Cref{sec:suppl:fail_renderer}); all loss weights and thresholds are listed in \Cref{tab:hyperparams}.

\usetikzlibrary{decorations.pathreplacing}
\providecommand{\LDelta}{\mathcal{L}_{\Delta}}
\providecommand{\archsmall}{\fontsize{6}{7}\selectfont}
\providecommand{\subt}[1]{\\[-1pt]{\archsmall\color{black!62}#1}}
\definecolor{bluec}{HTML}{2A78D6}
\definecolor{violc}{HTML}{6250D6}
\definecolor{orc}{HTML}{EB6834}
\definecolor{grayc}{HTML}{8A8984}
\definecolor{mutc}{HTML}{52514E}
\tikzset{
  archpic/.style={font=\sffamily\scriptsize, >={Stealth[length=1.6mm]}, line width=0.55pt, draw=mutc, text=black!85},
  mod/.style={draw=bluec!70, fill=bluec!9, rounded corners=3pt, align=center, inner xsep=6pt, inner ysep=4pt,
              minimum height=0.8cm},
  frz/.style={mod, draw=grayc, fill=grayc!14},
  dat/.style={draw=mutc!80, fill=white, rounded corners=2.4mm, inner xsep=5pt, inner ysep=2.5pt, minimum height=0.5cm,
              align=center},
  gt/.style={dat, draw=orc, fill=orc!11},
  loss/.style={circle, draw=orc, fill=white, line width=0.85pt, inner sep=0pt, minimum size=0.4cm,
               path picture={\draw[orc, line width=0.75pt] (path picture bounding box.north west) -- (path picture bounding box.south east)
                             (path picture bounding box.north east) -- (path picture bounding box.south west);}},
  lname/.style={text=orc!85!black, inner sep=1.5pt},
  fwd/.style={->, draw=mutc},
  grd/.style={->, draw=orc, densely dashed, line width=0.8pt},
  memgrp/.style={draw=violc!55, fill=violc!5, rounded corners=5pt},
  cmp/.style={font=\sffamily\scriptsize\bfseries, text=black!55, inner sep=1pt},
  note/.style={font=\sffamily\archsmall, text=black!62, inner sep=1pt, align=center},
  pics/slots/.style={code={
    \foreach \k in {7,...,0}{
      \draw[violc!80, fill=violc!\the\numexpr 4+\k*3\relax, line width=0.4pt] (\k*0.05,\k*0.05) rectangle ++(1.3,0.85);}
    \draw[step=0.26, violc!45, line width=0.25pt] (0,0) grid (1.3,0.85);
    \draw[violc!80, line width=0.4pt] (0,0) rectangle (1.3,0.85);}},
  pics/slotsmed/.style={code={
    \foreach \k in {7,...,0}{
      \draw[violc!80, fill=violc!\the\numexpr 4+\k*3\relax, line width=0.35pt] (\k*0.045,\k*0.045) rectangle ++(1.1,0.5);}
    \draw[step=0.22, violc!45, line width=0.2pt] (0,0) grid (1.1,0.5);
    \draw[violc!80, line width=0.35pt] (0,0) rectangle (1.1,0.5);}},
  pics/slotsmini/.style={code={
    \foreach \k in {3,...,0}{
      \draw[violc!80, fill=violc!\the\numexpr 6+\k*5\relax, line width=0.35pt] (\k*0.05,\k*0.05) rectangle ++(0.5,0.3);}
    \draw[step=0.1, violc!45, line width=0.2pt] (0,0) grid (0.5,0.3);}},
}

\definecolor{tealc}{HTML}{1B8A80}
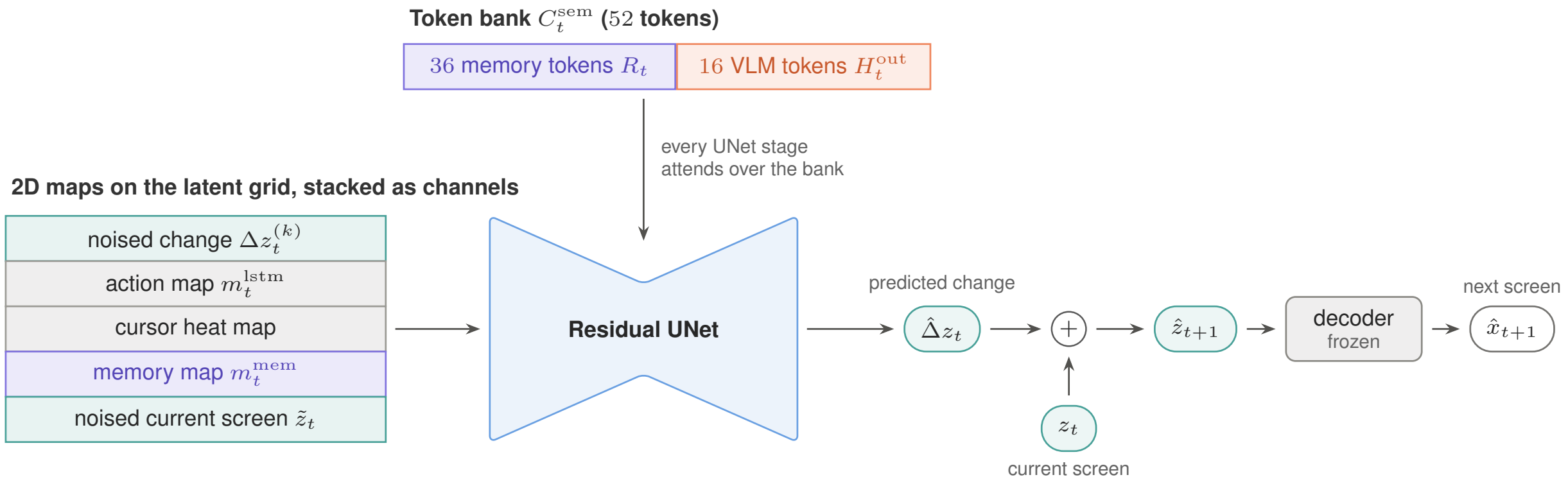
\begin{figure*}[t]
  \centering
  \resizebox{0.92\textwidth}{!}{%
  \begin{tikzpicture}[archpic,
      arr/.style={fwd, shorten >=3pt, shorten <=3pt, rounded corners=4pt},
      bar/.style={minimum width=4.2cm, minimum height=0.5cm, inner sep=0pt, line width=0.55pt, align=center},
      pill/.style={dat, minimum height=0.5cm},
      ttl/.style={font=\sffamily\scriptsize\bfseries, text=black!80, inner sep=1.5pt}]

    \node[ttl, anchor=south west] at (0,3.42) {2D maps on the latent grid, stacked as channels};
    \node[bar, draw=tealc!80, fill=tealc!10]  (b1) at (2.10,3.05) {noised change $\Delta\z^{(k)}_t$};
    \node[bar, draw=grayc!85, fill=grayc!14]  (b2) at (2.10,2.55) {action map $m^{\mathrm{lstm}}_t$};
    \node[bar, draw=grayc!85, fill=grayc!14]  (b3) at (2.10,2.05) {cursor heat map};
    \node[bar, draw=violc!80, fill=violc!12, text=violc!85!black] (b4) at (2.10,1.55) {memory map $m^{\mathrm{mem}}_t$};
    \node[bar, draw=tealc!80, fill=tealc!10]  (b5) at (2.10,1.05) {noised current screen $\tilde{\z}_t$};

    \filldraw[draw=bluec!70, fill=bluec!9, rounded corners=2pt, line width=0.55pt]
      (5.35,3.30) -- (7.05,2.55) -- (8.75,3.30) -- (8.75,0.80) -- (7.05,1.55) -- (5.35,0.80) -- cycle;
    \node[font=\sffamily\scriptsize\bfseries, text=black!85] (unet) at (7.05,2.05) {Residual UNet};
    \draw[arr] (4.20,2.05) -- (5.35,2.05);

    \node[ttl, anchor=south west] at (4.40,5.27) {Token bank $\Csem$ ($52$ tokens)};
    \node[bar, minimum width=3.0cm, draw=violc!80, fill=violc!12, text=violc!85!black] (t1) at (5.90,4.95) {$36$ memory tokens $R_t$};
    \node[bar, minimum width=2.8cm, draw=orc!80, fill=orc!12, text=orc!85!black, anchor=west] (t2) at (t1.east) {$16$ VLM tokens $\Hout$};
    \draw[arr] (7.05,4.70) -- (7.05,2.92);
    \node[note, anchor=west, align=left] at (7.20,3.95) {every UNet stage\\attends over the bank};

    \node[pill, draw=tealc!80, fill=tealc!8] (dz) at (10.35,2.05) {$\hat{\Delta}\z_t$};
    \node[note, anchor=south] at (10.35,2.40) {predicted change};
    \node[circle, draw=black!65, fill=white, line width=0.5pt, inner sep=0.5pt, minimum size=0.38cm,
          font=\sffamily\scriptsize\bfseries] (add) at (11.75,2.05) {$+$};
    \node[pill, draw=tealc!80, fill=tealc!8] (zt) at (11.75,0.95) {$\z_t$};
    \node[note, anchor=north] at (11.75,0.62) {current screen};
    \node[pill, draw=tealc!80, fill=tealc!8] (zn) at (13.15,2.05) {$\hat{\z}_{t+1}$};
    \node[frz, minimum width=1.5cm, minimum height=0.6cm] (dec) at (14.90,2.05) {decoder\subt{frozen}};
    \node[pill] (xn) at (16.65,2.05) {$\hat{\x}_{t+1}$};
    \node[note, anchor=south] at (16.65,2.40) {next screen};

    \draw[arr] (8.75,2.05) -- (dz);
    \draw[arr] (dz) -- (add);
    \draw[arr] (zt) -- (add);
    \draw[arr] (add) -- (zn);
    \draw[arr] (zn) -- (dec);
    \draw[arr] (dec) -- (xn);
  \end{tikzpicture}%
  }
  \caption{\textbf{Zoom-in on Module~3 (Interface) of Main Figure~2: memory and compute reach the renderer as a 2D map stacked with the other inputs, and as tokens that every UNet stage attends over.}
  \emph{Left, the stacked maps.} Five maps share the latent grid and are concatenated as channels at the UNet input:
  the noised change $\Delta\z^{(k)}_t$, which is the quantity being denoised (the difference between the next and the current screen latent, at diffusion step $k$);
  the action map $m^{\mathrm{lstm}}_t$, in which a pretrained LSTM summarizes recent mouse and keyboard input;
  a heat map peaked at the cursor;
  the memory map $m^{\mathrm{mem}}_t$, which holds what each screen location reads from the slot memory;
  and a noised copy $\tilde{\z}_t$ of the current screen, corrupted so that the model cannot simply copy it.
  \emph{Top, the token bank.} Some content has no screen position: the $36$ memory tokens $R_t$ ($4$ read heads, each returning $1$ semantic and $8$ visual tokens) and the $16$ VLM tokens $\Hout$, which carry the computed outcome. Both are projected to a common width to form the $52$-token bank $\Csem$. At each of its $7$ stages, every location of the UNet attends over the bank and adds the result through a learned gate that starts small, so the pretrained renderer is not disturbed at initialization.
  \emph{Right, the output.} The UNet predicts only the change in the latent, $\hat{\Delta}\z_t$. Adding it to the current latent $\z_t$ gives the next latent, which a frozen decoder turns into the next screen $\hat{\x}_{t+1}$; regions that do not change are therefore carried over untouched. The UNet is trained with $\Lren$ (\cref{eq:suppl_lren}). Schematic.}
  \label{fig:unetcond_suppl}
\end{figure*}

\paragraph{How do we harness a VLM's symbolic reasoning without paying for test-time text decoding?}
Approximate dynamics look convincing in natural video, but typing \texttt{3+3} in a terminal must produce \texttt{6} every time.
So, we need our model to have an understanding of symbolic computation. However, learning exact arithmetic from pixels alone is prohibitively slow. We therefore asked whether we could reach for a pretrained VLM here: most pretrained VLMs \textit{already know} how to compute; the one remaining question is whether we can route a VLM's answer into the renderer.
We hit two problems when we went down this path. Reading prompt features directly from the VLM without text supervision does not reveal the answer. Waiting for the VLM to autoregressively generate the answer is far more accurate; however, this dramatically slows down inference, as we must wait for a full autoregressive rollout of the VLM on every time-step.

Our post-prefix causal query bridge (\Cref{fig:vlm_bridge}, zooming into Module~2 of Main Figure~2) resolves this trade-off with a single ordering trick.
During training, we append $N_{\mathrm{out}}{=}16$ learnable query tokens $\Qout$ immediately after the multimodal prefix $[\x_t;\, \tilde{R}_t;\, a_t]$ and strictly before the target outcome string $y_t$.
Under causal attention, $\Qout$ reads the prefix without seeing $y_t$ ($\partial\Hout/\partial y_t = 0$), so at test time we simply drop $y_t$ and compute the query hidden states $\Hout \in \R^{16 \times 2048}$ in one forward pass.
During training, however, $y_t$ attends back over $\Qout$: next-token cross-entropy $\Lexe$ over $y_t$ forces the VLM to retrieve cleared variables from $\tilde{R}_t$, execute the operation, and pack the answer into $\Hout$.
Finally, an auxiliary linear head trained with cross-entropy $\Lprobe$ predicts the first $16$ tokens of $y_t$ from $\Hout$ through the frozen language-model output head, keeping the answer linearly readable by the renderer.
Removing both execution losses drops cleared-screen arithmetic accuracy to $0.0\%$ (\Cref{sec:suppl:fail_compute}).

\subsection{Architecture Footprint and Resolution Bridging}
\label{sec:suppl:arch}
A practical neural computer must keep memory overhead constant as sessions grow long and scale to high-resolution typography without blowing up parameter count or latency.

\paragraph{How much overhead do Memory and Compute add, and how are user actions encoded?}
Because off-screen state is compressed into a fixed bank of $64$ slots rather than a growing frame buffer or KV cache, memory cost is independent of rollout length: the slot memory adds $7.78\text{M}$ trainable parameters and holds a constant $0.41$\,MB FP32 state per episode. In summary, the fixed memory component adds very little overhead, leaving substantial room to scale it up.

On the compute side, we keep the VLM backbone frozen and adapt only its language layers with LoRA~\cite{hu2021lora} ($r{=}16$; $17.4\text{M}$ parameters for \texttt{Qwen3-VL-2B-Instruct}~\cite{bai2025qwen3vl}).
For rendering, pretrained short-term action LSTMs~\cite{rivard2025neuralos} supply the local conditioning map $m^{\mathrm{lstm}}_t$, while gated 2D cross-attention layers in the Residual UNet (\Cref{fig:unetcond_suppl}) inject the projected memory and compute tokens $\Csem \in \R^{52 \times 512}$.
Because symbolic reasoning and 2D rendering need different views of user input, each step provides both a verbatim keystroke string $a_t$ for the VLM (preserving exact character order and shortcuts such as \texttt{Ctrl+L}) and a 2D cursor heat map, click flags, and multi-hot key vector for the UNet and LSTMs.

\begin{figure}[t]
\centering
\includegraphics[width=\linewidth]{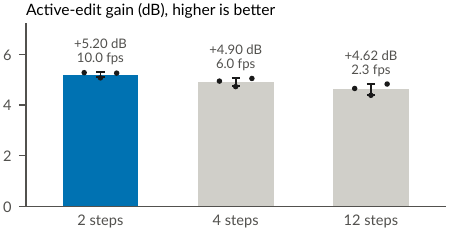}
\caption{\textbf{Two DDIM steps suffice at $1024{\times}768$: more steps are slower and do not improve active-edit gain.}
Active-edit gain of \textbf{Ours} (the model of Main Table~1B) at $2$, $4$ and $12$ sampling steps, with the end-to-end frame rate on one GH200 GPU above each bar; we use $2$ steps (blue). Gain falls slightly with more steps, by $0.6$\,dB from $2$ to $12$.
Bars are means, error bars s.d.\ and dots the $3$ trained seeds, on $862$ active transitions per seed.}
\label{fig:t6_pareto}
\end{figure}

\paragraph{How does the same memory scale to $1024{\times}768$ at interactive frame rates?}
The 4-channel $512{\times}384$ Stable Diffusion VAE caps terminal legibility at $56.3\%$ OCR, so scaling to $1024{\times}768$ with the 16-channel \texttt{FLUX.1-schnell} f8 VAE~\cite{blackforestlabs2024flux} ($96{\times}128$ latents) is essential for crisp text.
To decouple memory size from pixel resolution, the memory bank always operates on a fixed $48{\times}64$ spatial grid: moving from $512{\times}384$ to $1024{\times}768$ adds zero memory parameters because we simply $2{\times}2$ average-pool the latents to $48{\times}64$ before memory read and write, and bilinearly upsample $m^{\mathrm{mem}}_t$ back to $96{\times}128$.
To keep generation interactive, we sample with $2$ DDIM steps~\cite{song2021denoising} at $1024{\times}768$ (while every $512{\times}384$ model uses $32$ steps to match \textsc{NeuralOS}; \Cref{tab:budgets}).
As \Cref{fig:t6_pareto} shows, active-edit gain at $1024{\times}768$ is $+5.20 \pm 0.11$\,dB at $2$ steps, $+4.90 \pm 0.16$\,dB at $4$, and $+4.62 \pm 0.22$\,dB at $12$, so extra steps cost speed without improving quality; we have not tested the step count at $512{\times}384$.
At $2$ steps the full model runs at $10.0 \pm 0.4$\,fps end to end on one NVIDIA GH200 GPU over three seeds; Main Table~1B reports the first seed ($10.5$\,fps).

\subsection{Evaluation Suites and Scoring Protocols}
\label{sec:suppl:eval_protocol}
Why do we evaluate functional benchmarks alongside PSNR and LPIPS?
Roughly $95\%$ of a desktop does not change on each time-step, so a model that simply copies the previous frame---or, worse, hallucinates plausible-looking nonsense when reopening a closed file---can still achieve respectable single-step perceptual scores. Evaluating desktop world models solely on perceptual metrics such as PSNR and LPIPS is therefore a major blind spot of existing benchmarks.

To separate a working computer from a stateless frame interpolator, we pair standard perceptual metrics with direct tests of object permanence, causal memory use, and symbolic execution.

We evaluate in two separate settings: a controlled $512{\times}384$ suite that matches \textsc{NeuralOS}~\cite{rivard2025neuralos} in codec and resolution for fair baseline comparisons, and a $1024{\times}768$ suite using the \texttt{FLUX.1-schnell} autoencoder.
Models are not compared across resolutions.
\begin{itemize}[leftmargin=1.3em,itemsep=2pt,topsep=2pt]
  \item \textbf{Hidden-shape recall and same-pane overwrite.}
  Each matched triplet pairs three VS Code episodes whose hidden file holds a triangle, circle, or star.
  At the reopen frame (downsampled to $256{\times}192$ in $[0,1]$), we extract the region where the three ground-truth targets $T_u$ ($u \in \{0,1,2\}$) differ, $\mathcal{R} = \{(y,x) : \max_c(\max_u T_u - \min_u T_u) > 0.1\}$ ($|\mathcal{R}| \ge 500$ pixels).
  \emph{$3$-way recall accuracy} is the fraction of generated frames $\hat{\x}$ whose mean $L_1$ error on $\mathcal{R}$ is lowest for the true shape ($33.3\%$ chance - a model that copies the current frame scores exactly $33.3\%$).
  The \emph{Desktop} dev set has $32$ matched triplets, of which $18$ have an edited hidden file and $14$ an unedited one.
  Edited-file recall uses the $18$ edited triplets: $54$ episodes, each generated with $3$ noise samples, give $162$ evaluations per trained model.
  Overwrite uses all $32$ triplets: $96$ episodes with $3$ noise samples give $288$ evaluations per $\Delta t$.
  In this supplement, $\pm$ is the standard deviation over independently trained models.
  The dev triplets are divided once into a $17$-triplet half and a $15$-triplet half; the ablations (\Cref{sec:suppl:ablations}) are scored on the first ($9$ edited triplets, $81$ evaluations per trained model).
  Edited-file recall, overwrite and donor-swap results are in Main Table~2.

  \item \textbf{XFCE File Manager persistence.}
  Matched episodes create \texttt{New Folder} inside \texttt{Documents}, close the File Manager for $64$, $128$, or $256$ steps, and reopen \texttt{Documents}.
  We show these episodes qualitatively (Main Figure~3 and \Cref{sec:suppl:qualitative}).

  \item \textbf{Counterfactual donor memory swap.}
  To test whether recall is causally driven by the explicit slot bank rather than by shortcuts in the current frame or LSTMs, we hold $\x_t$, $a_t$, and the LSTM states fixed at the reopen frame of episode $i$ and replace $M_t \leftarrow M^{\mathrm{donor}}_t$ from an episode $j$ of the same triplet with a different shape.
  \emph{Donor-swap follow} is the fraction of trials in which the generated frame has lower mean $L_1$ error, on the same region $\mathcal{R}$ as above, to the donor's shape than to the original ($50.0\%$ chance).
  The model is trained with $\mathcal{L}_{\mathrm{swap}}$ (\Cref{eq:suppl_lswap}), which rewards sensitivity of the memory prediction to such a swap.

  \item \textbf{Perceptual fidelity and active-edit gain.}
  Single-step quality is measured with AlexNet LPIPS v0.1~\cite{zhang2018unreasonable}, InceptionV3 FID~\cite{heusel2017gans}, PSNR, and SSIM.
  To discount trivial frame copying, on active transitions ($\ge 64$ pixels changing by $>2/255$) we also report \emph{active-edit gain}: changed-pixel PSNR minus the changed-pixel PSNR of copying $\x_t$.
  $1024{\times}768$ scores use $862$ transitions per seed, and the $512{\times}384$ scores of \Cref{tab:suppl_neuralos_zeroshot} use $930$ \emph{Desktop} and $701$ \emph{Maths} frames.

  \item \textbf{CLI execution and terminal legibility.}
  The \emph{Maths} dev split has $324$ scored transitions across four Python REPL families, of which $124$ are cleared-operand transitions where \texttt{Ctrl+L} wipes the screen before the query.
  \emph{Cleared-screen answers (\%)} is greedy exact match of the VLM's decoded answer, and \emph{$\Hout$ probe (\%)} is argmax exact match from the single-pass linear head.
  \emph{Terminal OCR (\%)} is the fraction of earlier terminal lines ($n{=}157$) that \texttt{EasyOCR} recovers, as a case-insensitive substring, from the generated frame upscaled $2{\times}$.
\end{itemize}

\subsection{Training and Optimization}
\label{sec:suppl:training}
How do we train a memory to survive long occlusions under bounded GPU memory?
Every model trains on windows of $128$ consecutive frames (\Cref{fig:unroll}).
The first $64$ frames are a warm-up: the model runs forward on them without gradients, so that by frame $65$ its memory already holds what the episode showed earlier, such as a file that was edited and then closed.
The remaining $64$ frames are trained with truncated BPTT, with a loss on $8$ of them.
We use \texttt{AdamW} ($\beta_1{=}0.9$, $\beta_2{=}0.999$, weight decay $0.01$), a linear learning-rate warm-up to a constant rate ($200$ steps at $512{\times}384$; $150$ to $300$ at $1024{\times}768$), global gradient-norm clipping at $1.0$, and a $T{=}1{,}000$ diffusion schedule.
\Cref{tab:budgets} gives the training budget and sampling settings of every model, and \Cref{tab:hyperparams} the remaining settings of the $512{\times}384$ model.
All $512{\times}384$ models are initialized from \textsc{NeuralOS}'s \texttt{joint-272k} checkpoint; the baseline is fine-tuned with the published \textsc{NeuralOS} recipe.
The $1024{\times}768$ runs of \textbf{Ours} stop at a wall-clock limit, so their step counts are means over seeds ($867$, $834$ and $826$ steps on the Dense Interactive Mixture; $1{,}510$, $2{,}105$ and $2{,}069$ on the Broad Mixture).
Notably, every budget difference favors the baselines rather than \textbf{Ours}: fine-tuned \textsc{NeuralOS} sees three times as many windows as our $512{\times}384$ model, and \textsc{Copy-Only} sees about four times as many as ours on the Dense Interactive Mixture.
The $1024{\times}768$ models use the memory and loss settings of \Cref{tab:hyperparams} with a learning rate of $10^{-5}$ for pretrained weights and $10^{-4}$ ($1.5{\times}10^{-4}$ on the Dense Interactive Mixture) for new modules.

\section{Model Hyperparameters}
\label{sec:suppl:hyperparams}
\Cref{tab:hyperparams} lists the settings of the $512{\times}384$ model, as trained for the ablation experiment of \Cref{sec:suppl:ablations}; training budgets are in \Cref{tab:budgets}.

\begin{table}[t]
  \centering
  \footnotesize
  \setlength{\tabcolsep}{1.5pt}
  \begin{tabular}{@{}lll@{}}
    \toprule
    Symbol & Value & Role \\
    \midrule
    $M, H_r$ & $64,\; 4$ & Slots and read heads \\
    -- & $64,\; 512,\; 8{\times}128$ & Key, semantic, visual sizes \\
    $\mu_{\mathrm{EMA}}$ & $0.996$ & JEPA target momentum \\
    $\alpha_s, \beta_s$ & $4.0,\; 1.25$ & Surprise-gate scale, offset \\
    $\tau_w, \tau_r, \tau_a$ & $0.1,\; 0.1,\; 0.05$ & Write, read, alloc.\ temp. \\
    $\lambda_{\mathrm{prec}}, u_{\mathrm{free}}$ & $5,\; 0.1$ & Alloc.\ penalty, threshold \\
    $\lambda_{\mathrm{align}}, \lambda_q$ & $0.5,\; 0.5$ & Alignment, query weights \\
    -- & $0.07$ & InfoNCE temperature \\
    $\delta_{\mathrm{stat}}, \delta_{\mathrm{chg}}$ & $0.02,\; 0.35$ & Static, changed thresh. \\
    $\lambda_{\mathrm{VICReg}}, \lambda_{\mathrm{swap}}, \rho$ & $0.25,\; 0.5,\; 0.1$ & VICReg, swap, margin \\
    $N_{\mathrm{out}}$ & $16$ & VLM outcome queries \\
    $r$ & $16$ & LoRA rank \\
    $\lambda_{\mathrm{JEPA}}, \lambda_{\mathrm{exe}}, \lambda_{\mathrm{probe}}$ & $0.5,\; 0.2,\; 0.1$ & Joint loss weights \\
    $\omega_{\mathrm{chg}}, \lambda_{\mathrm{static}}$ & $4,\; 0.1$ & Rendering loss (\Cref{eq:suppl_lren}) \\
    $g_0$ & $0.25$ & Cross-attn.\ gate init \\
    $p_{\mathrm{drop}}$ & $0.5$ & Blank screen to the VLM \\
    -- & $0.3$ & Reference-copy dropout \\
    -- & $10^{-4},\; 10^{-5}$ & LR: new, pretrained \\
    -- & $200,\; 3{,}000$ & LR warm-up, total steps \\
    -- & $16$ & Batch size \\
    -- & $8$ & Loss frames per window \\
    \bottomrule
  \end{tabular}
  \caption{\textbf{Settings of the $512{\times}384$ model.} Symbols follow Main Section~3 and \Cref{sec:suppl:losses}; rows without a symbol are described in \Cref{sec:suppl:training}. $s_c$ is the per-channel standard deviation of the training residuals.}
  \label{tab:hyperparams}
\end{table}

\section{Data Corpus Provenance and \texorpdfstring{$1024{\times}768$}{1024x768} Mixture}
\label{sec:suppl:data}
\begin{table*}[t]
  \centering
  \footnotesize
  \setlength{\tabcolsep}{5pt}
  \begin{tabular}{@{}llrrl@{}}
    \toprule
    Source & Origin / License & Train / Dev ep. & Share & Notes \\
    \midrule
    \emph{Desktop} (VS Code) & In-house XFCE & $768$ / $96$ & $12.5\%$ & $32$ matched triplets \\
    \emph{Maths} (Python REPL) & In-house XFCE & $4{,}000$ / $300$ & $12.5\%$ & $4$ arithmetic families \\
    AgentNet~\cite{wang2025opencua} & \texttt{xlangai/AgentNet} (MIT) & $21{,}957$ / $425$ & \multirow{3}{*}{$75.0\%$} & $2{,}500$ ep.\ at $1024{\times}768$ \\
    ProCUA-SFT~\cite{jung2026procua} & \texttt{nvidia/ProCUA-SFT} (CC BY 4.0) & $92{,}964$ / $396$ & & $2{,}500$ ep.\ at $1024{\times}768$ \\
    CUA-Gym\footnotemark & \texttt{p-doom/cuagym-qwen35-rollouts} (CC0) & $54{,}286$ / $531$ & & $2{,}500$ ep.\ at $1024{\times}768$ \\
    \midrule
    Total & $5$ sources ($>2.1$\,TB raw) & $173{,}975$ / $1{,}748$ & $100.0\%$ & $12{,}268$ ep.\ at $1024{\times}768$ \\
    \bottomrule
  \end{tabular}
  \caption{\textbf{The $1024{\times}768$ Broad Mixture combines $75\%$ public GUI trajectories with $25\%$ controlled XFCE tasks.}
  The cleaned pool holds $173{,}975$ training trajectories ($8.83\text{M}$ steps), and a stratified \texttt{FLUX.1-schnell} subset encodes $12{,}268$ trajectories ($694{,}266$ training windows) into FP16 $\R^{16\times 96\times 128}$ latents.}
  \label{tab:data_corpus}
\end{table*}
\footnotetext{\url{https://huggingface.co/datasets/p-doom/cuagym-qwen35-rollouts}}
\begin{table}[t]
  \centering
  \footnotesize
  \setlength{\tabcolsep}{2.6pt}
  \begin{tabular}{@{}lrrrr@{}}
    \toprule
    Model & Steps & \shortstack{Win.\\/step} & \shortstack{Win.\\total} & \shortstack{DDIM\\steps} \\
    \midrule
    \multicolumn{5}{@{}l}{\emph{$512{\times}384$}} \\
    \textbf{Ours} (\Cref{sec:suppl:ablations}) & $3{,}000$ & $16$ & $48$k & $32$ \\
    \textsc{NeuralOS}, fine-tuned & $9{,}000$ & $16$ & $144$k & $32$ \\
    \midrule
    \multicolumn{5}{@{}l}{\emph{$1024{\times}768$}} \\
    \textbf{Ours}, Dense Interactive & $842$ & $16$ & $13$k & $2$ \\
    \textbf{Ours}, Broad & $1{,}895$ & $16$ & $30$k & $2$ \\
    \textsc{Copy-Only} & $3{,}500$ & $16$ & $56$k & $2$ \\
    \bottomrule
  \end{tabular}
  \caption{\textbf{Training budget and sampling settings of every model.} Win./step is the number of $128$-frame training windows per optimizer step, and Win.\ total their product with the step count.}
  \label{tab:budgets}
\end{table}
\begin{figure*}[t]
  \centering
  \includegraphics[width=\textwidth]{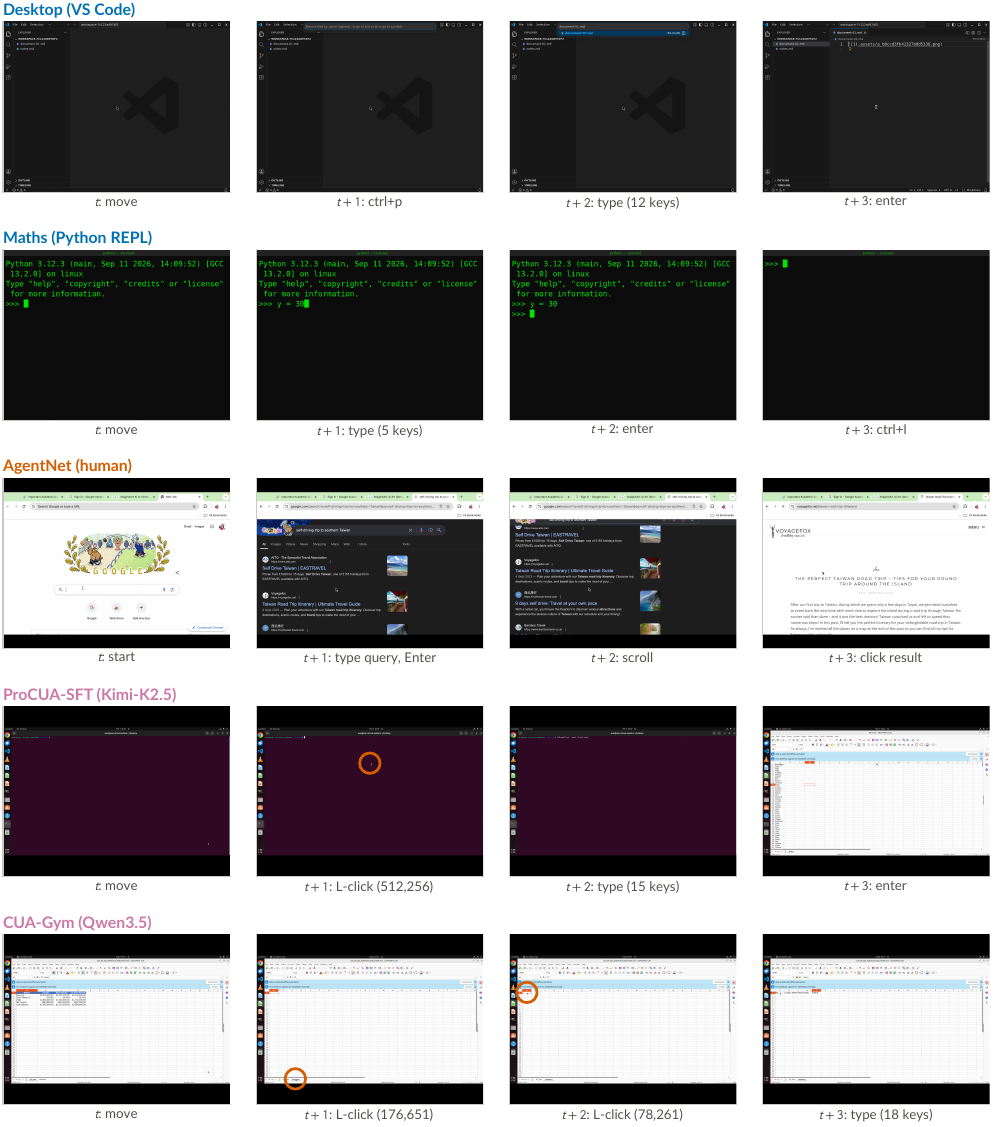}
  \caption{\textbf{Representative four-frame trajectories across all five training sources.}
  Top two rows: controlled XFCE \emph{Desktop} persistence and \emph{Maths} \texttt{Ctrl+L} arithmetic. Bottom three rows: letterboxed public GUI corpora (orange circles mark clicks). Each frame is labelled with the action that led to it.}
  \label{fig:data_examples}
\end{figure*}

\begin{figure*}[t]
  \centering
  \includegraphics[width=\textwidth]{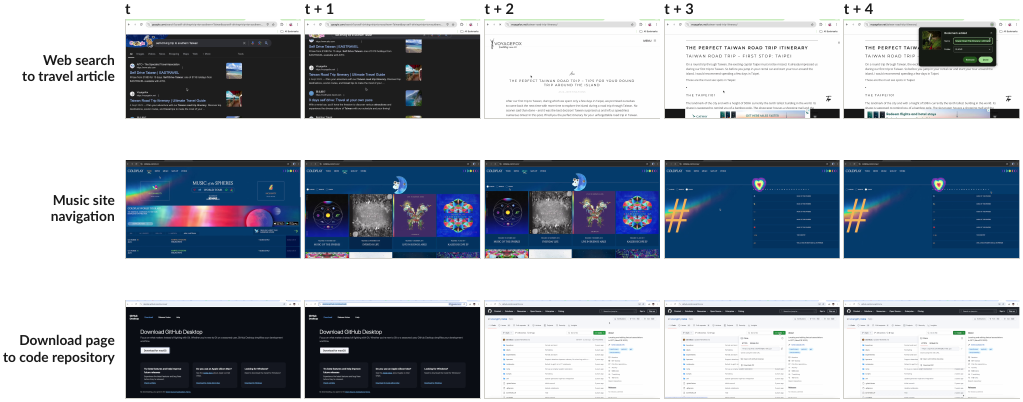}
  \caption{\textbf{Three AgentNet episodes from the $1024{\times}768$ open-world corpus: five consecutive recorded steps each.}
  Frames are shown at full resolution with the letterbox bars cropped. Episodes are among the first dev episodes with six consecutive steps of visible change.}
  \label{fig:suppl_agentnet_active_gui}
\end{figure*}

\begin{figure*}[t]
  \centering
  \includegraphics[width=\textwidth]{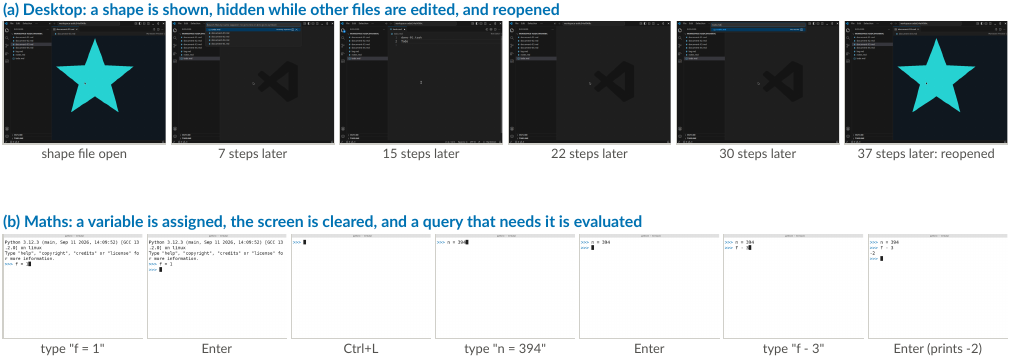}
  \caption{\textbf{The two controlled tasks, each shown as one complete recorded episode.}
  (a)~\emph{Desktop}: the reopened frame can only be drawn correctly from memory of the first.
  (b)~\emph{Maths}: the value of \texttt{f} is no longer on screen when \texttt{f - 3} is evaluated.
  All frames are recordings at $512{\times}384$; each is labelled with the action that led to it.}
  \label{fig:suppl_agentnet_gt_candidates}
\end{figure*}

Public GUI corpora offer broad application diversity but rarely revisit closed files or test arithmetic after the screen is cleared.
Pairing three public desktop corpora with our controlled XFCE recordings gives $173{,}975$ cleaned training trajectories across all five sources ($169{,}207$ public and $4{,}768$ in-house XFCE, \Cref{tab:data_corpus}).
\Cref{fig:data_examples} shows all five sources, \Cref{fig:suppl_agentnet_active_gui} shows three AgentNet episodes at full resolution, and \Cref{fig:suppl_agentnet_gt_candidates} shows one complete episode of each controlled task.

\paragraph{Alignment and sanitization.}
The sources differ in aspect ratio, action logging, and frame timing, so we normalize each before training.
We letterbox all $16{:}9$ recordings onto a $4{:}3$ canvas and parse \texttt{pyautogui} scripts with a non-executing abstract syntax tree (AST) parser.
This removes $194$ corrupted trajectories ($45$ ProCUA-SFT records with misparsed $(0,0)$ scroll coordinates and $149$ AgentNet records with missing gap frames) and corrects a one-frame action offset in ProCUA-SFT, so that every step obeys $(\x_t, a_t) \to \x_{t+1}$.
Episodes shorter than $64$ frames are left-padded with zero latents, which are masked out of all losses.

\section{Four Ablations at $3{,}000$ Steps}
\label{sec:suppl:three_pillar}
\label{sec:suppl:ablations}

The ablations are separate, shorter experiments.
Every arm, including the full model, was trained for only $3{,}000$ steps (\Cref{sec:suppl:training}), so the numbers in \Cref{sec:suppl:fail_jepa,sec:suppl:fail_addressing,sec:suppl:fail_compute,sec:suppl:fail_renderer} should be compared with each other and not with the main tables, which report longer training.

In this setting, the full model reaches $40.3 \pm 0.7\%$ edited-file recall, $28.2 \pm 2.1\%$ cleared-screen answers and $43.5 \pm 2.2\%$ Terminal OCR.
Each ablation removes one component from the full model\footnote{The full model and the arms without the execution loss or the residual target have three seeds. The arms without the memory loss or the surprise gate have two.}.
Answer accuracy is counted over $124$ transitions, so it moves in steps of $0.8$ points; several arms that keep the execution loss average to the same value, $28.2\%$ ($35$ of $124$).

\subsection{Memory Supervision (\textsc{Memory-JEPA})}
\label{sec:suppl:fail_jepa}
The memory objective is intended to supply a learning signal across an occlusion, where a hidden file affects only a small patch of the single reopen frame.
Without the memory objective ($\Lpred{=}0$), edited-file recall is $32.1\%$ (seeds $25.9\%$ and $38.3\%$), against $40.3 \pm 0.7\%$ for the full model of the same experiment; cleared-screen answers ($28.2\%$) and Terminal OCR ($45.2\%$) are unchanged.

\subsection{Surprise Gating}
\label{sec:suppl:fail_addressing}
\label{sec:suppl:slot_capacity}
\label{sec:suppl:slots}
Writing on every frame ($\eta_t \equiv 1$) gives $35.8\%$ edited-file recall (seeds $33.3\%$ and $38.3\%$), against $40.3 \pm 0.7\%$ for the full model of the same experiment, which uses the surprise gate~\cite{rao1999predictive,friston2010free,yang2025cambrians}. 

\subsection{Symbolic Compute Supervision}
\label{sec:suppl:fail_compute}
Cleared-screen answer accuracy scores the VLM's decoded text, not the rendered screen.
With the execution losses, the full model answers $28.2 \pm 2.1\%$ of the $124$ cleared-screen queries ($15.5\%$ of arithmetic queries and $38.7\%$ of number-recall queries), against $65.5 \pm 0.5\%$ of the $200$ queries whose operands are still on screen.
Without them ($\Lexe{=}0$, $\Lprobe{=}0$), cleared-screen answer accuracy is $0.0\%$ in all three seeds, and $\Hout$ probe accuracy falls from $23.7 \pm 2.8\%$ to $0.0\%$.
The answer does not arise from rendering supervision alone.
\begin{figure*}[t]
  \centering
  \vspace{-2pt}
  \sfbanner{after the first frame (at $t={17}$), the model is auto-regressively fed its own previous output.}
  \includegraphics[width=\linewidth]{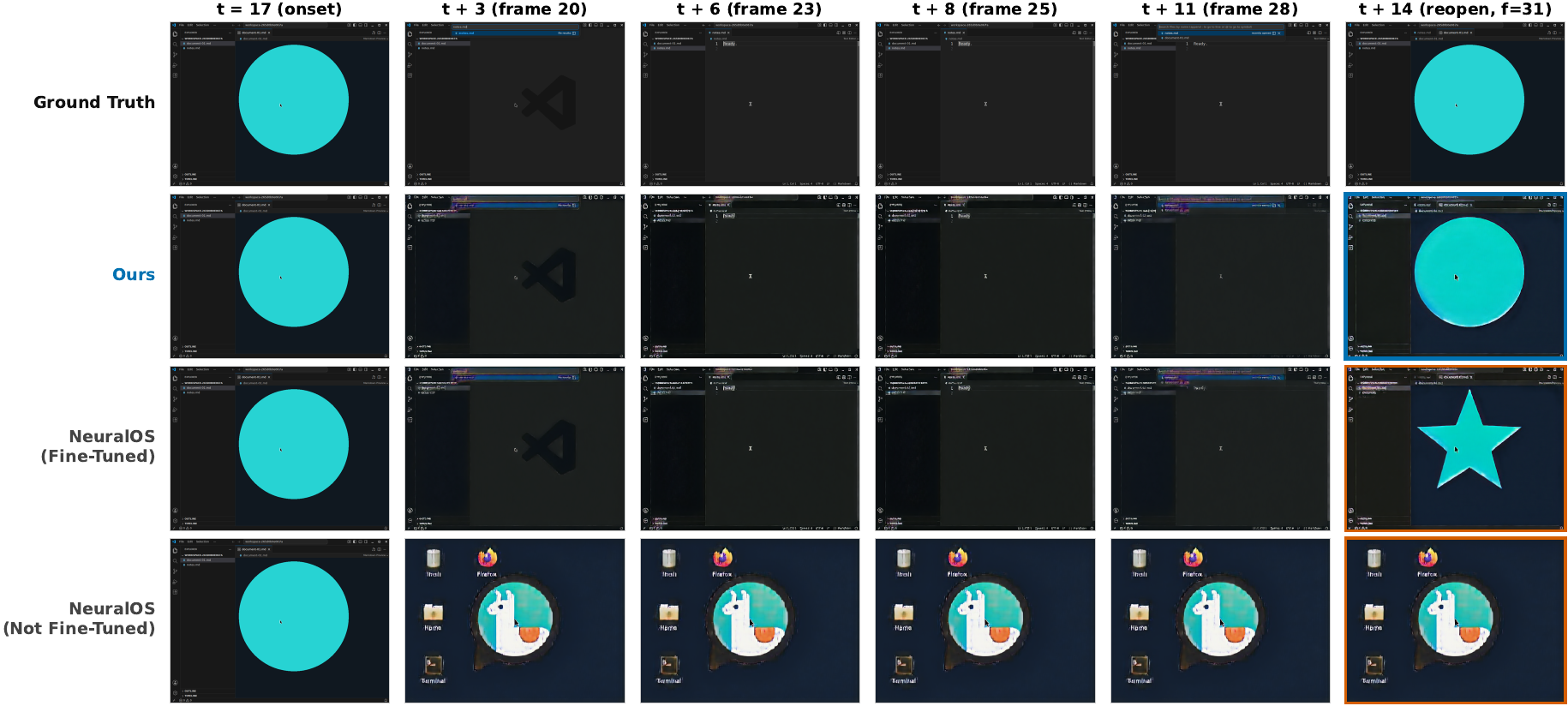}
  \vspace{-6pt}
  \caption{\textbf{\textbf{Ours} recovers the edited circle after $14$ closed-loop VS Code steps, whereas \textsc{NeuralOS} reverts to the pre-edit star.}
  Student-forced: from $t{=}17$ the model is given only its own previous output. Unmodified \textsc{NeuralOS} (\texttt{joint-272k}) snaps back to its training desktop.}
  \label{fig:rollout_vscode_r1_512}
  \vspace{-4pt}
\end{figure*}

\subsection{Residual Target; Screen-Dropout Setting}
\label{sec:suppl:fail_renderer}
\label{sec:suppl:use}
Regenerating the full frame at every step blurs small glyphs.
Replacing the residual target $\Delta\z_t$ with full-frame $x_0$ prediction lowers Terminal OCR from $43.5 \pm 2.2\%$ to $0.2 \pm 0.4\%$ and leaves cleared-screen answers unchanged ($28.2 \pm 2.4\%$).
In this experiment the residual target trades recall for legible text: the full-frame arm has higher edited-file recall ($65.4 \pm 10.5\%$ against $40.3 \pm 0.7\%$).
During training the VLM sees a blank screen in place of $\x_t$ with probability $p_{\mathrm{drop}}{=}0.5$, and the reference copy $\tilde{\z}_t$ is dropped with probability $0.3$; we do not report an ablation of either setting.

\section{Baseline Verification and Closed-Loop Rollouts}
\label{sec:suppl:rollout_horizons}
\label{sec:suppl:rollouts}

In a closed-loop rollout the model's own output replaces the ground-truth frame at every step, so errors can compound.
\Cref{fig:rollout_vscode_r1_512} shows one such student-forced rollout in VS Code, \Cref{sec:suppl:qualitative} shows further examples, and Main Table~1A reports PSNR after $261$ self-generated steps.
\begin{table*}[t]
  \centering
  \small
  \setlength{\tabcolsep}{6.5pt}
  \begin{tabular}{@{}llccccc@{}}
    \toprule
    Split (frames) & \textsc{NeuralOS} ($512{\times}384$) & LPIPS $\downarrow$ & PSNR $\uparrow$ & SSIM $\uparrow$ & FID $\downarrow$ & Terminal OCR (\%) $\uparrow$ \\
    \midrule
    \multirow{2}{*}{\emph{Desktop} ($930$)} & Public release (zero-shot) & $0.708$ & $9.84$ & $0.565$ & $376.8$ & -- \\
     & Fine-tuned ($9\text{k}$ steps) & $\mathbf{0.113}$ & $\mathbf{25.66}$ & $\mathbf{0.914}$ & $\mathbf{67.9}$ & -- \\
    \midrule
    \multirow{2}{*}{\emph{Maths} ($701$)} & Public release (zero-shot) & $0.714$ & $9.04$ & $0.425$ & $466.1$ & $0.0$ \\
     & Fine-tuned ($9\text{k}$ steps) & $\mathbf{0.127}$ & $\mathbf{24.05}$ & $\mathbf{0.795}$ & $\mathbf{85.2}$ & $0.0$ \\
    \bottomrule
  \end{tabular}
  \caption{\textbf{Fine-tuning \textsc{NeuralOS} on our corpus adds $+15$--$16$\,dB over the unmodified public release.}
  Evaluated on identical $512{\times}384$ dev frames ($32$ DDIM steps, seed $0$). Zero-shot \texttt{joint-272k} renders its own training desktop, and even after fine-tuning Terminal OCR stays at $0.0\%$ ($n{=}157$).}
  \label{tab:suppl_neuralos_zeroshot}
\end{table*}

\paragraph{Zero-shot \textsc{NeuralOS}.}
The public \texttt{joint-272k} checkpoint, evaluated without fine-tuning on our XFCE suite, follows clicks but renders its own training desktop ($9.84$\,dB PSNR and $0.708$ LPIPS on \emph{Desktop}, $9.04$\,dB and $0.714$ on \emph{Maths}, \Cref{tab:suppl_neuralos_zeroshot}).
Fine-tuning it on our corpus adds $15.8$\,dB on \emph{Desktop} ($25.66$\,dB) and $15.0$\,dB on \emph{Maths} ($24.05$\,dB), and we use the fine-tuned model as the baseline throughout.
\Cref{tab:suppl_neuralos_zeroshot} is a single evaluation seed, so its values differ slightly from the three-seed means of Main Table~1A.

\section{High-Resolution \texorpdfstring{$1024{\times}768$}{1024x768} Training on the Broad Mixture}
\label{sec:suppl:1024_flagship}
\label{sec:suppl:hires}
We also train the $1024{\times}768$ model on the $694{,}266$-window Broad Mixture (\Cref{tab:data_corpus}), in addition to the Dense Interactive Mixture of Main Table~1B.
Training ran for $21.5$\,h on $4$ GH200 GPUs, which covers $0.044$ epochs.
Between step $500$ and the last checkpoint, whole-frame PSNR rises from $27.8$ to $29.6$\,dB, active-edit gain from $+0.09$ to $+2.97$\,dB, and AlexNet LPIPS falls from $0.074$ to $0.056$ (\Cref{fig:training_progression}).
These are rendering metrics on the $862$ active dev transitions used for Main Table~1B: $474$ from VS Code, $324$ from Maths and $64$ from AgentNet. This set is $93\%$ in-house data, which is all of the Dense Interactive Mixture and a quarter of the Broad Mixture, so it favors the model trained on the former. We report no memory or compute metric for this run.

\section{Training Progression and Compute Budget}
\label{sec:suppl:scaling}
\begin{figure*}[t]
\centering
\includegraphics[width=0.66\textwidth]{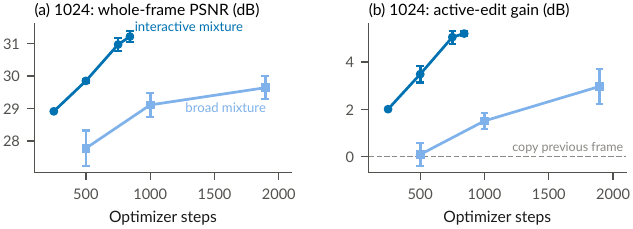}
\caption{\textbf{Rendering quality at $1024{\times}768$ is still rising when training stops.}
(a,~b) Whole-frame PSNR and active-edit gain on the Dense Interactive ($6.3$\,h) and Broad ($21.5$\,h) mixtures on $4$ GH200 GPUs.
Error bars are s.d.\ over $3$ trained seeds.}
\label{fig:training_progression}
\end{figure*}


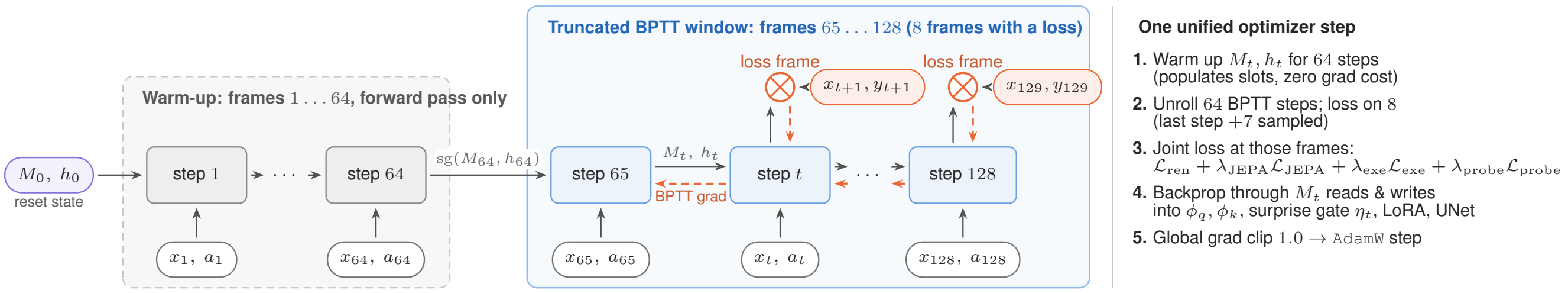
\begin{figure*}[t]
  \centering
  \resizebox{\textwidth}{!}{%
  \begin{tikzpicture}[archpic,
      cfwd/.style={fwd, shorten >=1.5pt, shorten <=1.5pt, rounded corners=4pt},
      cgrd/.style={grd, shorten >=1.5pt, shorten <=1.5pt, rounded corners=4pt}]

    \filldraw[rounded corners=4pt, draw=black!22, fill=black!3, densely dashed, line width=0.55pt]
      (0.80,-0.55) rectangle (5.45,2.45);
    \node[font=\sffamily\scriptsize\bfseries, text=black!65, anchor=north west] at (0.95,2.37)
      {Warm-up: frames $1\dots 64$, forward pass only};

    \filldraw[rounded corners=4pt, draw=bluec!50, fill=bluec!4, line width=0.6pt]
      (6.55,-0.55) rectangle (14.55,3.45);
    \node[font=\sffamily\scriptsize\bfseries, text=bluec!85!black, anchor=north west] at (6.72,3.37)
      {Truncated BPTT window: frames $65\dots 128$ ($8$ frames with a loss)};

    \node[dat, draw=violc!80, fill=violc!8] (s0) at (-0.25,1.05) {$M_0,\ h_0$};
    \node[note, anchor=north] at (s0.south) {reset state};

    \node[dat] (inw1) at (1.85,-0.15) {$x_1,\ a_1$};
    \node[frz, minimum width=1.42cm] (mw1) at (1.85,1.05) {step $1$};
    \draw[cfwd] (inw1) -- (mw1);

    \node at (3.12,1.05) {$\cdots$};

    \node[dat] (inw2) at (4.40,-0.15) {$x_{64},\ a_{64}$};
    \node[frz, minimum width=1.42cm] (mw2) at (4.40,1.05) {step $64$};
    \draw[cfwd] (inw2) -- (mw2);

    \node[dat] (inb1) at (7.60,-0.15) {$x_{65},\ a_{65}$};
    \node[mod, minimum width=1.42cm] (mb1) at (7.60,1.05) {step $65$};
    \draw[cfwd] (inb1) -- (mb1);

    \node[dat] (inbt) at (10.15,-0.15) {$x_t,\ a_t$};
    \node[mod, minimum width=1.42cm] (mbt) at (10.15,1.05) {step $t$};
    \draw[cfwd] (inbt) -- (mbt);

    \node at (11.42,1.05) {$\cdots$};

    \node[dat] (inb2) at (12.75,-0.15) {$x_{128},\ a_{128}$};
    \node[mod, minimum width=1.52cm] (mb2) at (12.75,1.05) {step $128$};
    \draw[cfwd] (inb2) -- (mb2);

    \draw[cfwd] (s0.east) -- (mw1.west);
    \draw[cfwd] (mw1.east) -- (2.86,1.05);
    \draw[cfwd] (3.38,1.05) -- (mw2.west);
    \draw[cfwd] (mw2.east) -- node[note, above=2pt, fill=white, inner sep=1.2pt]{$\mathrm{sg}(M_{64}, h_{64})$} (mb1.west);

    \draw[cfwd] ([yshift=3.5pt]mb1.east) -- ([yshift=3.5pt]mbt.west);
    \draw[cgrd] ([yshift=-3.5pt]mbt.west) -- ([yshift=-3.5pt]mb1.east);

    \draw[cfwd] ([yshift=3.5pt]mbt.east) -- (11.16,1.17);
    \draw[cgrd] (11.16,0.93) -- ([yshift=-3.5pt]mbt.east);
    \draw[cfwd] (11.68,1.17) -- ([yshift=3.5pt]mb2.west);
    \draw[cgrd] ([yshift=-3.5pt]mb2.west) -- (11.68,0.93);

    \node[note, anchor=south] at (8.88,1.22) {$M_t,\ h_t$};
    \node[note, text=orc!85!black, anchor=north] at (8.88,0.88) {BPTT grad};

    \foreach \nd/\tg in {mbt/t{+}1, mb2/129}{
      \node[loss] (l\nd) at (\nd |- 0,2.30) {};
      \draw[cfwd] ([xshift=-4pt]\nd.north) -- ([xshift=-4pt]l\nd.south);
      \draw[cgrd] ([xshift=4pt]l\nd.south) -- ([xshift=4pt]\nd.north);
      \node[gt, anchor=west] (g\nd) at ([xshift=6pt]l\nd.east) {$x_{\tg}, y_{\tg}$};
      \draw[cfwd] (g\nd.west) -- (l\nd.east);
      \node[lname, anchor=south] at (l\nd.north) {loss frame};
    }

    \draw[black!22] (14.88,-0.55) -- (14.88,3.45);
    \node[anchor=north west, font=\sffamily\scriptsize\bfseries, text=black!85] at (15.12,3.38) {One unified optimizer step};
    \node[anchor=north west, align=left, inner sep=1pt, font=\sffamily\scriptsize] at (15.12,2.84) {%
      \textbf{1.} Warm up $M_t, h_t$ for $64$ steps\\[-1pt]
      \phantom{\textbf{1.} }(populates slots, zero grad cost)\\[3pt]
      \textbf{2.} Unroll $64$ BPTT steps; loss on $8$\\[-1pt]
      \phantom{\textbf{2.} }(last step $+ 7$ sampled)\\[3pt]
      \textbf{3.} Joint loss at those frames:\\[-1pt]
      \phantom{\textbf{3.} }$\Lren + \lambda_{\mathrm{JEPA}}\Lpred + \lambda_{\mathrm{exe}}\Lexe + \lambda_{\mathrm{probe}}\Lprobe$\\[3pt]
      \textbf{4.} Backprop through $M_t$ reads \& writes\\[-1pt]
      \phantom{\textbf{4.} }into $\phi_q, \phi_k$, surprise gate $\eta_t$, LoRA, UNet\\[3pt]
      \textbf{5.} Global grad clip $1.0$ $\to$ \texttt{AdamW} step};
  \end{tikzpicture}%
  }
  \caption{\textbf{Each training window is $128$ frames: $64$ warm-up frames that only fill memory, then $64$ frames that are trained on.}
  The warm-up frames are run forward without gradients, so that the memory $M_t$ and the LSTM state $h_t$ already hold what happened earlier in the episode (for example a file that was edited and then closed).
  The state $(M_{64}, h_{64})$ is then detached, and the losses ($\Lren$, $\Lpred$, $\Lexe + \Lprobe$) are applied at $8$ of the next $64$ frames and backpropagated through the memory reads and writes within those frames.}
  \label{fig:unroll}
\end{figure*}

The $1024{\times}768$ models train on the $128$-frame windows of \Cref{sec:suppl:training} for $6.3$\,h (Dense Interactive) and $21.5$\,h (Broad) on $4$ NVIDIA GH200 GPUs, and their PSNR and active-edit gain are still rising at the last checkpoint (\Cref{fig:training_progression}).

\section{Additional Qualitative Rollouts}
\label{sec:suppl:qualitative}
\begin{figure*}[t]
  \centering
  \tfbanner
  \includegraphics[width=0.95\textwidth]{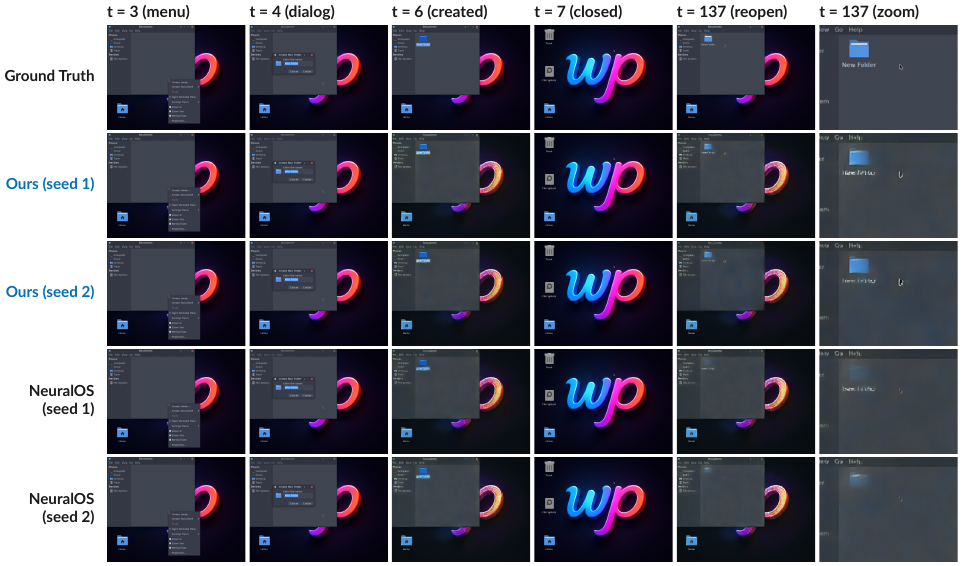}
  \caption{\textbf{Both seeds of \textbf{Ours} retain \texttt{New Folder} after a $128$-step occlusion, while fine-tuned \textsc{NeuralOS} blurs or erases it.}
  Teacher-forced: every frame is predicted from the real frames before it. A second episode of the XFCE File Manager task of Main Figure~3 ($512{\times}384$): folder created at $t{=}3..6$, reopened at $t{=}137$ (rightmost column: $2.5{\times}$ zoom).}
  \label{fig:suppl_xfce_desktop_g128}
\end{figure*}

\begin{figure*}[t]
  \centering
  \sfbanner{after the first frame, the model is auto-regressively fed its own previous output.}
  \includegraphics[width=0.84\textwidth]{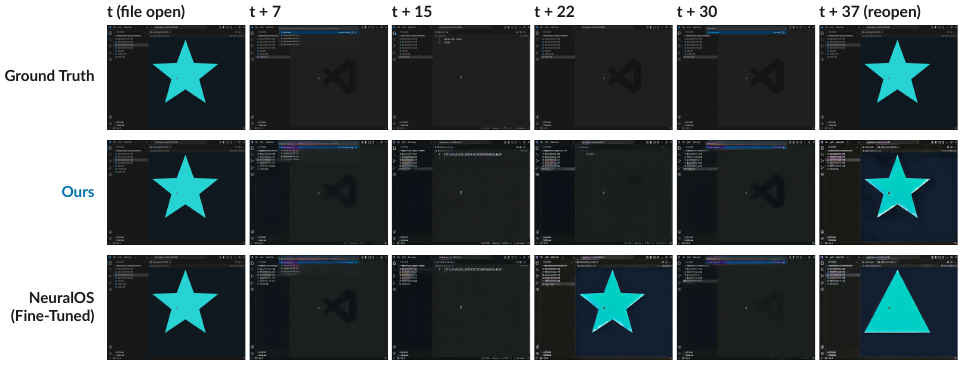}
  \vspace{-4pt}
  \caption{\textbf{\textbf{Ours} recovers the hidden star over $37$ closed-loop VS Code steps, whereas fine-tuned \textsc{NeuralOS} draws a triangle.}
  Student-forced: the model is given only its own previous output ($512{\times}384$; $32$ occluded steps in \texttt{todo.md}, $24.93$\,dB on reopen). \textsc{NeuralOS} also draws the star at $t{+}22$, while the file is still closed, before drawing a triangle on reopen.}
  \label{fig:suppl_rollout_revisit_n032_r4}
  \vspace{-4pt}
\end{figure*}

\begin{figure*}[t]
  \centering
  \sfbanner{after the first frame, the model is auto-regressively fed its own previous output.}
  \includegraphics[width=0.96\textwidth]{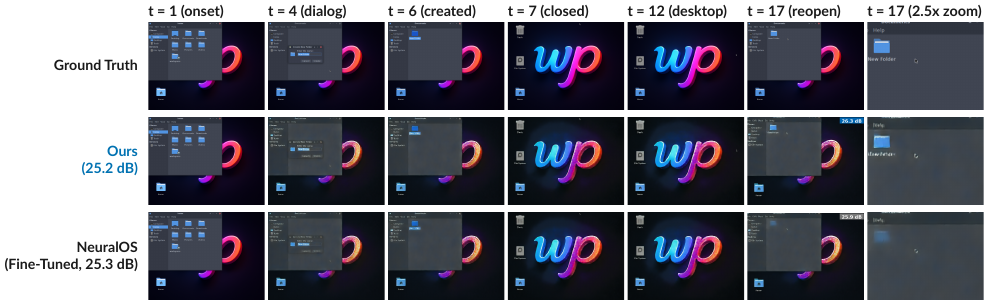}
  \vspace{-4pt}
  \caption{\textbf{\textbf{Ours} keeps the \texttt{New Folder} icon across $16$ student-forced File Manager steps.}
  Student-forced: every frame after $t{=}1$ is generated from the model's own previous output ($512{\times}384$, rightmost column $2.5{\times}$ zoom): \textbf{Ours} retains the icon and label, while fine-tuned \textsc{NeuralOS} nearly erases it. Whole-frame PSNR is the same for the two models ($25.2$ and $25.3$\,dB); they differ in the folder region shown in the zoom.}
  \label{fig:suppl_xfce_sf_r00}
  \vspace{-4pt}
\end{figure*}

\begin{figure*}[p]
  \centering
  \sfbanner{after the first frame, the model is auto-regressively fed its own previous output.}
  \includegraphics[width=0.96\textwidth]{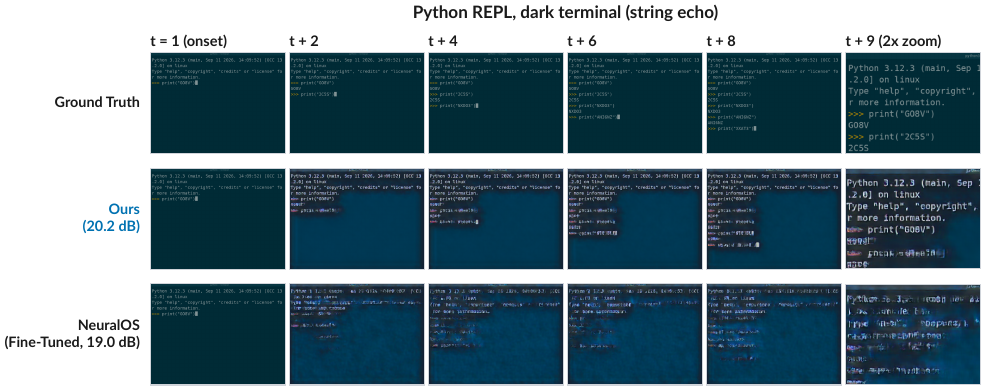}
  \vspace{-4pt}
  \caption{\textbf{\textbf{Ours} keeps the terminal header and the first command legible across $9$ student-forced REPL steps, while fine-tuned \textsc{NeuralOS} loses the header within $2$ steps.}
  Student-forced: the model is given only its own previous output. $512{\times}384$ rollout on a dark-theme terminal (rightmost column: $2{\times}$ zoom). Later lines blur for both models.}
  \label{fig:suppl_maths_sf}
\end{figure*}

Four further $512{\times}384$ rollout strips complement Main Figures~3 and~4.
Each is a single episode chosen by hand to illustrate a behavior; aggregate results are in Main Tables~1 and~2.
Each figure carries a banner stating its protocol.
In a \emph{teacher-forced} rollout every frame is predicted from the real frames before it, so the test isolates memory across the gap.
In a \emph{student-forced} rollout the model is given only its own previous output, so errors can compound.
Teacher-forced: on a second episode of the File Manager task of Main Figure~3, \textbf{Ours} redraws the \texttt{New Folder} icon after a $128$-step occlusion (\Cref{fig:suppl_xfce_desktop_g128}).
Student-forced: \textbf{Ours} redraws the hidden star after $37$ VS Code steps (\Cref{fig:suppl_rollout_revisit_n032_r4}), keeps the created folder across $16$ File Manager steps (\Cref{fig:suppl_xfce_sf_r00}), and keeps the terminal header legible across $9$ Python REPL steps (\Cref{fig:suppl_maths_sf}).

\section{Generalization to Unseen Desktops}
\label{sec:suppl:unseen}
\begin{figure*}[t]
  \centering
  \begin{minipage}[c]{0.5\textwidth}
    \includegraphics[width=\linewidth]{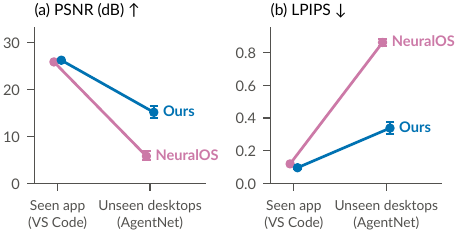}
  \end{minipage}\hfill
  \begin{minipage}[c]{0.46\textwidth}
  \caption{\textbf{Both models render VS Code with similar fidelity; on unseen desktops \textsc{NeuralOS} loses $20.0$\,dB and ours $11.1$\,dB.}
  Next-frame PSNR~(a; higher is better) and LPIPS~(b; a learned perceptual distance, lower is better) of the $512{\times}384$ models, both trained only on our VS Code and Python REPL recordings.
  Each frame is predicted from the true preceding frames and actions (teacher-forced).
  \emph{Seen app}: $30$ held-out VS Code episodes. \emph{Unseen desktops}: $30$ AgentNet episodes from Windows, macOS and Ubuntu.
  Each point averages the first $12$ frames of each episode, then the episodes; bars are $95\%$ bootstrap intervals over episodes.}
  \label{fig:suppl_seen_unseen}
  \end{minipage}
\end{figure*}

\begin{figure*}[t]
  \centering
  \includegraphics[width=\textwidth]{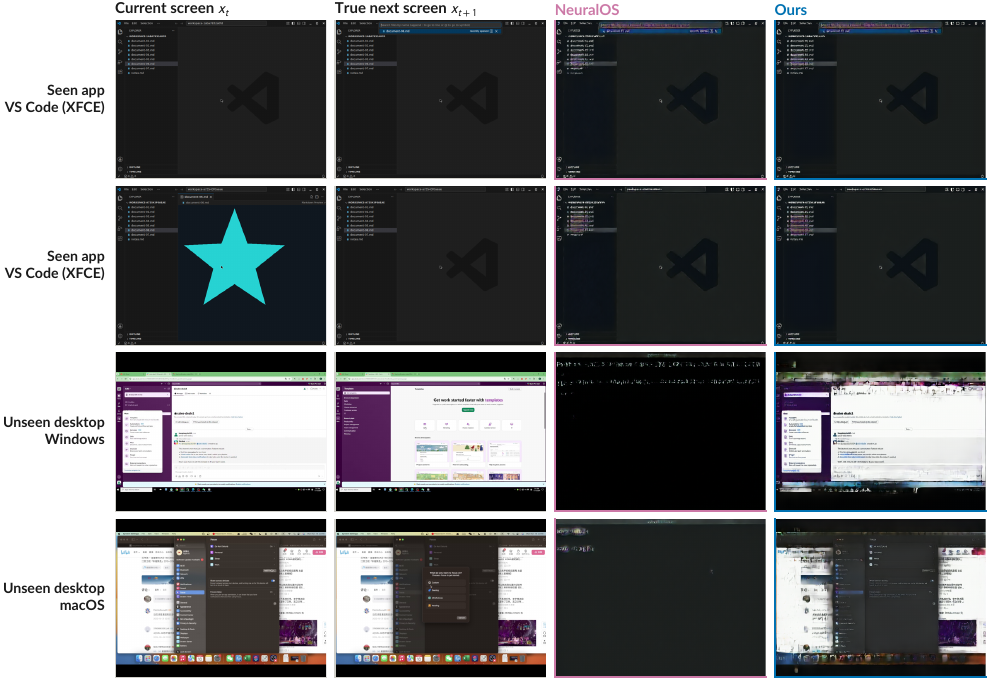}
  \caption{\textbf{On an unseen desktop \textsc{NeuralOS} replaces the screen with a dark one; ours keeps it.}
  Each row is one teacher-forced step: the model receives the true screens up to $\x_t$ and the user's action, and predicts $\x_{t+1}$.
  On VS Code (top two rows) both models redraw the screen correctly, including closing the file that shows the star.
  On Windows (third row) the true next screen is a new page of templates; on macOS (bottom row) a dialog opens.
  \textsc{NeuralOS} draws a dark screen, with illegible text in the Windows row; ours keeps the current screen with blotchy artifacts and draws neither the page nor the dialog.
  Rows are selected by rule: in each set, frames in which at least $2\%$ of pixels change are ranked by how much lower our LPIPS is than \textsc{NeuralOS}'s, and the frames at the one-third and two-thirds points are shown. Frames are at the native $512{\times}384$.}
  \label{fig:suppl_seen_unseen_examples}
\end{figure*}

\textbf{Ours} generalizes to unseen desktops better than \textsc{NeuralOS}~\cite{rivard2025neuralos}: trained on the same recordings, the two models render VS Code with similar fidelity, but on Windows, macOS and Ubuntu desktops \textsc{NeuralOS} loses $20.0$\,dB PSNR and ours $11.1$\,dB (\Cref{fig:suppl_seen_unseen}).

\paragraph{Setup.}
We compare the two $512{\times}384$ models of \Cref{tab:budgets}. Both start from the \textsc{NeuralOS} \texttt{joint-272k} checkpoint and are trained only on the Dense Interactive Mixture, our recordings of VS Code in an XFCE desktop (\emph{Desktop}) and of a Python REPL (\emph{Maths}); \textsc{NeuralOS} is fine-tuned for $9{,}000$ steps and ours for $3{,}000$.
Neither model has seen an AgentNet frame.
The \emph{seen-app} set is $30$ held-out VS Code episodes from \emph{Desktop}.
The \emph{unseen-desktop} set is $30$ AgentNet~\cite{wang2025opencua} dev episodes: $16$ from Windows, $9$ from macOS and $5$ from Ubuntu, covering browsers, office and presentation software, a terminal, system settings and maps; $4$ of them show VS Code under another operating system and theme.
Each frame is predicted teacher-forced, that is, from the true preceding frames and actions, so both models always receive the real current screen.
We score the first $12$ transitions of each episode, sampled with $32$ DDIM steps, for one training seed and one sampling seed per model.
We report PSNR and LPIPS~\cite{zhang2018unreasonable}, a learned perceptual distance computed on AlexNet features (lower is better), averaged within each episode and then over episodes, with $95\%$ bootstrap intervals over episodes.

\paragraph{Result.}
On VS Code the two models are close: $26.2$ against $25.8$\,dB PSNR and $0.097$ against $0.120$ LPIPS for ours and \textsc{NeuralOS}.
On unseen desktops \textsc{NeuralOS} falls to $5.8$\,dB and $0.86$ LPIPS and ours to $15.1$\,dB and $0.34$; the intervals do not overlap (\Cref{fig:suppl_seen_unseen}).
In the generated screens of \Cref{fig:suppl_seen_unseen_examples}, \textsc{NeuralOS} replaces a Windows or macOS screen with a dark screen, with illegible text in the Windows example, while ours keeps the screen it is given.
When an action opens a new page or a dialog on these desktops, ours keeps the old screen, with artifacts.
Our $1024{\times}768$ model, trained on the same mixture and scored on the same episodes after downsampling to $512{\times}384$, falls from $30.4$ to $19.2$\,dB.

\paragraph{Why \textsc{NeuralOS} fails.}
In \textsc{NeuralOS} the UNet that draws the next screen never sees the current one: the latent $\z_t$ of the current screen ($16{\times}48{\times}64 = 49{,}152$ numbers) enters the lower of two stacked LSTMs only as one attention read-out of $1{,}024$ numbers, and the UNet is conditioned only on an MLP projection of the upper LSTM's $4{,}096$-number output.
Whatever this state does not encode about the current screen must come from the UNet's weights, which have learned the screens of the training data.
The zero-shot result of \Cref{tab:suppl_neuralos_zeroshot} shows the same failure without our fine-tuning: the public \textsc{NeuralOS} release, shown our XFCE desktop, renders its own training desktop ($9.84$\,dB).

\section{Broader Impact and Ethics}
\label{sec:suppl:ethics}
Off-screen state lives in an explicit memory bank, so it can be inspected, replaced, or wiped without resetting the rest of the model (Main Section~5.3).
The main misuse risk is the generation of deceptive screenshots or `spoofed' application sessions, which motivates provenance watermarking for generative interface systems.
Our in-house data are XFCE virtual-machine recordings and contain no personal data.
The AgentNet examples in \Cref{fig:data_examples,fig:suppl_agentnet_active_gui} are screenshots distributed with that corpus; we show none that contain photographs of people or account details.
The public corpora are used under their licenses (\texttt{AgentNet} under MIT, \texttt{ProCUA-SFT} under CC BY 4.0, and \texttt{CUA-Gym} under CC0).
On publication, we will release the aligned five-source corpus, our conversion and alignment pipelines, and our training code.

\end{document}